\documentclass[11pt,a4paper,titlepage,oneside]{memoir}
\setlrmarginsandblock{4cm}{3cm}{*}
\setulmarginsandblock{4cm}{3cm}{*}

\usepackage[OT1]{fontenc}
\usepackage[british]{babel}
\usepackage[utf8]{inputenc}
\usepackage[sc]{mathpazo}
\usepackage{amsmath,amssymb,amsfonts,mathrsfs}
\usepackage[amsmath,thmmarks]{ntheorem}
\usepackage{graphicx}

\usepackage{caption}
\usepackage{subcaption}
\usepackage{longtable}
\usepackage{wrapfig}
\usepackage{enumitem}
\usepackage{pdfpages}
\usepackage{float}
\usepackage[export]{adjustbox}
\usepackage{algorithm}
\usepackage[noend]{algpseudocode}
\usepackage[urldate=long,backend=biber,sorting=none,natbib=true,style=numeric]{biblatex}

\usepackage{multirow}

\usepackage{varioref}

\usepackage{datetime}

\usepackage{mathtools}

\usepackage[h]{esvect}

\usepackage{array}

\usepackage{pstricks,pst-all}

\usepackage{listings}
\usepackage[activate]{pdfcprot}

\usepackage{booktabs}

\usepackage{ETHlogo}

\nonzeroparskip
\defaultlists

\makeatletter

\if@twoside
  \copypagestyle{chapter}{Ruled}
\else
  \copypagestyle{chapter}{ruled}
\fi
\makeoddhead{chapter}{}{}{}
\makeevenhead{chapter}{}{}{}
\makeheadrule{chapter}{\textwidth}{0pt}
\copypagestyle{abstract}{empty}

\makechapterstyle{bianchimod}{%
  \chapterstyle{default}
  \renewcommand*{\chapnamefont}{\normalfont\Large\sffamily}
  
  \renewcommand*{\printchaptername}{%
    \chapnamefont\centering\@chapapp}

  }

\chapterstyle{bianchimod}

\setsecheadstyle{\Large\bfseries\sffamily}
\setsubsecheadstyle{\large\bfseries\sffamily}
\setsubsubsecheadstyle{\bfseries\sffamily}
\setparaheadstyle{\normalsize\bfseries\sffamily}
\setsubparaheadstyle{\normalsize\itshape\sffamily}
\setsubparaindent{0pt}

\captionnamefont{\sffamily\bfseries\footnotesize}
\captiontitlefont{\sffamily\footnotesize}
\setsecnumdepth{subsection}
\settocdepth{subsection}

\pretitle{\vspace{0pt plus 0.7fill}\begin{center}\HUGE\sffamily\bfseries}
\posttitle{\end{center}\par}
\preauthor{\par\begin{center}\let\and\\\Large\sffamily}
\postauthor{\end{center}}
\predate{\par\begin{center}\Large\sffamily}
\postdate{\end{center}}

\def\@advisors{}
\newcommand{\advisors}[1]{\def\@advisors{#1}}
\def\@department{}
\newcommand{\department}[1]{\def\@department{#1}}
\def\@thesistype{}
\newcommand{\thesistype}[1]{\def\@thesistype{#1}}

\renewcommand{\maketitlehookb}{\vspace{1in}%
  \par\begin{center}\Large\sffamily\@thesistype\end{center}}

\renewcommand{\maketitlehookd}{%
  \vfill\par
  \begin{flushright}
    \sffamily
    \@advisors\par
    \@department, ETH Z\"urich
  \end{flushright}
}

\checkandfixthelayout

\makeatother

\theoremstyle{plain}
\numberwithin{equation}{chapter}

\theoremstyle{nonumberplain}
\theorembodyfont{\normalfont}
\theoremsymbol{\ensuremath{\square}}

\newcommand{\R}{\mathbb{R}}

\renewcommand{\epsilon}{\ensuremath\varepsilon}

\renewcommand{\phi}{\ensuremath{\varphi}}

\usepackage{eso-pic}
\usepackage{xspace}
\usepackage[justification=centering]{caption}
\usepackage{siunitx}
\usepackage{tikz}
\usepackage{calc}
\usepackage{pgf}
\usepackage{fp}
\usetikzlibrary{shapes,arrows,positioning,shadows,calc,fit,arrows,decorations.pathreplacing,decorations.markings,decorations.pathmorphing,patterns,shapes,shapes.multipart}
\pgfdeclarelayer{bg}
\pgfsetlayers{bg,main}

\newcommand{\argmin}{\operatorname*{arg\;min}}

\makeatletter
\def\BState{\State\hskip-\ALG@thistlm}
\makeatother

\usepackage{xparse}
\NewDocumentCommand{\ceil}{s O{} m}{%
	\IfBooleanTF{#1} 
	{\left\lceil#3\right\rceil} 
	{#2\lceil#3#2\rceil} 
}

\makeatletter
\DeclareRobustCommand\onedot{\futurelet\@let@token\@onedot}
\def\@onedot{\ifx\@let@token.\else.\null\fi\xspace}

\def\etal{\emph{et al}\onedot}
\makeatother

\usepackage{xcolor}	 
\definecolor{mygreen}{RGB}{44,85,17}
\definecolor{myblue}{RGB}{34,31,217}
\definecolor{mybrown}{RGB}{194,164,113}
\definecolor{myred}{RGB}{255,66,56}

\title{Efficient Convolutional Neural Networks for Pixelwise Classification on Heterogeneous Hardware Systems}
\author{Fabian Tschopp}
\thesistype{Technical Report}
\advisors{Supervisors: Prof.\ Dr.\ Angelika Steger, Dr.\ Jan Funke}
\department{Department of Computer Science}
\date{September 11, 2015}

\bibliography{technical_report.bib}
\begin{document}

\frontmatter

\begin{titlingpage}
  \calccentering{\unitlength}
  \begin{adjustwidth*}{\unitlength-24pt}{-\unitlength-24pt}
    \maketitle
  \end{adjustwidth*}
\end{titlingpage}

\begin{abstract}
This work presents and analyzes three convolutional neural network (CNN) models for efficient pixelwise classification of images. When using convolutional neural networks to classify single pixels in patches of a whole image, a lot of redundant computations are carried out when using sliding window networks.
This set of new architectures solve this issue by either removing redundant computations or using fully convolutional architectures that inherently predict many pixels at once.

The implementations of the three models are accessible through a new utility on top of the Caffe library. The utility provides support for a wide range of image input and output formats, pre-processing parameters and methods to equalize the label histogram during training. The Caffe library has been extended by new layers and a new backend for availability on a wider range of hardware such as CPUs and GPUs through OpenCL.

On AMD GPUs, speedups of $54\times$ (SK-Net), $437\times$ (U-Net) and $320\times$ (USK-Net) have been observed, taking the SK equivalent SW (sliding window) network as the baseline. The label throughput is up to one megapixel per second.

The analyzed neural networks have distinctive characteristics that apply during training or processing, and not every data set is suitable to every architecture.
The quality of the predictions is assessed on two neural tissue data sets, of which one is the ISBI 2012 challenge data set. Two different loss functions, Malis loss and Softmax loss, were used during training.

The whole pipeline, consisting of models, interface and modified Caffe library, is available as Open Source software under the working title \textit{Project Greentea}.

\end{abstract}

\newpage
\renewcommand{\abstractname}{Acknowledgements}
\begin{abstract}
\subsection*{University of Zurich, Institute of Neuroinformatics}
Firstly I would like to express my gratitude to my supervisor Dr. Jan Funke for his guidance, motivation and the opportunity to visit HHMI Janelia in Ashburn, Virginia, USA.\\
I also thank my supervisor Prof. Dr. Angelika Steger for collaborating with the Institute of Neuroinformatics, which made this research project possible. I thank Stephan Gerhard and Julien Martel for interesting discussions about neural networks and this technical report. 

\subsection*{Howard Hughes Medical Institute, Janelia}
Besides my advisors I would like to thank Dr. Srinivas Turaga and Dr. Stephan Saalfeld for their collaboration at HHMI Janelia, which inspired me to extend the scope of my research and gave me insight into the applications of neural networks for image segmentation in connectomics. Besides this, Janelia has the nicest campus of all research institutes that I have seen so far.

\subsection*{AMD (Advanced Micro Devices)}
I would like to thank AMD and especially Roy Taylor, Greg Stoner and Bruno Stefanizzi for the generous hardware sponsoring, which empowered a lot of the development on the Caffe library and enabled me to use neural network models beyond what is possible on regular hardware.
I also thank Timmy Liu for his assistance and development of clBLAS and Dr. Ing. Hervé Chevanne for providing drivers and support for the AMD GPUs.\\
Being a fan of AMD hardware and using their devices for over ten years, it was a pleasure for me to work together with AMD engineers and using their newest hardware and software technology for research.

\subsection*{Family}
Last but not the least, I would like to thank my family, my parents and my sister, for supporting me during my Bachelor studies at ETH Zurich.

\end{abstract}

\cleartorecto
\tableofcontents
\mainmatter
\chapter{Introduction}
\section{Convolutional Neural Networks}
Convolutional neural networks are forward-backward neural networks that are mostly based on convolutions with machine learnable kernels, pooling operations and element-wise non-linear activation functions. The networks can be employed for various image classification and object recognition tasks. A prominent example is the ImageNet / AlexNet \cite{NIPS2012_4824} for object recognition. Recent networks \cite{2015arXiv150504597R} can have very many, in this case over 20, layers and millions of learnable parameters.

This work is focused on classifying biomedical data, in particular neural tissue electron microscopy images (see Chapter \ref{ch:datasets}). The challenge with this kind of data sets is that training data is more scarce than with data sets that can be generated from everyday pictures such as handwritten letters or online collections of images. Annotating ground truth for neural tissue images is a lot of manual work, as every single pixel has to be labeled.

Consequently, improving training speeds is not a primary objective to optimize for. The data that has to be processed with a trained model afterwards however can easily reach terabyte-scale. It is therefore crucial to develop networks that are as efficient as possible in the forwarding step. This work presents three such efficient pixel classification networks.

When training, the classical mislabeling objectives such as Softmax or Cross-entropy loss might not be the most useful for pixelwise classifications of biological images, and using spatial context information to generate an error signal to train the neural networks can perform better. Therefore, this research also considers different training methods including the Malis \cite{2009arXiv0911.5372T} criterion.

To relate the objectives of this technical report with the title, it needs to be dissected into its components:
\begin{itemize}
	\item \textit{Efficient Convolutional Neural Networks} means the network models analyzed and designed are as efficient as possible in getting the task done - in this case, pixelwise classification of electron microscopy neural tissue images. 
	\item \textit{Pixelwise Classification}, as opposed to image classification, aims to propose a label to each pixel in a given image. It can also be seen as many separate image classifications of small patches in a bigger image. This gives rise to new optimization possibilities as the contexts for the predictions overlap spatially.
	\item \textit{Heterogeneous Hardware Systems} means the network models used should also run as efficient as possible on a variety of compute devices. This objective makes efficient neural networks more accessible to users and allows to use existing hardware and clusters to get segmentation tasks with neural networks done.
\end{itemize}

\section{Caffe Library}
Caffe stands for Convolutional Architecture for Fast Feature Embedding \cite{jia2014caffe}. It is a state-of-the-art neural network library that has been heavily optimized for the use with nVidias CUDA technology. In many cases, the library is therefore already very efficient using certain GPUs. What was missing until now \cite{Caffe} is fast CPU support (the current CPU backend is mostly single threaded) and support for GPUs and accelerator devices from AMD and Intel. The library is still under development and has a large community \cite{BVLCCaffe}.

Network models and trained weights (usually called model zoo) can be shared in Google's prototxt (network and learning configurations) and protocol buffer (trained weights and solver states) format.

The library is typically used on the command line with the Caffe binary or through a python (Pycaffe) interface (see Figure \ref{fig:greenteaoverview}). For more advanced and intrusive interfaces, C++ interfaces can be programmed on top of the library.

All models, utilities and backend additions programmed for this project are based on and around Caffe. The changes to the library are documented in Chapter \ref{ch:caffelibrary}.

\section{Pixelwise Classification}

\begin{figure}[H]
	\center
	\begin{subfigure}{0.5\textwidth}
		\center
		\includegraphics[scale=0.30]{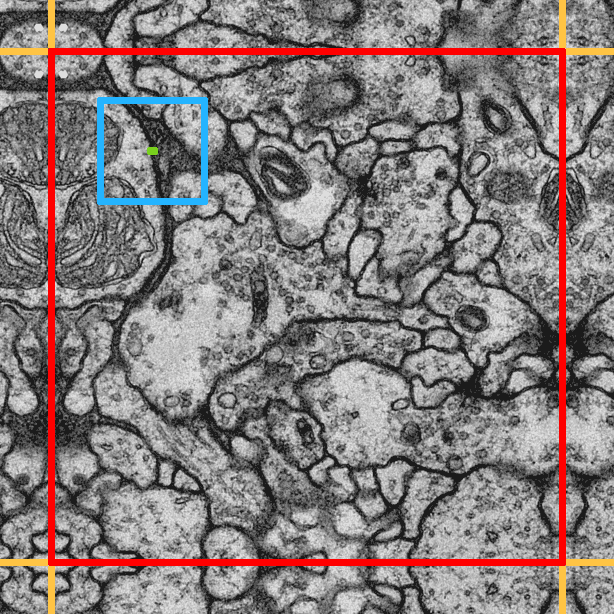}
		\caption{Input image to classify with border mirroring to extend the image. The green rectangle is a 4 by 1 pixel area to be labeled.}
	\end{subfigure}%
	\begin{subfigure}{.5\textwidth}
		\center
		\includegraphics[scale=0.30]{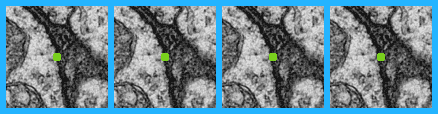}
		\caption{Generated minibatch input of size $n = 4$ and with a context of 102 by 102 pixels. The output classification for this input will be 4 by 1 pixels. The individual images in the minibatch are only shifted by one pixel each. The data overlaps and is copied into the network redundantly.}
	\end{subfigure}%
	\caption{Pixelwise image classification based on sliding window architectures. (Raw image source: ssTEM \cite{DS1ssTEM}, \cite{CaffeNeuralModels}). The surrounding context (blue rectangles) is what determines the labeling decision of the neural network.}
	\label{fig:pixelsliding}
\end{figure}

Pixelwise classification means labeling each pixel in an image based on a local context around the pixel. Figure \ref{fig:pixelsliding} shows how this works with sliding window networks that outputs a single pixel per input tile of size $v + w = 101 + 1 = 102$.

\begin{figure}[H]
	\center
	\begin{subfigure}{0.5\textwidth}
		\center
		\includegraphics[scale=0.30]{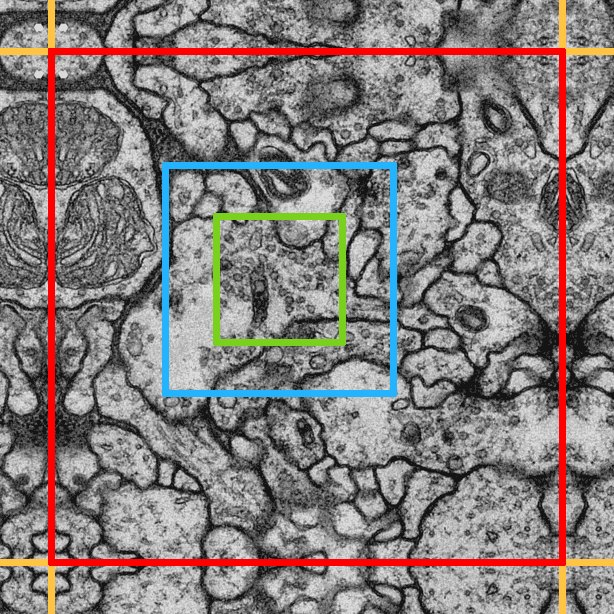}
		\caption{The green area to be labeled is 128 by 128 pixels ($w = 128$). The green patch with the blue context padding ($v = 101$) is directly what the networks take as input.}
	\end{subfigure}%
	\begin{subfigure}{.5\textwidth}
		\center
		\includegraphics[scale=0.30]{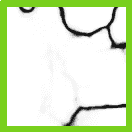}
		\caption{The data is passed through the network as large a tile of size $w+v$ by $w+v$ instead of a minibatch. There is no overlapping data being passed through the network redundantly and no duplicated convolution and pooling operations are carried out. The output prediction is a large patch ($w = 128$) instead of a stride of pixels from a minibatch (as in Figure \ref{fig:pixelsliding}).}
	\end{subfigure}%
	\caption{Pixelwise image classification based on strided kernel and fully convolutional architectures. (Raw image source: ssTEM \cite{DS1ssTEM}, \cite{CaffeNeuralModels}).}
	\label{fig:pixelpatch}
\end{figure}

While minibatch processing can output many pixels at once, this is still inefficient (see Sections \ref{sec:swnet} and \ref{sec:labelthroughput}). The work by Hongsheng Li \etal \cite{2014arXiv1412.4526L} allows to make existing SW networks more efficient while giving identical prediction results (see Section \ref{sec:swtosk}).\\ Alternatively, fully convolutional models (U and partially also USK) directly output a bigger patch, as depicted in Figure \ref{fig:pixelpatch}. This method of training and processing is called patch-based ($n = 1$, $w > 1$), as opposed to minibatch-based ($n > 1$, $w = 1$). A combination of both ($n > 1$, $w > 1$) is possible but only useful when the images in the data set can not be tiled with large $w \gg 1$ (see Section \ref{sec:devicememory}).

Minibatches can still have advantages during training (see Section \ref{sec:histeq}) because every element in the minibatch can be picked independently. During processing, networks that output large patches are always performing better.

In all cases, the images have to be extended (padded) by mirroring on the borders by $\frac{v}{2}$ pixels on each side if every pixel of the image is to be labeled.

\newpage
\section{Existing Work}

This technical report is based on the following existing work:
\begin{itemize}
	\item SW (sliding window) network designed by Julien Martel \cite{JulienMartel}, not published. The architecture is trimmed for segmenting the data set DS1.
	\item Strided kernel convolution and pooling kernels by Hongsheng Li \etal \cite{2014arXiv1412.4526L}. This is the fundamental approach in speeding up existing SW networks.
	\item Malis criterion, first introduced by Srinivas Turaga \etal \cite{2009arXiv0911.5372T}. The criterion supports an alternative way of training neural networks through affinity graphs, which is very specific and useful on biomedical data, where areas are separated by background borders.
	\item The Open Source Caffe library maintained by the Berkeley Vision and Learning Center \cite{BVLCCaffe}, \cite{jia2014caffe}.
	\item U network designed by Ronneberger \etal \cite{2015arXiv150504597R}. This model is also optimized for biomedical images and especially the data set DS2 (ISBI 2012 \cite{ISBI2012}).
	\item N-dimensional convolution kernels by Jeff Donahue \cite{ndconv}.
	\item Segmentation evaluation scripts of the ISBI 2012 challenge \cite{ISBI2012}, \cite{fijiscript}.
\end{itemize}

\section{New Contributions}
An overview of new contributions to the Caffe landscape in terms of models, utilities and library changes is given in Figure \ref{fig:greenteaoverview}.
This work tackles the given problem on all levels - from using efficient BLAS libraries over backend development and frontends for easy use to new network models.

On the side of neural network models, this project introduces two new neural network architectures, the SK-Net (Section \ref{sec:sknet}) and USK-Net (Section \ref{sec:usknet}).

A meta-analysis of the three efficient networks (SK, U, USK) is given based on:
\begin{itemize}
	\item Differences and characteristics of the network designs (Chapter \ref{ch:models}).
	\item Computational cost and efficiency (Chapter \ref{ch:benchmarks}).
	\item Image segmentation quality, assessed both numerically and visually (Chapter \ref{ch:results}) for the typical foreground-background two label classification.
\end{itemize}

In order to be able to train the network models easily on various data sets, the Caffe Neural Tool (Chapter \ref{ch:caffeneuraltool}) has been developed.

The Caffe library (see Chapter \ref{ch:caffelibrary}) has been extended with new layers and adaptions for compability with the new backend. The new layers also affect the functionality of the CUDA backend. OpenCL backend development (see Section \ref{sec:openclbackend}) was mostly focused on versatility and completeness, so that CPUs and all kinds of compute devices can be used on all network models. This includes compability to three different BLAS libraries: clBLAS, ViennaCL-BLAS and cBLAS.

\tikzstyle{cred} = [circle, draw, top color=white, bottom color=red!30, draw=red!50!black!100, drop shadow, text width=0.5cm, text centered, rounded corners, minimum height=0.5cm]
\tikzstyle{cgreen} = [circle, draw, top color=white, bottom color=green!30, draw=green!50!black!100, drop shadow, text width=0.5cm, text centered, rounded corners, minimum height=0.5cm]
\tikzstyle{cblue} = [circle, draw, top color=white, bottom color=blue!30, draw=blue!50!black!100, drop shadow, text width=0.5cm, text centered, rounded corners, minimum height=0.5cm]

\tikzstyle{plain} = [text width = 2.2cm, minimum height = 0.5cm]

\tikzstyle{sred} = [rectangle, draw, top color=white, bottom color=red!30, draw=red!50!black!100, drop shadow, minimum width=1cm, text centered, minimum height=0.5cm]
\tikzstyle{sgreen} = [rectangle, draw, top color=white, bottom color=green!30, draw=green!50!black!100, drop shadow, minimum width=1cm, text centered, minimum height=0.5cm]
\tikzstyle{sblue} = [rectangle, draw, top color=white, bottom color=blue!30, draw=blue!50!black!100, drop shadow, minimum width=1cm, text centered, minimum height=0.5cm]

\tikzstyle{sline} = [line width=1.5pt, draw, -latex']
\tikzstyle{dline} = [line width=1.5pt, draw, -latex', dotted]

\tikzstyle{box} = [rectangle, draw, top color=white, bottom color = yellow!30, draw=yellow!50!black!100, rounded corners]

\begin{figure}[H]
	\centering
	\begin{tikzpicture}[node distance = 2cm and 1cm, auto]
		\node [plain] (models) {\phantom{x}\\Models\\\phantom{x}};
		\node [plain, below of = models] (interfaces) {Frontend\\Interface\\\phantom{x}};
		\node [plain, below of = interfaces] (library) {\phantom{x}\\Library\\\phantom{x}};
		\node [plain, below of = library] (backend) {Backend /\\Compute\\Kernels};
		\node [plain, below of = backend] (blas) {\phantom{x}\\BLAS\\\phantom{x}};
		
		\node [right=9.0cm of models] (models_term) {};
		\node [right=9.0cm of interfaces] (interfaces_term) {};
		\node [right=9.0cm of library] (library_term) {};
		\node [right=9.0cm of backend] (backend_term) {};
		\node [right=9.0cm of blas] (blas_term) {};
		
		\node [sblue, right=0.0cm of models] (mod_sw) {SW};	
		\node [sgreen, right=2.5cm of mod_sw] (mod_sk) {SK};
		\node [sred, right=1.0cm of mod_sk] (mod_u) {U};
		\node [sgreen, right=1.0cm of mod_u] (mod_usk) {USK};
		
		\node [sblue, right=0.0cm of interfaces] (caffe_if) {Caffe binary};
		\node [sblue, right=1.0cm of caffe_if] (pycaffe_if) {Pycaffe};
		\node [sgreen, right=1.0cm of pycaffe_if] (caffe_nt_if) {Caffe Neural Tool};
		
		\node [sred, right=1.5cm of library] (caffe) {Caffe shared or static library};
				
		\node [sgreen, right=0.0cm of backend, align=center] (back_ocl) {Greentea\\(OpenCL)};
		\node [sblue, right=2.3cm of back_ocl, align=center] (back_cpu) {Caffe\\(CPU native)};
		\node [sred, right=0.5cm of back_cpu, align=center] (back_cuda) {Caffe\\(CUDA)};
		
		\node [sgreen, right=0.0cm of blas, align=center] (viennaclblas) {ViennaCL\\BLAS};
		\node [sgreen, right=0.5cm of viennaclblas, align=center] (clblas) {clBLAS\\\phantom{x}};
		\node [sblue, right=0.9cm of clblas, align=center] (cblas) {cBLAS\\\phantom{x}};
		\node [sblue, right=1.0cm of cblas, align=center] (cublas) {cuBLAS\\\phantom{x}};
			
		\path [dline] (mod_sw) -- (caffe_if);
		\path [dline] (mod_sw) -- (pycaffe_if);
		\path [sline] (mod_sk) -- (caffe_nt_if);
		\path [sline] (mod_sk) -- (pycaffe_if);
		\path [sline] (mod_u) -- (caffe_nt_if);
		\path [sline] (mod_u) -- (pycaffe_if);
		\path [sline] (mod_usk) -- (caffe_nt_if);
		\path [sline] (mod_usk) -- (pycaffe_if);
		
		\path [sline] (caffe_if) -- (caffe);
		\path [sline] (pycaffe_if) -- (caffe);
		\path [sline] (caffe_nt_if) -- (caffe);

		\path [sline] (caffe) -- (back_ocl);
		\path [dline] (caffe) -- (back_cpu);
		\path [sline] (caffe) -- (back_cuda);
		
		\path [sline] (back_ocl) -- (viennaclblas);
		\path [sline] (back_ocl) -- (clblas);
		\path [sline] (back_ocl) -- (cblas);
		
		\path [sline] (back_cpu) -- (cblas);
				
		\path [sline] (back_cuda) -- (cublas);
				
		\begin{pgfonlayer}{bg}
			\node[box,fit=(models) (models_term)] {};
		\end{pgfonlayer}
		
		\begin{pgfonlayer}{bg}
			\node[box,fit=(interfaces) (interfaces_term)] {};
		\end{pgfonlayer}
		
		\begin{pgfonlayer}{bg}
			\node[box,fit=(library) (library_term)] {};
		\end{pgfonlayer}
	
		\begin{pgfonlayer}{bg}
			\node[box,fit=(backend) (backend_term)] {};
		\end{pgfonlayer}	
	
		\begin{pgfonlayer}{bg}
			\node[box,fit=(blas) (blas_term)] {};
		\end{pgfonlayer}	
	\end{tikzpicture}
	\caption{\textit{Project Greentea} overview. Green boxes denote completely new additions to the Caffe landscape. Red boxes are parts that have been re-implemented or adapted from existing work to fit the needs of this project. Blue parts are mostly unchanged from existing work. Dashed arrows denote deprecated and only partially supported combinations.}
	\label{fig:greenteaoverview}
\end{figure}
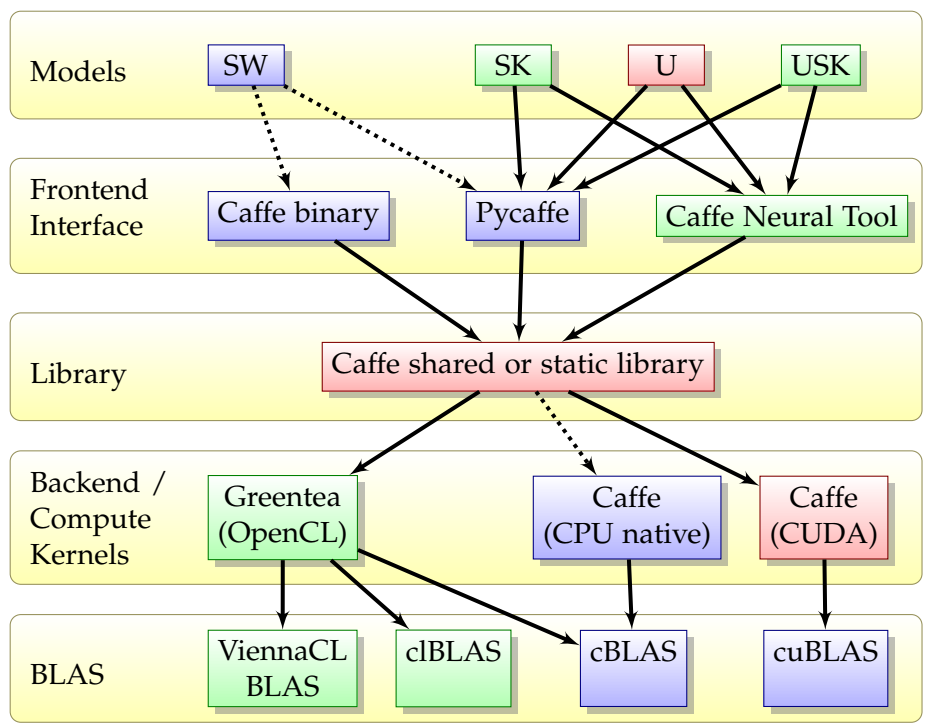

This report is also providing an introduction into segmentation and pixelwise classification with neural networks. It contains all the details necessary to understand existing models and readers should be able to easily design their own neural networks based on the findings of this research project.

The combination of the new contributions and the Caffe library infrastructure is summarized under the working title \textit{Project Greentea}. However, \textit{Greentea} also stands for the new OpenCL backend architecture which has been optimized for high flexibility (see Section \ref{sec:openclbackend}).

\textit{Greentea} was a name of my choice because I simply prefer greentea over coffee (Caffe). Greentea can not be used as an abbreviation like it is the case with Caffe, but as greentea is supposed to be good for the brain, using it for a neural network machine learning toolset seems to be appropriate.

\newpage
\section{Terminology}
The most used symbols and abbreviations in the report:
\begin{itemize}
	\item \textit{Forwarding, processing}: Computing data through a neural network from input to output.
	\item \textit{Backwarding, training, backpropagation}: Computing neural network gradients (diff maps) in the backward direction and updating the network weights.
	\item \textit{Data blob}: Memory blob containing feature maps of forward processing in the neural network.
	\item \textit{Diff blob}: Memory blob containing the differential / error signal map during backpropagation.
	\item \textit{BLAS}: Basic Linear Algebra Subprograms. Includes functions such as efficient matrix multiplications.
	\item \textit{DS1}: Data set 1, see Section \ref{sec:dataset1}.
	\item \textit{DS2}: Data set 2, see Section \ref{sec:dataset2}.
	\item \textit{SW}: Sliding window networks for pixelwise classification, see Section \ref{sec:swnet}.
	\item \textit{SK}: Strided kernel networks for pixelwise classification, see Section \ref{sec:sknet}.
	\item \textit{U}: Ronneberger \etal \cite{2015arXiv150504597R} network architecture, see Section \ref{sec:unet}.
	\item \textit{USK}: Network architecture combining SK and U aspects, see Section \ref{sec:usknet}.
	\item \textit{$f$}: Number of feature maps after ($f_{\text{out}}$) and before ($f_{\text{in}}$) a network layer.
	\item \textit{$w$}: Size (in each dimension) of a feature map in a network layer. When not indexed or otherwise noted, it refers to the output size of a network layer.
	\item \textit{$v$}: Size of the total network input padding (context).
	\item \textit{$p$}: Network layer padding.
	\item \textit{$s$}: Network layer stride.
	\item \textit{$k$}: Network layer kernel size.
	\item \textit{$d$}: Network layer kernel stride.
	\item \textit{$L$}: Set of layers with layers $l \in L$.
	\item \textit{$B$}: Set of memory blobs with blobs $b \in B$.
	\item \textit{$W$}: Set of network weights, $|W|$ denotes the number of weights.
	\item \textit{$M$}: Device or host memory usage.
	\item \textit{$q$}: Number of queues in the Caffe OpenCL backend.
	\item \textit{$n$}: Network minibatch size.
	\item \textit{$A$}: Affinity graph data, $\Delta A$ denotes the affinity graph diff.
	\item \textit{$I$}: Pixel image data, $\Delta I$ denotes the image diff.
\end{itemize}

Some symbols are used differently in some sections of the report and are explained in-place.
\chapter{Datasets}
\label{ch:datasets}
\section{DS1 - Segmented anisotropic ssTEM dataset of neural tissue}
\label{sec:dataset1}

\begin{figure}[H]
	\center
	\includegraphics[scale=0.5]{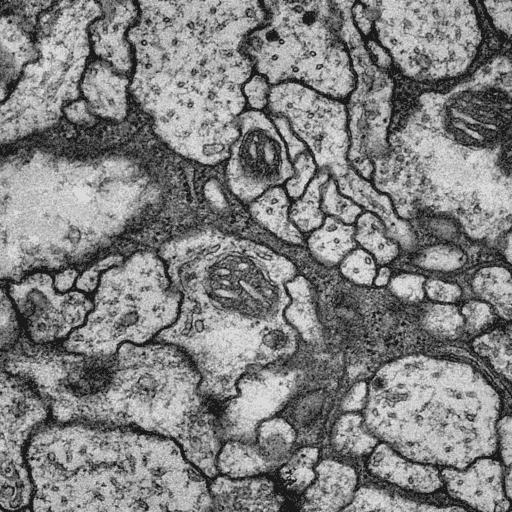}
	\caption{DS1 ssTEM raw image, 512 by 512 pixels of image 2 (right upper corner) (Source: ssTEM \cite{DS1ssTEM}, \cite{CaffeNeuralModels}).}
	\label{fig:ds1datasetraw}
\end{figure}

This data set shows neural tissue from a Drosophila larva ventral nerve cord and was acquired using serial section Transmission Electron Microscopy at HHMI Janelia Research Campus \cite{DS1ssTEM}.
The training data consists of 20 images of 1024 by 1024 pixels raw ssTEM and the corresponding segmentation.
It is segmented into nine different labels, which are consolidated into foreground and background for two label training and evaluation (see Section \ref{sec:ds1analysis}):
\begin{itemize}
	\item \#0: Horizontal cell membranes - background
	\item \#1: +\ang{45} to vertical cell membranes - background
	\item \#2: Vertical cell membranes - background
	\item \#3: -\ang{45} to vertical cell membranes - background
	\item \#4:  Cell membrane junctions - background
	\item \#5: Glia cells - background
	\item \#6: Mitochondria - foreground
	\item \#7: Synapses - background
	\item \#8: Cell interior - foreground
\end{itemize}
The idea behind label consolidation in this way is that the network, during training, can learn the separation borders between neural cells. This is especially important with the Malis criterion loss (see Section \ref{sec:malisloss}), which can only segment into foreground and background. With the Softmax loss, it is also possible to let the network learn the labels separately and combine them accordingly afterwards. The network has to learn the features separately either way, as the membrane for example depends on different orientations in the convolution filters. This means a network can be trained on two labels only and afterwards, all nine labels can be extracted by applying a short fine-tuning training phase to the network. 

\begin{figure}[H]
	\center
	\begin{subfigure}{0.5\textwidth}
		\center
		\includegraphics[scale=0.37]{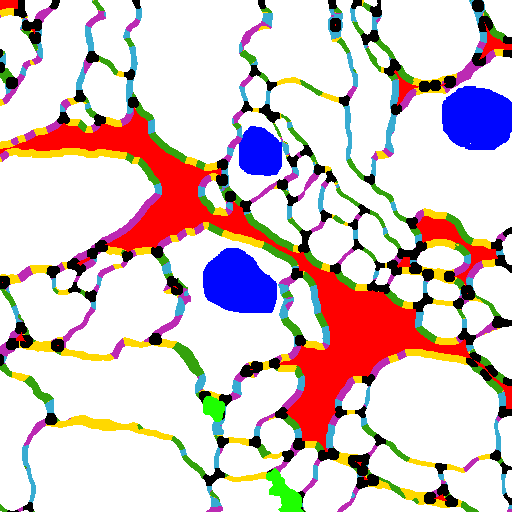}
		\caption{All 9 labels, the blue mitochondria and white cell interiors are the foreground.}
	\end{subfigure}%
	\begin{subfigure}{.5\textwidth}
		\center
		\includegraphics[scale=0.37]{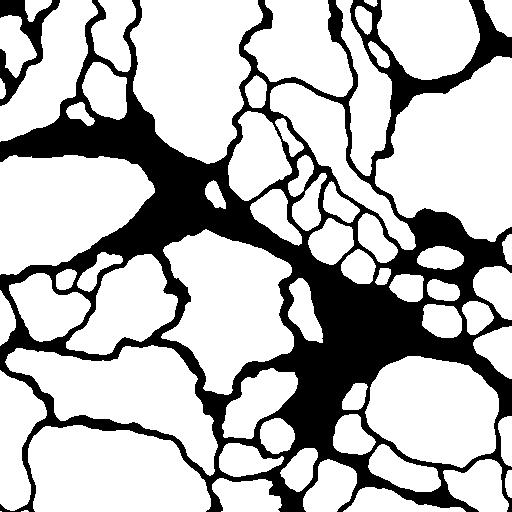}
		\caption{Consolidated labels, background-foreground segmentation.}
	\end{subfigure}%
	\caption{DS1 ssTEM label images, 512 by 512 pixels of image 2, corresponding to the raw image in Figure \ref{fig:ds1datasetraw} (Source: ssTEM \cite{DS1ssTEM}, \cite{CaffeNeuralModels}).}
	\label{fig:ds1datasetlabels}
\end{figure}

For evaluation, the training data set has been split into training and testing data because of the lack of a segmented test data stack:
\begin{itemize}
	\item Train images: 0, 1, 3, 4, 5, 6, 8, 9, 10, 11, 13, 14, 15, 16, 18 and 19 
	\item Test images: 2, 7, 12 and 17
\end{itemize}
As all slices of the data set are very similar, cross validation by splitting the data set in different ways was not applied. The total amount of pixels for training is therefore $16 \cdot 1024^2 \approx \SI{16}{Mpixel}$.

The data set with the corresponding test and train scripts is available in the Caffe Neural Models repository \cite{CaffeNeuralModels} as \textit{dataset\_01}.

\section{DS2 - ISBI 2012 dataset of neural tissue}
\label{sec:dataset2}
\begin{figure}[H]
	\center
	\includegraphics[scale=0.5]{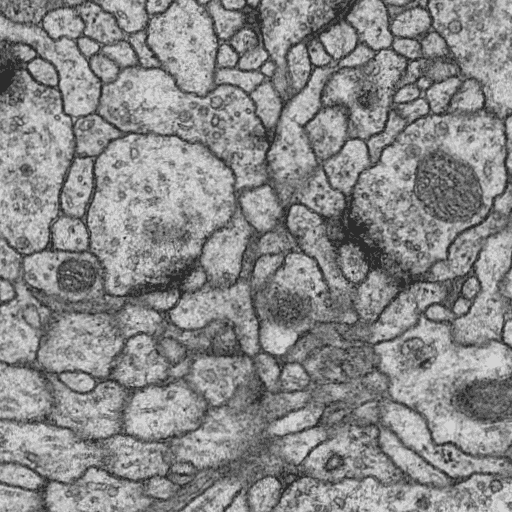}
	\caption{DS2 raw image, 512 by 512 pixels of training image 1 (Source: ISBI challenge \cite{ISBI2012}, \cite{CaffeNeuralModels}).}
	\label{fig:ds2datasetraw}
\end{figure}

This data set is from the ISBI 2012 challenge \cite{ISBI2012}, \cite{TrackEM}, \cite{Cardona2010}.
The raw images and corresponding segmentation images for training are 512 by 512 pixels. The neural tissue features have a similar scale to the data set DS1. The raw images have a bit less contrast and are more fuzzy.

The training set has 30 images, which gives a total of $30 \cdot 512^2 \approx \SI{7.8}{Mpixel}$, which is about half as much as on DS1.

The test data used on DS2 is a separate stack of 30 images of size 512 by 512 pixels. For those images, segmentation ground truth is not available for public download, thus the evaluation reported in Section \ref{sec:ds2analysis} is solely based on the reports of the official ISBI 2012 evaluation \cite{ISBI2012}, which is still open for new results.

For both stacks, the data spans 2 x 2 x 1.5 microns with a resolution of 4 by 4 by 50 nm/pixel \cite{ISBI2012}.

\begin{figure}[H]
	\center
	\includegraphics[scale=0.5]{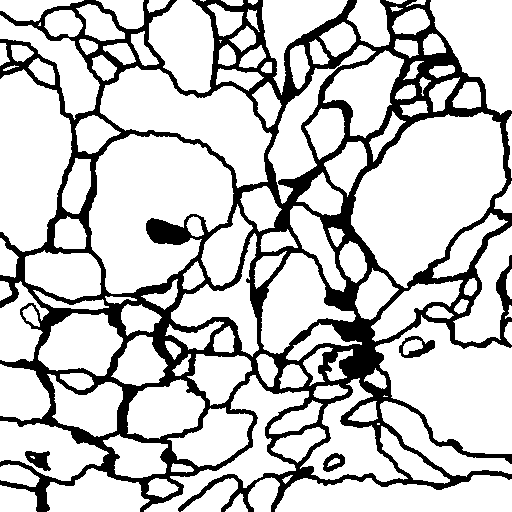}
	\caption{DS2 label image, 512 by 512 pixels of training image 1, corresponding to the raw image in Figure \ref{fig:ds2datasetraw} (Source: ISBI challenge \cite{ISBI2012}, \cite{CaffeNeuralModels}).}
	\label{fig:ds2datasetlabels}
\end{figure}

The data set with the corresponding test and train scripts is available in the Caffe Neural Models repository \cite{CaffeNeuralModels} as \textit{dataset\_02}.
\chapter{Models}
\label{ch:models}

\section{Introduction}
\label{sec:introduction}
This chapter describes how the network architectures were set up for training and processing the data sets DS1 and DS2.
They were configured for two label classification, however also nine labels were tested and even more could be learned through the Softmax loss. With Malis loss (Section \ref{sec:malisloss}), only foreground and background separation is implemented.

For all networks architectures, no special striding or padding was used. Therefore the striding parameter is set to be $s=1$, except for downsampling and sliding window pooling operations, which use $s=k$ (stride matches kernel size). The padding is always $p=0$. All networks use square size filters and feature maps, which simplifies the descriptions to just one value per parameter. The models can be generalized to arbitrary dimensions with different sizes in each dimension. At the time of the project, SK networks up to 6 dimensional can be configured with the modified Caffe library provided \cite{Caffe}.

The mini batch size used for processing in the SW network is $n = 256$, but the choice is arbitrary and only limited by GPU memory. With SK, U and USK networks, a minibatch size of one ($n = 1$) is sufficient to reach 100\% GPU utilization, and often the GPU memory is not sufficient for bigger mini batch sizes (see Section \ref{sec:devicememory}). It is rather useful to increase the network output size than using minibatches, if the data set allows it by having big enough input pictures.

All efficient networks presented here (excluding the sliding window) can be run with almost any size of output prediction maps. Only constraints on even divisibility with pooling operations, as well as the size constraints given by convolutions and strided kernels (see Section \ref{sec:sknet}) have to be met. The networks can therefore be run on different input image sizes, depending on memory requirements (see Section \ref{sec:devicememory}) and data set image size, without re-design and re-training of the networks. The results are numerically identical in this case.

The total padding ($v$) of the networks is a characteristic of the network itself and can not be changed without re-design and re-training. It describes the amount of context considered for each pixel prediction.

To fit the networks to data sets with features of different scales than DS1 and DS2 (see Chapter \ref{ch:datasets}), it may be necessary to adapt kernel sizes and layers to get good predictions after training. Here, the networks are configured so that a big mitochondrion (about 100 by 100 pixels) would fit into the context of a pixel prediction centered on the mitochondrion.

\section{Sliding Window (SW-Net)}
\label{sec:swnet}
Sliding window networks classify an image by taking a pixel and a border padding $v$ of some size around it as input and classify the center pixel by running the patch through a neural network. Then the next pixel is labeled by shifting the window patch by one pixel, classifying the neighboring pixel of the first one. The pixels can also be processed in a minibatch to increase GPU utilization and amortize direct memory access transfer times (from host to device memory). This is still very inefficient as most of the context of two neighboring pixels overlaps and the same filters are applied over the whole context. The redundant computations can be reduced for a patch of input pixels by using SK networks.

The sliding window network described here was developed by Julien Martel \cite{JulienMartel}. It is the baseline for calculating the speedups obtained with the SK, U and USK networks. The structure of the SW network was also used when designing the SK network and the core of the USK network. The reason why this was used as a basis is that it has already been trained on the data set DS1 and had good results.\\
Learning rate, weight decay and training parameters were available. However, no numerical evaluation or publication about the network architecture exists.
\begin{table}[H]
\begin{center}
	\begin{tabular}{|l|l|r|r|r|r|r|}
		\hline
		\textbf{Layer} & \textbf{Type} & $w$ & $f_{\text{in}}$ & $f_{\text{out}}$ & $k$ & $s$\\\hline
		data & MemoryData & 100 & 3 & 3 & 1 & 1\\\hline
		conv1 + relu1 & Convolution + ReLU & 94 & 3 & 48 & 7 & 1\\\hline
		pool1 & Max Pooling & 47 & 48 & 48 & 2 & 2\\\hline
		conv2 + relu2 & Convolution + ReLU & 43 & 48 & 128 & 5 & 1\\\hline
		pool2 & Max Pooling & 22 & 128 & 128 & 2 & 2\\\hline
		conv3 + relu3 & Convolution + ReLU & 20 & 128 & 192 & 3 & 1\\\hline
		pool3 & Max Pooling & 10 & 192 & 192 & 2 & 2\\\hline
		ip1 + relu4 & InnerProduct + ReLU & 1 & 192 & 1024 & 10 & 1\\\hline
		ip2 + relu5 & InnerProduct + ReLU & 1 & 1024 & 512 & 1 & 1\\\hline
		ip3 & InnerProduct & 1 & 512 & 2 & 1 & 1\\\hline
		prob & Softmax & 1 & 2 & 2 & 1 & 1\\\hline
	\end{tabular}
	\captionof{table}{Network setup for the SW model.}
	\label{tab:swnet}
\end{center}
\end{table}
The sliding window network has not been evaluated in-depth in terms of benchmarking and quality assessment. This is because the Caffe Neural Tool (see Chapter \ref{ch:caffeneuraltool}) used for detailed benchmarking and processing does not work with minibatches in its current form. Its segmentation performance should however be numerically equal to the SK network, which is derived from the SW network. Any differences would be due to different training methods (patches on SK versus minibatches on SW).
Information about how single layers speed up from SW to SK networks (both theory and experimental) can be found in the work of Hongsheng Li \etal \cite{2014arXiv1412.4526L}.

For weight initialization, the SW network uses random initialization drawn from a Gaussian distribution with $\mu = 0$ and $\sigma = 0.01$.

\section{SK-Net}
\label{sec:sknet}
\subsection{Converting SW Networks to SK}
\label{sec:swtosk}
Hongsheng Li \etal \cite{2014arXiv1412.4526L} provide a pseudo code (page 4) on how to convert a sliding window network to a strided kernel network. However, it is incomplete on consistency checking, kernel sizes and feature map output sizes. Also, the theory of converting inner product (fully connected) layers is not described. Therefore I provide a more complete version (see Algorithm \ref{alg:swtosk}), although without considering padding and striding. This enforces $s = 1$ and $p = 0$ in all layers of the SK network. Each data dimension (width, height, depth) can be processed separately for the kernel size ($k$), kernel stride ($d$) and output dimension ($w$). The algorithm is able to convert networks and find consistency issues fully automatized.

For Caffe prototxt network configurations, only the kernel size $k$ and kernel stride $d$ have to be provided. Output dimensions will be computed on the fly, given the network input size $w_{\text{SK}}^{(0)}$. Padding and striding parameters can be left away, they will default to the correct values.

\begin{figure}[H]
	\centering
	\begin{tikzpicture}[domain=0:2,label/.style={
		postaction={ decorate,transform shape,
			decoration={ markings, mark=at position .5 with \node #1;}}}]
	
	  \node (sks01) {\includegraphics[scale=0.5]{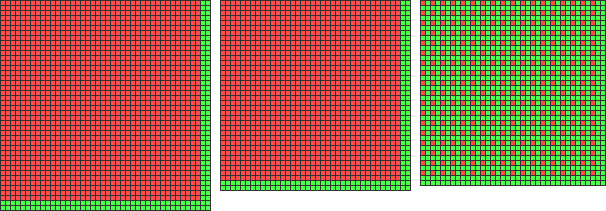}};
	  \node (sks02) [below=1.7cm of sks01] {\includegraphics[scale=0.5]{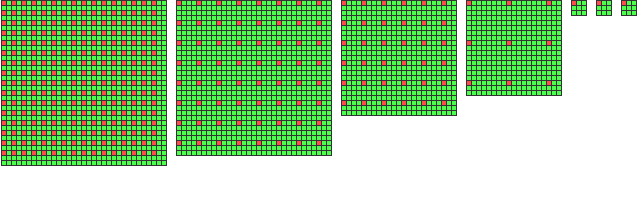}};

	  \draw[decorate,decoration={brace,amplitude=3pt,mirror}] 
		  (-5.35cm,-2.0cm) coordinate () -- (-1.65cm,-2.0cm) coordinate (); 
	  \node at (-3.5cm,-2.5cm) {$w^{(0)}_{\text{SK}}$};
	  
	  \draw[decorate,decoration={brace,amplitude=3pt}] 
	  	  (-5.35cm,2.0cm) coordinate () -- (-1.8cm,2.0cm) coordinate (); 
	  \node at (-3.5cm,2.5cm) {$w^{(0)}_{\text{SW}}$};
	  
	  \draw[decorate,decoration={brace,amplitude=3pt,mirror}] 
		  (-1.45cm,-1.6cm) coordinate () -- (1.9cm,-1.6cm) coordinate (); 
	  \node at (0.225cm,-2.1cm) {$w^{(1,2)}_{\text{SK}}$};

	  \draw[decorate,decoration={brace,amplitude=3pt}] 
		  (-1.45cm,2.0cm) coordinate () -- (1.7cm,2.0cm) coordinate (); 
	  \node at (0.125cm,2.5cm) {$w^{(1,2)}_{\text{SW}}$};
	  
	  \draw[decorate,decoration={brace,amplitude=3pt,mirror}] 
		  (2.1cm,-1.53cm) coordinate () -- (5.35cm,-1.53cm) coordinate (); 
	  \node at (3.725cm,-2.03cm) {$w^{(3)}_{\text{SK}}$};
	  
	  \draw[decorate,decoration={brace,amplitude=3pt,mirror}] 
		  (-5.6cm,-6.9cm) coordinate () -- (-2.7cm,-6.9cm) coordinate (); 
	  \node at (-4.15cm,-7.4cm) {$w^{(4,5)}_{\text{SK}}$};
	  
	  \draw[decorate,decoration={brace,amplitude=3pt,mirror}] 
		  (-2.5cm,-6.7cm) coordinate () -- (0.25cm,-6.7cm) coordinate (); 
	  \node at (-1.125cm,-7.2cm) {$w^{(6)}_{\text{SK}}$};
	  
	  \draw[decorate,decoration={brace,amplitude=3pt,mirror}] 
		  (0.40cm,-6.0cm) coordinate () -- (2.45cm,-6.0cm) coordinate (); 
	  \node at (1.425cm,-6.5cm) {$w^{(7,8)}_{\text{SK}}$};

	  \draw[decorate,decoration={brace,amplitude=3pt,mirror}] 
	  (2.6cm,-5.65cm) coordinate () -- (4.3cm,-5.65cm) coordinate (); 
	  \node at (3.45cm,-6.15cm) {$w^{(9)}_{\text{SK}}$};
	  	  
	  \draw[decorate,decoration={brace,amplitude=3pt}] 
	  	  (3.5cm,2.0cm) coordinate () -- (3.65cm,2.0cm) coordinate (); 
	  \node at (3.575cm,2.4cm) {$d=2$};
	  
	  \draw[decorate,decoration={brace,amplitude=3pt}] 
		  (-4.55cm,-3.65cm) coordinate () -- (-4.40cm,-3.65cm) coordinate (); 
	  \node at (-4.475cm,-3.25cm) {$d=2$};
	  
	  \draw[decorate,decoration={brace,amplitude=3pt}] 
		  (-1.45cm,-3.65cm) coordinate () -- (-1.10cm,-3.65cm) coordinate (); 
	  \node at (-1.275cm,-3.25cm) {$d=4$};
	  
	  \draw[decorate,decoration={brace,amplitude=3pt}] 
		  (1.50cm,-3.65cm) coordinate () -- (1.85cm,-3.65cm) coordinate (); 
	  \node at (1.675cm,-3.25cm) {$d=4$};
	  
	  \draw[decorate,decoration={brace,amplitude=3pt}] 
		  (3.3cm,-3.65cm) coordinate () -- (4.0cm,-3.65cm) coordinate (); 
	  \node at (3.65cm,-3.25cm) {$d=8$};
	  
	  \draw [->, line width=1.25pt, label={[above]{conv1}}] (-1.9cm,2.3cm) coordinate () [bend left=45] to (-1.2cm,2.3cm) coordinate ();
	  
	  \draw [->, line width=1.25pt, label={[above]{pool1}}] (1.6cm,2.3cm) coordinate () [bend left=45] to (2.3cm,2.3cm) coordinate ();
	  
	  \draw [->, line width=1.25pt, label={[above]{conv2}}] (5.1cm,2.3cm) coordinate () [bend left=45] to (5.8cm,2.3cm) coordinate ();
	 
	  \draw [->, line width=1.25pt, label={[above]{pool2}}] (-2.9cm,-3.4cm) coordinate () [bend left=45] to (-2.2cm,-3.4cm) coordinate ();
	  
	  \draw [->, line width=1.25pt, label={[above]{conv3}}] (0.1cm,-3.4cm) coordinate () [bend left=45] to (0.8cm,-3.4cm) coordinate ();

	  \draw [->, line width=1.25pt, label={[above]{pool3}}] (2.3cm,-3.4cm) coordinate () [bend left=45] to (3.0cm,-3.4cm) coordinate ();

	  \draw [->, line width=1.25pt, label={[above]{ip1}}] (4.1cm,-3.4cm) coordinate () [bend left=45] to (4.5cm,-3.4cm) coordinate ();
	  
	  \draw [->, line width=1.25pt, label={[below]{ip2}}] (4.6cm,-4.4cm) coordinate () [bend right=45] to (5.0cm,-4.4cm) coordinate ();
	  
	  \draw [->, bend right=45, line width=1.25pt, label={[above]{ip3}}] (5.1cm,-3.4cm) coordinate () [bend left=45] to (5.5cm,-3.4cm) coordinate ();
	  
	\end{tikzpicture}
	\caption{SK (strided kernel) feature maps. The kernel and feature map sizes used here are smaller than in the actual SK network in order to fit it into a reasonably sized figure. The input size is $w_{\text{SK}}^{(0)}=w_{\text{SW}}^{(0)}+2$ and consequently the output is 3 by 3 pixels as inferred from Algorithm \ref{alg:swtosk}. The red squares trace the information flow in the original SW network. Convolutions and their ReLU activations are not displayed separately.}
	\label{fig:sks}
\end{figure}
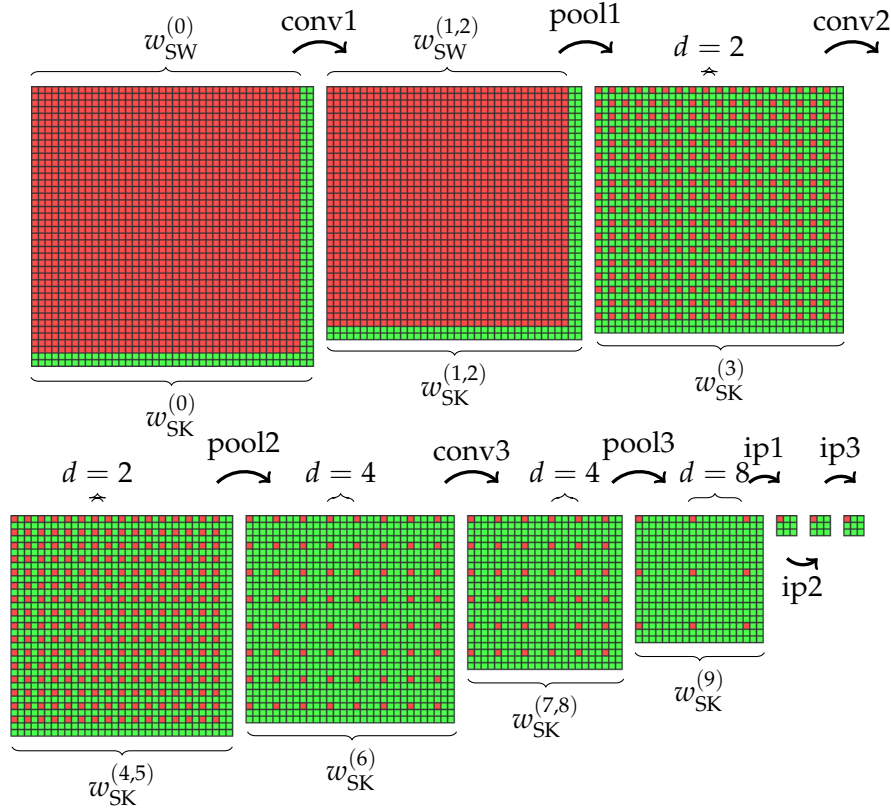

Algorithm \ref{alg:swtosk} assumes that $|L_{\text{SW}}| = |L_{\text{SK}}| = N$, not taking into account the input data layer, which is at $i = 0$.\\
If a layer type is not handled in a special \textit{if}-case, it is handled by using the exact same configuration as in the original network. This fails however if the layer does anything other than an element-wise operation (this implies $k=1$), because of the kernel stride $d > 1$. During the conversion, the initial input size $w^{(0)}_{\text{SK}}$ is equal to what the SW network used. Afterwards, an arbitrary input size $w^{(0)}_{\text{SK}} \geq w^{(0)}_{\text{SW}}$ can be used during training and processing, and the output will be of size $w^{(N)}_{\text{SK}} = w^{(0)}_{\text{SK}} - w^{(0)}_{\text{SW}} + 1$.\\\newline
Using $w^{(0)}_{\text{SW}} = w^{(0)}_{\text{SK}}$ also helps to prove that the results stay numerically the same. In this case, the strides introduced by pooling operations will be implicitly ignored. They will not be taken into account at the first inner product layer (\textit{ip1}), which will span the whole feature map size, because the external kernel size is
\begin{equation}
w^{(i-1)}_{\text{SK}} = (k^{(i)}_{\text{SK}} - 1) d^{(i)}_{\text{SK}} + 1 \Longrightarrow w^{(i)}_{\text{SK}} = 1
\end{equation}
and
\begin{equation}
w^{(i-1)}_{\text{SW}} = k^{(i)}_{\text{SW}} = k^{(i)}_{\text{SK}} \Longrightarrow w^{(i)}_{\text{SW}} = 1
\end{equation}
at that layer. This property can be visualized as well, as in Figure \ref{fig:sks}, where the first inner product layer (\textit{ip1}) has a kernel size of $k_{\text{SK}} = 3 = w_{\text{SW}}^{(9)}$.

For the originally inner product (fully connected) layers in SW-Net, which are now normal convolutions, (\textit{ip1} to \textit{ip3}) there are no computational savings compared to the SW network anymore. This is clear from the observation that the \textit{ip1} layer isolates the context of each pixel and no overlappings in the feature maps exist after this layer.\\
The number of input ($f_{\text{in}}$) and output ($f_{\text{out}}$) feature maps remains exactly the same for SW and SK networks in all layers.\\

\begin{algorithm}[H]
	\caption{Convert SW-Net to SK-Net}\label{euclid}
	\label{alg:swtosk}
	\begin{algorithmic}[1]
		\Procedure{Convert}{}
		\State $\forall i\in [1,N]. s^{(i)}_{\text{SK}} \gets 1$
		\State $\forall i\in [1,N]. p^{(i)}_{\text{SK}} \gets 0$
		\State $w^{(0)}_{\text{SK}} \gets w^{(0)}_{\text{SW}}$
		\State $d_{\text{temp}} \gets 1$
		\For{$i=1$; $i\leq N$; $i \gets i + 1$}
		\If{$l_{\text{SW}}^{(i)} =$ convolution}
		\State $l_{\text{SK}}^{(i)} \gets $ convolution SK
		\State $k^{(i)}_{\text{SK}} \gets k^{(i)}_{\text{SW}}$
		\State $d^{(i)}_{\text{SK}} \gets d_{temp}$
		\State $w^{(i)}_{\text{SK}} \gets w^{(i-1)}_{\text{SK}} - (k^{(i)}_{\text{SK}} - 1) \cdot d^{(i)}_{\text{SK}}$
		\Comment $w^{(i)}_{\text{SW}} \gets w^{(i-1)}_{\text{SW}} - (k^{(i)}_{\text{SW}} - 1)$
		\ElsIf{$l_{\text{SW}}^{(i)}$ = pooling}
		\If{$w^{(i-1)}_{\text{SW}} \mod{k^{(i)}_{\text{SW}}} \neq 0 \vee {k^{(i)}_{\text{SW}}} \neq {s^{(i)}_{\text{SW}}}$} \Return \textit{error}
		\EndIf
		\State $l^{(i)}_{\text{SK}} \gets $ pooling SK
		\State $k^{(i)}_{\text{SK}} \gets k^{(i)}_{\text{SW}}$
		\State $d^{(i)}_{\text{SK}} \gets d_{temp}$
		\State $w^{(i)}_{\text{SK}} \gets w^{(i-1)}_{\text{SK}} - (k^{(i)}_{\text{SK}} - 1) \cdot d^{(i)}_{\text{SK}}$
		\Comment $w^{(i)}_{\text{SW}} \gets \ceil[\bigg]{\frac{w^{(i-1)}_{\text{SW}}}{k^{(i)}_{\text{SW}}}}$
		\State $d_{temp} \gets d_{temp} \cdot k^{(i)}_{\text{SK}}$
		\ElsIf{$l_{\text{SW}}^{(i)}$ = inner product}
		\State $l^{(i)}_{\text{SK}} \gets $ convolution SK
		\State $k^{(i)}_{\text{SK}} \gets w^{(i-1)}_{\text{SW}}$
		\Comment $k^{(i)}_{\text{SW}} = w^{(i-1)}_{\text{SW}}$ is implicit
		\State $d^{(i)}_{\text{SK}} \gets d_{temp}$
		\State $w^{(i)}_{\text{SK}} \gets w^{(i-1)}_{\text{SK}} - (k^{(i)}_{\text{SK}} - 1) \cdot d^{(i)}_{\text{SK}}$
		\Comment $w^{(i)}_{\text{SW}} \gets 1$
		\State $d_{temp} \gets 1$
		\Else
		\If{$k^{(i)}_{\text{SW}} > 1$} \Return \textit{error}
		\EndIf
		\State $l_{\text{SK}} \gets l_{\text{SW}}$
		\State $w^{(i)}_{\text{SK}} \gets w^{(i-1)}_{\text{SK}}$
		\Comment $w^{(i)}_{\text{SW}} \gets w^{(i-1)}_{\text{SW}}$
		\EndIf
		\EndFor
		\If{$d_{temp} = 1$} \Return \textit{success}
		\Else $ $ \Return \textit{error}
		\EndIf
		\EndProcedure
	\end{algorithmic}
\end{algorithm}

Observations on Algorithm \ref{alg:swtosk}:
\begin{itemize}
	\item \textit{Line 13}: Implies only downsamlpling pooling layers can be converted. Pooling with $s=1$ instead of $s=k$ would need to be handled like convolutions in terms of kernel and output size, and would not change the kernel stride $d$. This has not been assessed further.
	\item \textit{Line 17}: Caffe can handle downsampling poolings that overlap the border in SW networks, which causes implicit zero padding. This is not allowed in SK networks, therefore \textit{Line 13} checks if $w \mod k = 0$. Continuing without this check would cause the new strided pooling layer to overlap on data that would normally be separated by a stride, and also output feature maps of wrong sizes to continue.
	\item \textit{Line 18}: Using downsampling is the only operation that increases the kernel stride. All other operations either keep the context local (convolutions) or have $k=1$ (element-wise operations).
	\item \textit{Lines 11, 17, 23}: Interestingly, all converted layer types with kernel sizes now decrease the feature map sizes by the same formula. Convolution and inner product layers implicitly inherit this behavior. Pooling does this because it separates output pixels by a stride equal to the downsampling kernel size $k$.
	\item \textit{Line 29}: Having a kernel stride $d > 1$ in the last layer implies pixels in the output feature maps are not independent from each other. In this case, the network has not been converted correctly and possibly lacks at least one inner product layer.
\end{itemize}

\subsection{SK Network}
\label{sec:sknetwork}
When processing the network in Table \ref{tab:swnet}, the condition $w^{(i-1)}_{\text{SW}} \mod{k^{(i)}_{\text{SW}}} \neq 0$ given by Algorithm \ref{alg:swtosk} is actually violated by the second pooling layer, having $43 \mod{2} \neq 0$. It is easy to fix this by starting at the last layer of the network and computing \begin{equation}
	w^{(i-1)}_{\text{SW}} = k^{(i)}_{\text{SW}}w^{(i)}_{\text{SW}}
\end{equation}
for pooling layers and
\begin{equation}
	w^{(i-1)}_{\text{SW}} = (k^{(i)}_{\text{SW}} - 1) + w^{(i)}_{\text{SW}}
\end{equation} for inner product and convolution layers. The corrected network is given in Table \ref{tab:swnetcorr}. Converting this to SK results in the network in Table \ref{tab:sknet}.

\begin{table}[H]
\begin{center}
	\begin{tabular}{|l|l|r|r|r|r|r|}
		\hline
		\textbf{Layer} & \textbf{Type} & $w$ & $f_{\text{in}}$ & $f_{\text{out}}$ & $k$ & $s$\\\hline
		data & MemoryData & 102 & 3 & 3 & 1 & 1\\\hline
		conv1 + relu1 & Convolution + ReLU & 96 & 3 & 48 & 7 & 1\\\hline
		pool1 & Max Pooling & 48 & 48 & 48 & 2 & 2\\\hline
		conv2 + relu2 & Convolution + ReLU & 44 & 48 & 128 & 5 & 1\\\hline
		pool2 & Max Pooling & 22 & 128 & 128 & 2 & 2\\\hline
		conv3 + relu3 & Convolution + ReLU & 20 & 128 & 192 & 3 & 1\\\hline
		pool3 & Max Pooling & 10 & 192 & 192 & 2 & 2\\\hline
		ip1 + relu4 & InnerProduct + ReLU & 1 & 192 & 1024 & 10 & 1\\\hline
		ip2 + relu5 & InnerProduct + ReLU & 1 & 1024 & 512 & 1 & 1\\\hline
		ip3 & InnerProduct & 1 & 512 & 2 & 1 & 1\\\hline
		prob & Softmax & 1 & 2 & 2 & 1 & 1\\\hline
	\end{tabular}
	\captionof{table}{Network setup for the corrected SW model.}
	\label{tab:swnetcorr}
\end{center}
\end{table}
\begin{table}[H]
\begin{center}
	\begin{tabular}{|l|l|r|r|r|r|r|r|}
		\hline
		\textbf{Layer} & \textbf{Type} & $w$ & $f_{\text{in}}$ & $f_{\text{out}}$ & $k$ & $s$ & $d$\\\hline
		data & MemoryData & 229 & 3 & 3 & 1 & 1 & 1\\\hline
		conv1 + relu1 & Conv. SK + ReLU & 223 & 3 & 48 & 7 & 1 & 1\\\hline
		pool1 & Max Pool. SK & 222 & 48 & 48 & 2 & 1 & 1\\\hline
		conv2 + relu2 & Conv. SK + ReLU & 214 & 48 & 128 & 5 & 1 & 2\\\hline
		pool2 & Max Pool. SK & 212 & 128 & 128 & 2 & 1 & 2\\\hline
		conv3 + relu3 & Conv. SK + ReLU & 204 & 128 & 192 & 3 & 1& 4\\\hline
		pool3 & Max Pool. SK & 200 & 192 & 192 & 2 & 1 & 4\\\hline
		ip1 + relu4 & Conv. SK + ReLU & 128 & 192 & 1024 & 10 & 1 & 8\\\hline
		ip2 + relu5 & Conv. SK + ReLU & 128 & 1024 & 512 & 1 & 1 & 1\\\hline
		ip3 & Conv. SK & 128 & 512 & 2 & 1 & 1 & 1\\\hline
		prob & Softmax & 128 & 2 & 2 & 1 & 1 & 1\\\hline
	\end{tabular}
	\caption{SK network configuration.}
	\label{tab:sknet}
\end{center}
\end{table}
The final three inner product layers from the SW network actually become convolutions with special properties: For \textit{ip1}, the rules stay the same as for converted convolution layers. Afterwards, \textit{ip2} and \textit{ip3} can have an arbitrary kernel stride $d$ because the kernel size $k$ only spans one pixel in each feature map. To not cause confusion, it should be configured so that $d=1$ for those layers.

The network still has one issue, which is not nice but acceptable and the network will still work. The issue is that there is no center pixel, because
\begin{equation}
	(w^{(0)}_{\text{SW}} \hat{=} 102) \mod{2} = 0
\end{equation}
Each patch in the original network has a context of 102 pixels. When converting to SK and classifying a patch of 128 by 128 pixels as given in Table \ref{tab:sknet}, the padding to add in the beginning actually becomes $v = 101$ pixels. This padding can not be split up into a border around the patch to classify. To simplify this issue, a border of 51 pixels on each side is assumed and then cropped by one pixel on the bottom and right side. This results the same behavior as running a 102 by 102 pixel sliding window network across the input, which was also padded with 51 pixels in the corrected version and 50 pixels in the original version.

To estimate the number of free parameters (all convolution weights $|W|$) that can be trained in a model, the following formula is used:
\begin{equation}
	|W| = \sum_{l^{(i)}\in L_{\text{conv.}}}f^{(i)}_{\text{in}}\cdot f^{(i)}_{\text{out}}\cdot (k^{(i)})^2
	\label{eq:freeparams}
\end{equation}
Using the values in Table \ref{tab:sknet}, this gives $|W| \approx 20.5\cdot10^6$ parameters, of which most ($\approx 19.6 \cdot 10^6$) are within the \textit{ip1} layer.

For weight initialization, the SK network uses random initialization drawn from a Gaussian distribution with $\mu = 0$ and $\sigma = 0.01$.

{\tikzset{
	skstride/.style n args={3}{
		path picture={
			\pgfmathparse{(#2*0.002+0.1)}\let\va\pgfmathresult
			\node[anchor=north] at (path picture bounding box.north){
				\includegraphics[height=12cm, width=\va cm]{data/skchunks/d#3.png}
			};}, draw, inner sep=0.0cm, text centered, minimum height=#1*0.02cm, minimum width=(#2*0.002cm+0.05cm)}
}

\definecolor{convcolor}{RGB}{255, 127, 14}
\definecolor{poolcolor}{RGB}{31, 119, 180}
\definecolor{upconvcolor}{RGB}{140, 86, 75}
\definecolor{othercolor}{RGB}{219, 219, 141}

\tikzstyle{convline} = [line width=3.0pt, draw, -latex', color=convcolor]
\tikzstyle{poolline} = [line width=3.0pt, draw, -latex', color=poolcolor]
\tikzstyle{upconvline} = [line width=3.0pt, draw, -latex', color=upconvcolor]
\tikzstyle{otherline} = [line width=3.0pt, draw, -latex', color=othercolor]

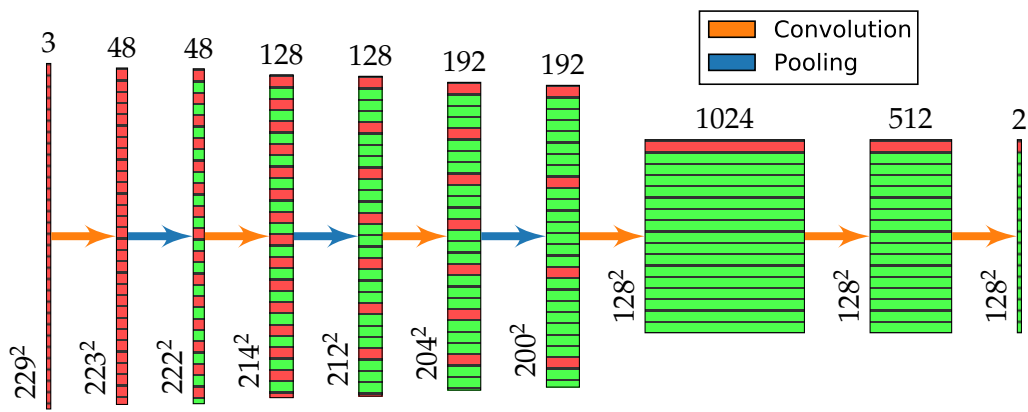
\begin{figure}[H]
	\centering
	\scalebox{1.0}{
	\begin{tikzpicture}[node distance = 1cm and 0.85cm, auto]
	
		\node[skstride={229}{3}{1}](data){};
		\node[rotate=0, anchor=south] at (data.north) {$3$};
		\node[rotate=90, anchor=south west] at (data.south west) {$229^2$};
		
		\node[skstride={223}{48}{1}, right=of data](conv1){};
		\node[rotate=0, anchor=south] at (conv1.north) {$48$};
		\node[rotate=90, anchor=south west] at (conv1.south west) {$223^2$};
		
		\node[skstride={222}{48}{2}, right=of conv1](pool1){};
		\node[rotate=0, anchor=south] at (pool1.north) {$48$};
		\node[rotate=90, anchor=south west] at (pool1.south west) {$222^2$};
		
		\node[skstride={214}{128}{2}, right=of pool1](conv2){};
		\node[rotate=0, anchor=south] at (conv2.north) {$128$};
		\node[rotate=90, anchor=south west] at (conv2.south west) {$214^2$};
		
		\node[skstride={212}{128}{4}, right=of conv2](pool2){};
		\node[rotate=0, anchor=south] at (pool2.north) {$128$};
		\node[rotate=90, anchor=south west] at (pool2.south west) {$212^2$};
		
		\node[skstride={204}{192}{4}, right=of pool2](conv3){};
		\node[rotate=0, anchor=south] at (conv3.north) {$192$};
		\node[rotate=90, anchor=south west] at (conv3.south west) {$204^2$};
		
		\node[skstride={200}{192}{8}, right=of conv3](pool3){};
		\node[rotate=0, anchor=south] at (pool3.north) {$192$};
		\node[rotate=90, anchor=south west] at (pool3.south west) {$200^2$};
		
		\node[skstride={128}{1024}{x}, right=of pool3](ip1){};
		\node[rotate=0, anchor=south] at (ip1.north) {$1024$};
		\node[rotate=90, anchor=south west] at (ip1.south west) {$128^2$};
		
		\node[skstride={128}{512}{x}, right=of ip1](ip2){};
		\node[rotate=0, anchor=south] at (ip2.north) {$512$};
		\node[rotate=90, anchor=south west] at (ip2.south west) {$128^2$};

		\node[skstride={128}{2}{x}, right=of ip2](ip3){};
		\node[rotate=0, anchor=south] at (ip3.north) {$2$};
		\node[rotate=90, anchor=south west] at (ip3.south west) {$128^2$};
		
		\path [convline] (data) -- (conv1);
		\path [poolline] (conv1) -- (pool1);
		\path [convline] (pool1) -- (conv2);
		\path [poolline] (conv2) -- (pool2);
		\path [convline] (pool2) -- (conv3);
		\path [poolline] (conv3) -- (pool3);
		\path [convline] (pool3) -- (ip1);
		\path [convline] (ip1) -- (ip2);
		\path [convline] (ip2) -- (ip3);
		
		\node[] at (10cm,2.5cm) {\includegraphics[scale=0.6]{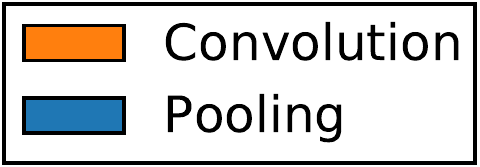}};
	\end{tikzpicture}}
	\caption{SK network configuration visualization. Green-red striped blocks represent feature maps with a kernel stride ($d > 1$). Vertical numbers represent the size of the feature maps while the horizontal numbers represent the number of feature maps.}
	\label{fig:skfmaps}
\end{figure}}

A directed acyclic graph representation of the network can be found in the appendix \ref{app:sknet}.

\section{U-Net}
\label{sec:unet}
The U-Net presented here is the network configuration as described in the Ronneberger \etal paper \cite{2015arXiv150504597R}. Table \ref{tab:unet} describes the network in the same style as Table \ref{tab:sknet} for the SK network in order to compare them. The layer names are chosen in the same style as with SW and SK networks.
The U-Net has contracting and expanding sections:
\begin{itemize}
	\item Contracting: Two convolutions followed by one max pooling layer.
	\item Expanding: Deconvolution followed by a convolution to reduce the number of feature maps, a mergecrop and two convoluton layers.	
\end{itemize}
The source code for running U-Net as well as the prototxt configuration files were not available for download at the time of this project, thus the network presented here, which is my own interpretation, might differ from the original design. The paper does not give all details, such as how the \textit{MergeCrop} layer and \textit{Upconvolution} work. The configurations of this U-Net included in the Caffe Neural Models \cite{CaffeNeuralModels}  are therefore incompatible to the original work \cite{2015arXiv150504597R}. Segmentation results should be comparable.

\begin{center}
	\begin{longtable}{|l|l|r|r|r|r|r|}
		\hline
		\textbf{Layer} & \textbf{Type} & $w$ & $f_{\text{in}}$ & $f_{\text{out}}$ & $k$ & $s$\\\hline
		data & MemoryData & 572 & 3 & 3 & 1 & 1\\\hline
		conv1 + relu1 & Convolution + ReLU & 570 & 3 & 64 & 3 & 1\\\hline
		conv2 + relu2 & Convolution + ReLU & 568 & 64 & 64 & 3 & 1\\\hline
		pool1 & Max Pooling & 284 & 64 & 64 & 2 & 2\\\hline
		conv3 + relu3 & Convolution + ReLU & 282 & 64 & 128 & 3 & 1\\\hline
		conv4 + relu4 & Convolution + ReLU & 280 & 128 & 128 & 3 & 1\\\hline
		pool2 & Max Pooling & 140 & 128 & 128 & 2 & 2\\\hline
		conv5 + relu5 & Convolution + ReLU & 138 & 128 & 256 & 3 & 1\\\hline
		conv6 + relu6 & Convolution + ReLU & 136 & 256 & 256 & 3 & 1\\\hline
		pool3 & Max Pooling & 68 & 256 & 256 & 2 & 2\\\hline
		conv7 + relu7 & Convolution + ReLU & 66 & 256 & 512 & 3 & 1\\\hline
		conv8 + relu8 & Convolution + ReLU & 64 & 512 & 512 & 3 & 1\\\hline
		pool4 & Max Pooling & 32 & 512 & 512 & 2 & 2\\\hline
		conv9 + relu9 & Convolution + ReLU & 30 & 512 & 1024 & 3 & 1\\\hline
		conv10 + relu10 & Convolution + ReLU & 28 & 1024 & 1024 & 3 & 1\\\hline
		upconv1 & Deconvolution & 56 & 1024 & 1024 & 2 & 2\\\hline
		conv11 & Convolution & 56 & 1024 & 512 & 1 & 1\\\hline
		mergecrop1 & MergeCrop & 56 & 512 + 512 & 1024 & 1 & 1\\\hline
		conv12 + relu11 & Convolution + ReLU & 54 & 1024 & 512 & 3 & 1\\\hline
		conv13 + relu12 & Convolution + ReLU & 52 & 512 & 512 & 3 & 1\\\hline
		upconv2 & Deconvolution & 104 & 512 & 512 & 2 & 2\\\hline
		conv14 & Convolution & 104 & 512 & 256 & 1 & 1\\\hline
		mergecrop2 & MergeCrop & 104 & 256 + 256 & 512 & 1 & 1\\\hline
		conv15 + relu13 & Convolution + ReLU & 102 & 512 & 256 & 3 & 1\\\hline
		conv16 + relu14 & Convolution + ReLU & 100 & 256 & 256 & 3 & 1\\\hline
		upconv3 & Deconvolution & 200 & 256 & 256 & 2 & 2\\\hline
		conv17 & Convolution & 200 & 256 & 128 & 1 & 1\\\hline
		mergecrop3 & MergeCrop & 200 & 128 + 128 & 256 & 1 & 1\\\hline
		conv18 + relu15 & Convolution + ReLU & 198 & 256 & 128 & 3 & 1\\\hline
		conv19 + relu16 & Convolution + ReLU & 196 & 128 & 128 & 3 & 1\\\hline
		upconv4 & Deconvolution & 392 & 128 & 128 & 2 & 2\\\hline
		conv20 & Convolution & 392 & 128 & 64 & 1 & 1\\\hline
		mergecrop4 & MergeCrop & 392 & 64 + 64 & 128 & 1 & 1\\\hline
		conv21 + relu17 & Convolution + ReLU & 390 & 128 & 64 & 3 & 1\\\hline
		conv22 + relu18 & Convolution + ReLU & 388 & 64 & 64 & 3 & 1\\\hline
		ip1 & Convolution & 388 & 64 & 2 & 1 & 1\\\hline
		prob & Softmax & 388 & 2 & 2 & 1 & 1\\\hline
		\caption{U network configuration.}
		\label{tab:unet}
	\end{longtable}
\end{center}

The U-Net architecture has $|W| \approx 29\cdot10^6$ parameters, using Equation \ref{eq:freeparams} and the corresponding values in Table \ref{tab:unet}. Thus there are about $30\%$ more parameters than in SK-Net. The number of weights rises towards the middle of the U network, due to having the same convolution kernel size (3 by 3) throughout the network and more feature maps with every contracting step.

For weight initialization, the U network uses random initialization drawn from a Gaussian distribution with $\mu = 0$ and $\sigma = \sqrt{2/(f_{\text{in}}\cdot k^2)}$. This is the same as used in the original paper \cite{2015arXiv150504597R}. With $\sigma = 0.01$ as in SK and SW models, the network could not be trained properly.

The upconvolutions are set to nearest neighbor interpolation, which causes each pixel in the $f_{\text{in}}$ maps to fill exactly four pixels (two by two) in the $f_{\text{out}}$ maps.

{\tikzset{
	skstride/.style n args={3}{
		path picture={
			\pgfmathparse{(#2*0.002+0.1)}\let\va\pgfmathresult
			\node[anchor=north] at (path picture bounding box.north){
				\includegraphics[height=12cm, width=\va cm]{data/skchunks/d#3.png}
			};}, draw, inner sep=0.0cm, text centered, minimum height=#1*0.008cm, minimum width=(#2*0.0005cm+0.05cm)}
}

\definecolor{convcolor}{RGB}{255, 127, 14}
\definecolor{poolcolor}{RGB}{31, 119, 180}
\definecolor{upconvcolor}{RGB}{140, 86, 75}
\definecolor{othercolor}{RGB}{219, 219, 141}

\tikzstyle{convline} = [line width=3.0pt, draw, -latex', color=convcolor]
\tikzstyle{poolline} = [line width=3.0pt, draw, -latex', color=poolcolor]
\tikzstyle{upconvline} = [line width=3.0pt, draw, -latex', color=upconvcolor]
\tikzstyle{otherline} = [line width=3.0pt, draw, -latex', color=othercolor]

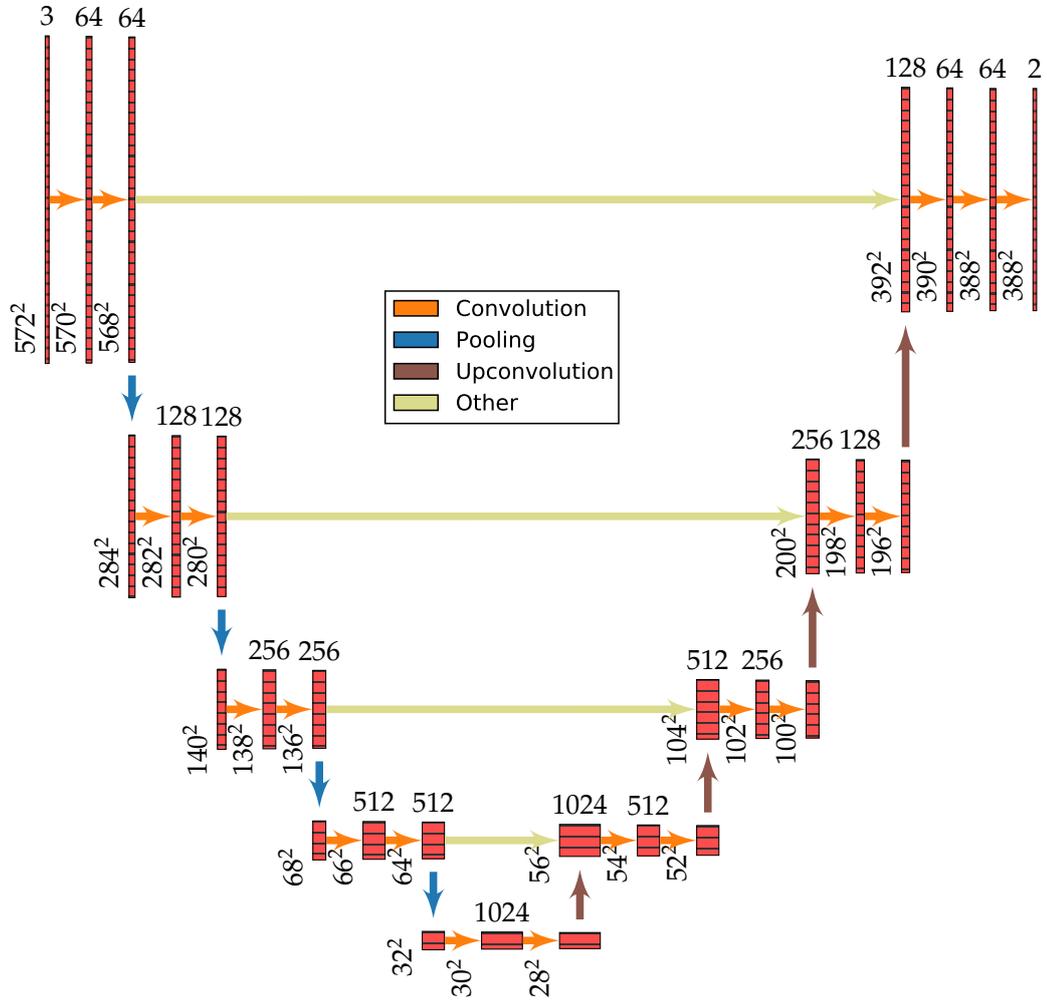
\begin{figure}[H]
	\centering
	\scalebox{0.95}{
	\begin{tikzpicture}[node distance = 1cm and 0.5cm, auto]
	
		\node[skstride={572}{3}{1}](data){};
		\node[rotate=0, anchor=south] at (data.north) {$3$};
		\node[rotate=90, anchor=south west] at (data.south west) {$572^2$};
		
		\node[skstride={570}{64}{1}, right=of data](conv1){};
		\node[rotate=0, anchor=south] at (conv1.north) {$64$};
		\node[rotate=90, anchor=south west] at (conv1.south west) {$570^2$};
		
		\node[skstride={568}{64}{1}, right=of conv1](conv2){};
		\node[rotate=0, anchor=south] at (conv2.north) {$64$};
		\node[rotate=90, anchor=south west] at (conv2.south west) {$568^2$};

		\node[skstride={284}{64}{1}, below=of conv2](pool1){};
		\node[rotate=90, anchor=south west] at (pool1.south west) {$284^2$};
		
		\node[skstride={282}{128}{1}, right=of pool1](conv3){};
		\node[rotate=0, anchor=south] at (conv3.north) {$128$};
		\node[rotate=90, anchor=south west] at (conv3.south west) {$282^2$};

		\node[skstride={280}{128}{1}, right=of conv3](conv4){};
		\node[rotate=0, anchor=south] at (conv4.north) {$128$};
		\node[rotate=90, anchor=south west] at (conv4.south west) {$280^2$};

		\node[skstride={140}{128}{1}, below=of conv4](pool2){};
		\node[rotate=90, anchor=south west, xshift=-0.5cm] at (pool2.south west) {$140^2$};

		\node[skstride={138}{256}{1}, right=of pool2](conv5){};
		\node[rotate=0, anchor=south] at (conv5.north) {$256$};
		\node[rotate=90, anchor=south west, xshift=-0.5cm] at (conv5.south west) {$138^2$};
		
		\node[skstride={136}{256}{1}, right=of conv5](conv6){};
		\node[rotate=0, anchor=south] at (conv6.north) {$256$};
		\node[rotate=90, anchor=south west, xshift=-0.5cm] at (conv6.south west) {$136^2$};
		
		\node[skstride={68}{256}{1}, below=of conv6](pool3){};
		\node[rotate=90, anchor=south west, xshift=-0.5cm] at (pool3.south west) {$68^2$};
		
		\node[skstride={66}{512}{1}, right=of pool3](conv7){};
		\node[rotate=0, anchor=south] at (conv7.north) {$512$};
		\node[rotate=90, anchor=south west, xshift=-0.5cm] at (conv7.south west) {$66^2$};
		
		\node[skstride={64}{512}{1}, right=of conv7](conv8){};
		\node[rotate=0, anchor=south] at (conv8.north) {$512$};
		\node[rotate=90, anchor=south west, xshift=-0.5cm] at (conv8.south west) {$64^2$};
		
		\node[skstride={32}{512}{1}, below=of conv8](pool4){};
		\node[rotate=90, anchor=south west, xshift=-0.5cm] at (pool4.south west) {$32^2$};
		
		\node[skstride={30}{1024}{1}, right=of pool4](conv9){};
		\node[rotate=0, anchor=south] at (conv9.north) {$1024$};
		\node[rotate=90, anchor=south west, xshift=-0.8cm] at (conv9.south west) {$30^2$};
		
		\node[skstride={28}{1024}{1}, right=of conv9](conv10){};
		\node[rotate=90, anchor=south west, xshift=-0.8cm] at (conv10.south west) {$28^2$};

		\node[skstride={56}{1024}{1}, above=of conv10, yshift=0.048cm](upconv1){};
		\node[rotate=0, anchor=south] at (upconv1.north) {$1024$};
		\node[rotate=90, anchor=south west, xshift=-0.5cm] at (upconv1.south west) {$56^2$};

		\node[skstride={54}{512}{1}, right=of upconv1](conv12){};
		\node[rotate=0, anchor=south] at (conv12.north) {$512$};
		\node[rotate=90, anchor=south west, xshift=-0.5cm] at (conv12.south west) {$54^2$};
		
		\node[skstride={52}{512}{1}, right=of conv12](conv13){};
		\node[rotate=90, anchor=south west, xshift=-0.5cm] at (conv13.south west) {$52^2$};

		\node[skstride={104}{512}{1}, above=of conv13, yshift=0.192cm](upconv2){};
		\node[rotate=0, anchor=south] at (upconv2.north) {$512$};
		\node[rotate=90, anchor=south west, xshift=-0.5cm] at (upconv2.south west) {$104^2$};
		
		\node[skstride={102}{256}{1}, right=of upconv2](conv15){};
		\node[rotate=0, anchor=south] at (conv15.north) {$256$};
		\node[rotate=90, anchor=south west, xshift=-0.5cm] at (conv15.south west) {$102^2$};
		
		\node[skstride={100}{256}{1}, right=of conv15](conv16){};
		\node[rotate=90, anchor=south west, xshift=-0.5cm] at (conv16.south west) {$100^2$};
		
		\node[skstride={200}{256}{1}, above=of conv16, yshift=0.48cm](upconv3){};
		\node[rotate=0, anchor=south] at (upconv3.north) {$256$};
		\node[rotate=90, anchor=south west, xshift=-0.2cm] at (upconv3.south west) {$200^2$};
		
		\node[skstride={198}{128}{1}, right=of upconv3](conv18){};
		\node[rotate=0, anchor=south] at (conv18.north) {$128$};
		\node[rotate=90, anchor=south west, xshift=-0.2cm] at (conv18.south west) {$198^2$};
		
		\node[skstride={196}{128}{1}, right=of conv18](conv19){};
		\node[rotate=90, anchor=south west, xshift=-0.2cm] at (conv19.south west) {$196^2$};
		
		\node[skstride={392}{128}{1}, above=of conv19, yshift=1.056cm](upconv4){};
		\node[rotate=0, anchor=south] at (upconv4.north) {$128$};
		\node[rotate=90, anchor=south west] at (upconv4.south west) {$392^2$};
		
		\node[skstride={390}{64}{1}, right=of upconv4](conv21){};
		\node[rotate=0, anchor=south] at (conv21.north) {$64$};
		\node[rotate=90, anchor=south west] at (conv21.south west) {$390^2$};
		
		\node[skstride={388}{64}{1}, right=of conv21](conv22){};
		\node[rotate=0, anchor=south] at (conv22.north) {$64$};
		\node[rotate=90, anchor=south west] at (conv22.south west) {$388^2$};
		
		\node[skstride={388}{2}{1}, right=of conv22](ip1){};
		\node[rotate=0, anchor=south] at (ip1.north) {$2$};
		\node[rotate=90, anchor=south west] at (ip1.south west) {$388^2$};
		
		\path [convline] (data) -- (conv1);
		\path [convline] (conv1) -- (conv2);
		\path [convline] (pool1) -- (conv3);
		\path [convline] (conv3) -- (conv4);
		\path [convline] (pool2) -- (conv5);
		\path [convline] (conv5) -- (conv6);
		\path [convline] (pool3) -- (conv7);
		\path [convline] (conv7) -- (conv8);
		\path [convline] (pool4) -- (conv9);
		\path [convline] (conv9) -- (conv10);
			
		\path [poolline, shorten >=5pt , shorten <=5pt] (conv2) -- (pool1);
		\path [poolline, shorten >=5pt , shorten <=5pt] (conv4) -- (pool2);
		\path [poolline, shorten >=5pt , shorten <=5pt] (conv6) -- (pool3);
		\path [poolline, shorten >=5pt , shorten <=5pt] (conv8) -- (pool4);
		
		\path [otherline] (conv2) -- (upconv4);
		\path [otherline] (conv4) -- (upconv3);
		\path [otherline] (conv6) -- (upconv2);
		\path [otherline] (conv8) -- (upconv1);
		
		\path [upconvline, shorten >=5pt , shorten <=5pt] (conv10) -- (upconv1);
		\path [upconvline, shorten >=5pt , shorten <=5pt] (conv13) -- (upconv2);
		\path [upconvline, shorten >=5pt , shorten <=5pt] (conv16) -- (upconv3);
		\path [upconvline, shorten >=5pt , shorten <=5pt] (conv19) -- (upconv4);
		
		\path [convline] (upconv1) -- (conv12);
		\path [convline] (conv12) -- (conv13);
		\path [convline] (upconv2) -- (conv15);
		\path [convline] (conv15) -- (conv16);
		\path [convline] (upconv3) -- (conv18);
		\path [convline] (conv18) -- (conv19);
		\path [convline] (upconv4) -- (conv21);
		\path [convline] (conv21) -- (conv22);
		\path [convline] (conv22) -- (ip1);
		
		\node[] at (6.3cm,-2.2cm) {\includegraphics[scale=0.6]{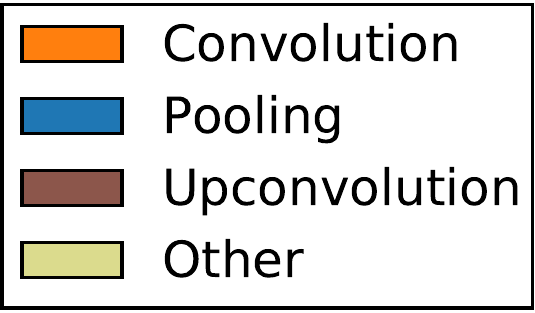}};
	\end{tikzpicture}}
	\caption{U network configuration visualization. Vertical numbers represent the size of the feature maps while the horizontal numbers represent the number of feature maps.}
	\label{fig:ufmaps}
\end{figure}}

Figure \ref{fig:ufmaps} represents the network in a style identical to the one used by Ronneberger \etal \cite{2015arXiv150504597R}.
A directed acyclic graph representation of the network can be found in the appendix \ref{app:unet}.

\section{USK-Net}
The USK network architecture combines ideas from the U and SK models. The majority of the convolutions and therefore free parameters can be trained on downsampled feature maps by using one or more contracting path steps (and their expanding counterpart) from the U-Net.\\
After one contracting step, the same sequence of layers as in the SK network is applied, however with different kernel sizes to match the sizes of the inputs and outputs as required by the U subnetwork.
The network has been configured so that:

\begin{itemize}
	\item The network outputs 512 by 512 pixel labels.
	\item The context considered for each pixel classification is 180 by 180 pixels.
	\item As a result, the network input is 692 by 692 pixels.
	\item The SK network part (\textit{conv3} to \textit{ip2}) is required to accept 344 by 344 pixels as input and outputs 258 by 258 pixels. It therefore sees a context of 86 pixels. This is on once by a factor of two downsampled feature maps. 
\end{itemize}

\begin{figure}[H]
	\centering
	\includegraphics[width=0.9\textwidth]{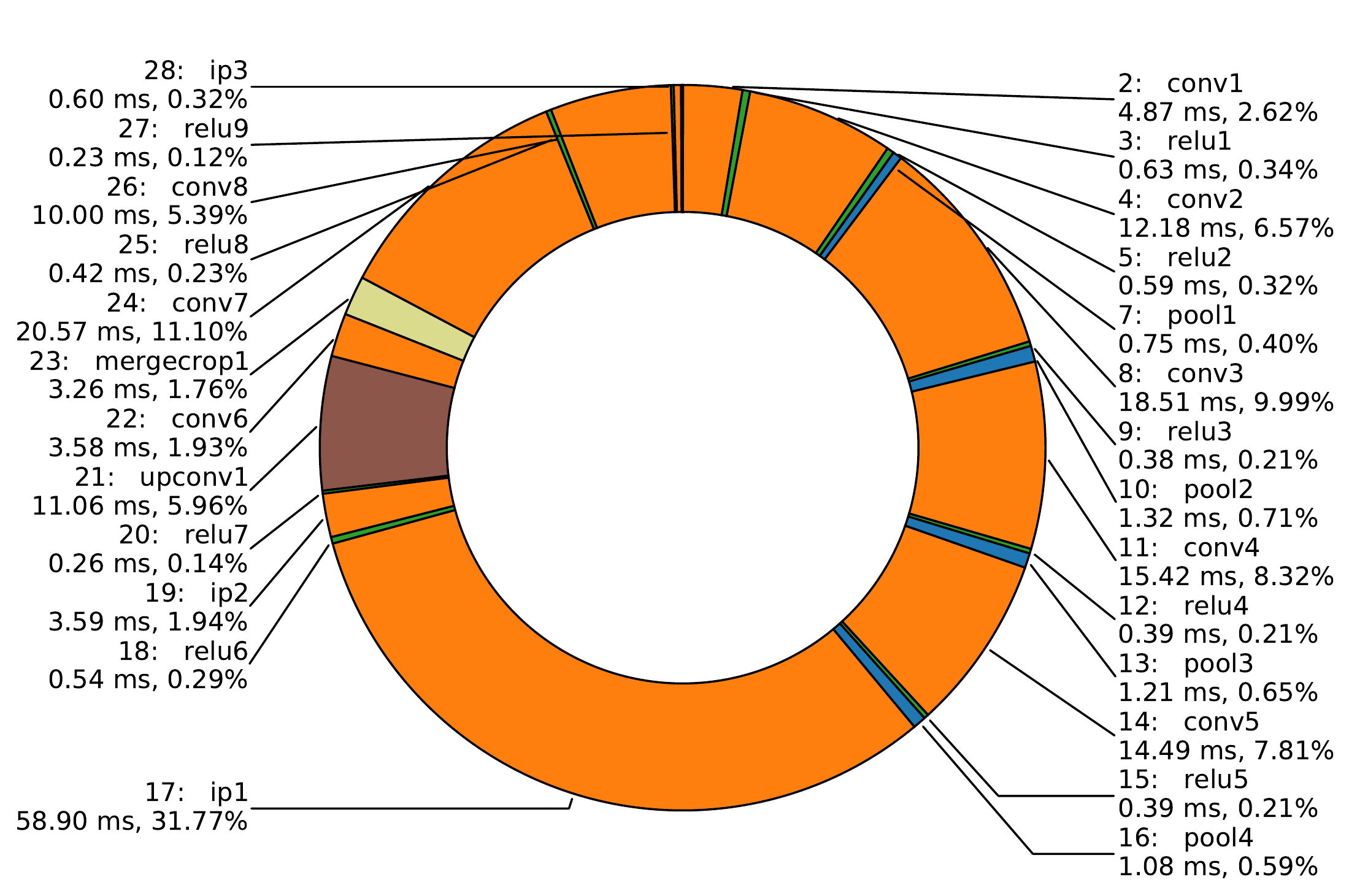}
	\caption{Layer-wise timings of USK-Net, excluding layers that contribute less than \SI{0.1}{\%} of the total forwarding time.}
	\label{fig:uskfwlayerwise}
\end{figure}

\begin{figure}[H]
	\centering
	\includegraphics[width=0.9\textwidth]{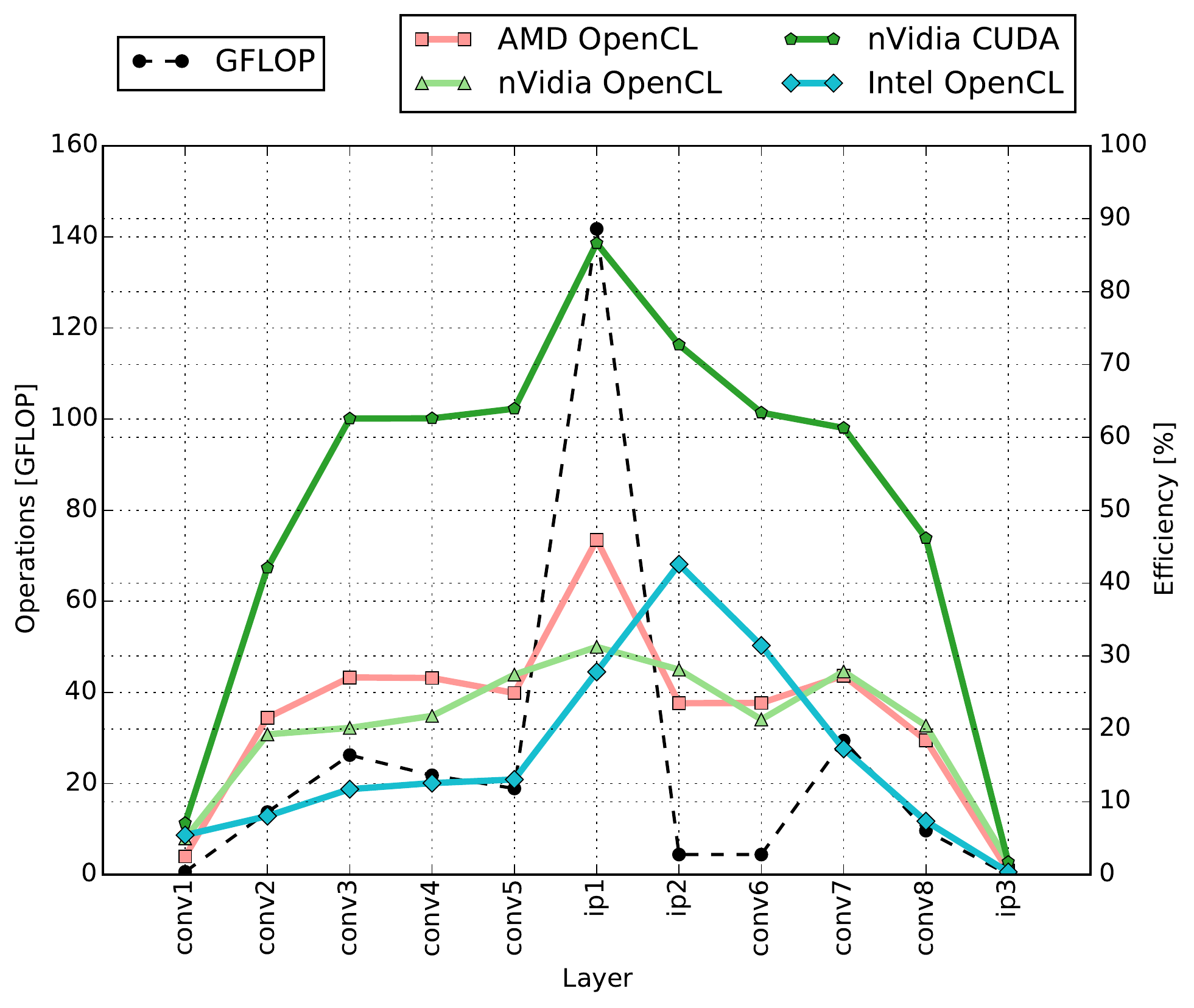}
	\caption{FLOP efficiency of the convolution layers in USK-Net.}
	\label{fig:uskefficiency}
\end{figure}

Prior experiments with two contracting and expanding steps were not very successful, as many features of both datasets (DS1 and DS2) vanished at two times downsampling and the network also became much harder to train because it gets very deep. The network mainly relied on the filter results in the two contracting and expanding step pairs (U subnet) to classify the given input. The signals coming from the SK subnetwork were largely ignored by setting their weights close to zero in the convolution following after the first mergecrop layer.\\
Using only one downsampling step fixed this issue and the SK subnetwork contributed properly after training. As the SK subnetwork contains the main computational costs (see Figures \ref{fig:uskfwlayerwise} and \ref{fig:uskefficiency} in Section \ref{sec:uskbenchmark}) and also carries most parameters, the features have to be meaningful enough at the beginning of the subnetwork, after the first few layers of the U subnetwork.

\label{sec:usknet}
\begin{center}
	\begin{longtable}{|l|l|r|r|r|r|r|r|l|}
		\hline
		\textbf{Layer} & \textbf{Type} & $w$ & $f_{\text{in}}$ & $f_{\text{out}}$ & $k$ & $s$ & $d$ & \textbf{Subnet}\\\hline
		data & MemoryData & 692 & 3 & 3 & 1 & 1 & 1 & U\\\hline
		conv1 + relu1 & Conv. + ReLU & 690 & 3 & 64 & 3 & 1 & 1 & U\\\hline
		conv2 + relu2 & Conv. + ReLU & 688 & 64 & 64 & 3 & 1 & 1 & U\\\hline
		pool1 & Max Pooling & 344 & 64 & 64 & 2 & 2 & 1 & U\\\hline
		conv3 + relu3 & Conv. SK + ReLU & 339 & 64 & 128 & 6 & 1 & 1 & SK\\\hline
		pool2 & Max Pool. SK & 338 & 128 & 128 & 2 & 1 & 1 & SK\\\hline
		conv4 + relu4 & Conv. SK + ReLU & 332& 128 & 128 & 4 & 1 & 2 & SK\\\hline
		pool3 & Max Pool. SK & 330 & 128 & 128 & 2 & 1 & 2 & SK\\\hline
		conv5 + relu5 & Conv. SK + ReLU & 318 & 128 & 128 & 4 & 1 & 4 & SK\\\hline
		pool4 & Max Pool. SK & 314 & 128 & 128 & 2 & 1 & 4 & SK\\\hline
		ip1 + relu6 & Conv. SK + ReLU & 258 & 128 & 512 & 8 & 1 & 8 & SK\\\hline
		ip2 + relu7 & Conv. SK + ReLU & 258 & 512 & 256 & 1 & 1 & 1 & SK\\\hline
		upconv1 & Deconv. & 516 & 256 & 256 & 2 & 2 & 1 & U\\\hline
		conv6 & Conv. & 516 & 256 & 128 & 1 & 1 & 1 & U\\\hline
		mergecrop1 & MergeCrop & 516 & 128 + 64 & 192 & 1 & 1 & 1 & U\\\hline
		conv7 + relu8 & Conv. + ReLU & 514 & 192 & 128 & 3 & 1 & 1 & U\\\hline
		conv8 + relu9 & Conv. + ReLU & 512 & 128 & 64 & 3 & 1 & 1 & U\\\hline
		ip3 & Conv. & 512 & 64 & 2 & 1 & 1 & 1 & U\\\hline
		prob & Softmax & 512 & 2 & 2 & 1 & 1 & 1 & U\\\hline
		\caption{USK network configuration.}
		\label{tab:usknet}
	\end{longtable}
\end{center}

The USK-Net architecture has $|W| \approx 5.5\cdot10^6$ parameters, using Equation \ref{eq:freeparams} and the corresponding values in Table \ref{tab:usknet}. This is a fraction (about $25\%$) of the SK- and U-Net weights. The savings mainly come from reducing the \textit{ip1} layer, which now only has $\approx 4.2 \cdot 10^6$ weights. While the \textit{ip1} layer is still the most expensive one, the network is more balanced than SK-Net. The inner product layers are less important, because the U subnetwork merges and convolves the feature maps from the beginning of the network together with upsampled signals from the \textit{ip2} layer. Figure \ref{fig:uskfmaps} displays the balanced feature maps on the two processing paths.\\
Furthermore, the USK-Net inherits the advantage of having bigger kernel sizes than only 3 by 3 (U-Net), going up to 8 by 8 in the \textit{ip1} and 6 by 6 in the \textit{conv2} layer.\\
This means the USK-Net can learn features with looking at a bigger context inside the feature maps, while still almost reaching the speed of U-Net (see Section \ref{sec:labelthroughput}) in forward processing.

{\tikzset{
		skstride/.style n args={3}{
			path picture={
				\pgfmathparse{(#2*0.002+0.1)}\let\va\pgfmathresult
				\node[anchor=north] at (path picture bounding box.north){
					\includegraphics[height=12cm, width=\va cm]{data/skchunks/d#3.png}
				};}, draw, inner sep=0.0cm, text centered, minimum height=#1*0.01cm, minimum width=(#2*0.002cm+0.05cm)}
}

\definecolor{convcolor}{RGB}{255, 127, 14}
\definecolor{poolcolor}{RGB}{31, 119, 180}
\definecolor{upconvcolor}{RGB}{140, 86, 75}
\definecolor{othercolor}{RGB}{219, 219, 141}

\tikzstyle{convline} = [line width=3.0pt, draw, -latex', color=convcolor]
\tikzstyle{poolline} = [line width=3.0pt, draw, -latex', color=poolcolor]
\tikzstyle{upconvline} = [line width=3.0pt, draw, -latex', color=upconvcolor]
\tikzstyle{otherline} = [line width=3.0pt, draw, -latex', color=othercolor]

\begin{figure}[H]
	\centering
	\scalebox{1.0}{
		\begin{tikzpicture}[node distance = 1cm and 0.65cm, auto]
		
		\node[skstride={692}{3}{1}](data){};
		\node[rotate=0, anchor=south] at (data.north) {$3$};
		\node[rotate=90, anchor=south west] at (data.south west) {$692^2$};
		
		\node[skstride={690}{64}{1}, right=of data](conv1){};
		\node[rotate=0, anchor=south] at (conv1.north) {$64$};
		\node[rotate=90, anchor=south west] at (conv1.south west) {$690^2$};
		
		\node[skstride={688}{64}{1}, right=of conv1](conv2){};
		\node[rotate=0, anchor=south] at (conv2.north) {$64$};
		\node[rotate=90, anchor=south west] at (conv2.south west) {$688^2$};
		
		\node[skstride={344}{64}{1}, below=of conv2](pool1){};
		\node[rotate=90, anchor=south west] at (pool1.south west) {$344^2$};
		
		\node[skstride={339}{128}{1}, right=of pool1](conv3){};
		\node[rotate=0, anchor=south] at (conv3.north) {$128$};
		\node[rotate=90, anchor=south west] at (conv3.south west) {$339^2$};
		
		\node[skstride={338}{128}{2}, right=of conv3](pool2){};
		\node[rotate=0, anchor=south] at (pool2.north) {$128$};
		\node[rotate=90, anchor=south west] at (pool2.south west) {$338^2$};
		
		\node[skstride={332}{128}{2}, right=of pool2](conv4){};
		\node[rotate=0, anchor=south] at (conv4.north) {$128$};
		\node[rotate=90, anchor=south west] at (conv4.south west) {$332^2$};
		
		\node[skstride={330}{128}{4}, right=of conv4](pool3){};
		\node[rotate=0, anchor=south] at (pool3.north) {$128$};
		\node[rotate=90, anchor=south west] at (pool3.south west) {$330^2$};
		
		\node[skstride={318}{128}{4}, right=of pool3](conv5){};
		\node[rotate=0, anchor=south] at (conv5.north) {$128$};
		\node[rotate=90, anchor=south west] at (conv5.south west) {$318^2$};
		
		\node[skstride={314}{128}{8}, right=of conv5](pool4){};
		\node[rotate=0, anchor=south] at (pool4.north) {$128$};
		\node[rotate=90, anchor=south west] at (pool4.south west) {$314^2$};
		
		\node[skstride={258}{512}{x}, right=of pool4](ip1){};
		\node[rotate=0, anchor=south] at (ip1.north) {$512$};
		\node[rotate=90, anchor=south west] at (ip1.south west) {$258^2$};
		
		\node[skstride={258}{256}{x}, right=of ip1](ip2){};
		\node[rotate=90, anchor=south west] at (ip2.south west) {$258^2$};
		
		\node[skstride={516}{192}{1}, above=of ip2, yshift=1.29cm](upconv1){};
		\node[rotate=0, anchor=south] at (upconv1.north) {$192$};
		\node[rotate=90, anchor=south west] at (upconv1.south west) {$516^2$};
		
		\node[skstride={514}{128}{1}, right=of upconv1](conv7){};
		\node[rotate=0, anchor=south] at (conv7.north) {$128$};
		\node[rotate=90, anchor=south west] at (conv7.south west) {$514^2$};
		
		\node[skstride={512}{64}{1}, right=of conv7](conv8){};
		\node[rotate=0, anchor=south] at (conv8.north) {$64$};
		\node[rotate=90, anchor=south west] at (conv8.south west) {$512^2$};
		
		\node[skstride={512}{2}{1}, right=of conv8](ip3){};
		\node[rotate=0, anchor=south] at (ip3.north) {$2$};
		\node[rotate=90, anchor=south west] at (ip3.south west) {$512^2$};
		
		\path [convline] (data) -- (conv1);
		\path [convline] (conv1) -- (conv2);
		
		\path [poolline, shorten >=5pt , shorten <=5pt] (conv2) -- (pool1);

		\path [convline] (pool1) -- (conv3);
		\path [poolline] (conv3) -- (pool2);
		\path [convline] (pool2) -- (conv4);
		\path [poolline] (conv4) -- (pool3);
		\path [convline] (pool3) -- (conv5);
		\path [poolline] (conv5) -- (pool4);
		\path [convline] (pool4) -- (ip1);
		\path [convline] (ip1) -- (ip2);

		\path [convline] (upconv1) -- (conv7);	
		\path [convline] (conv7) -- (conv8);
		\path [convline] (conv8) -- (ip3);

		\path [upconvline, shorten >=5pt , shorten <=5pt] (ip2) -- (upconv1);
		\path [otherline] (conv2) -- (upconv1);
		
		\node[] at (6cm,2.0cm) {\includegraphics[scale=0.6]{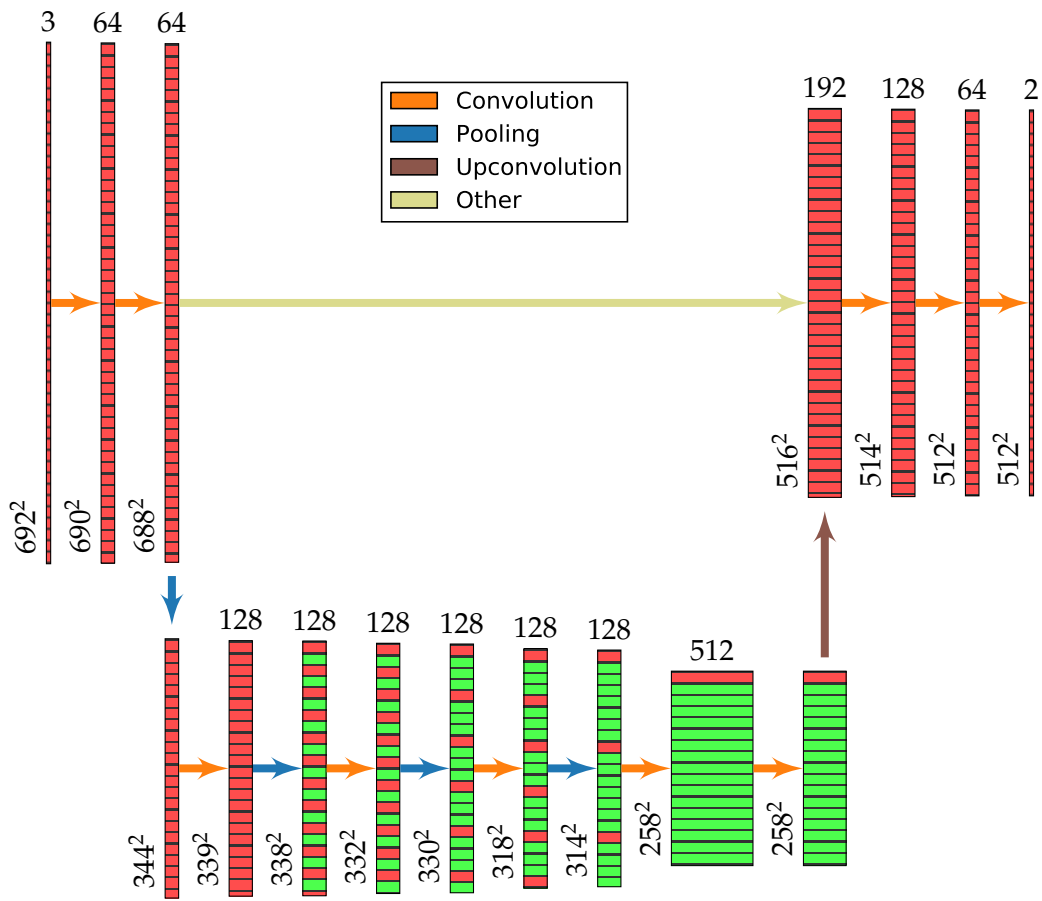}};
	\end{tikzpicture}}
	\caption{USK network configuration visualization. Green-red striped blocks represent feature maps with a kernel stride ($d > 1$). The U subnetwork blocks are just displayed red because the kernel stride does not apply there. Vertical numbers represent the size of the feature maps while the horizontal numbers represent the number of feature maps.}
	\label{fig:uskfmaps}
\end{figure}}

With less weights and less layers (less depth) than the U-Net, it is easier to train and also gave better segmentation results on the two evaluation data sets (see Chapter \ref{ch:results}).

The USK network uses random Gaussian weight initialization with $\mu = 0$ and $\sigma = 0.01$ for the SK subnet and $\sigma = \sqrt{2/(f_{\text{in}}\cdot k^2)}$ for the U subnet. Other initializations, such as $\sigma = 0.01$ for the whole network caused the network to either disable many neurons (causing feature maps with zero activation) or stopped the loss from decreasing early during training.

A directed acyclic graph representation of the network can be found in the appendix \ref{app:usknet}.
\chapter{Caffe Neural Tool}
\label{ch:caffeneuraltool}
\section{Functionality}
As the standard Caffe binary \cite{BVLCCaffe} does not support training with patches, a new interface had to be written on top of the Caffe library. One option is to use the Pycaffe interface with custom python code to load images and pre-process them for patch training and processing as depicted in Figure \ref{fig:pixelpatch}.

I chose to implement a C++ interface similar to the original binary, because OpenCV and OpenMP can be used for efficient, parallelized algorithms to preprocess raw and label images.

The Caffe Neural Tool \cite{CaffeNeuralTool} can be configured for training, processing and benchmarking with a prototxt file similar to the configuration files used to set up networks and solvers (learning configurations) in Caffe. The two most important functionalities are histogram equalization during training and the image preprocessor, which can prepare the label and raw images in various ways before filling up the neural network.

Template configurations for training and processing on the data sets DS1 and DS2 as well as benchmark scripts for U-, SK- and USK-Net are available in the Caffe Neural Models repository \cite{CaffeNeuralModels}.

For training, the following parameters have to be provided:
\begin{itemize}
	\item A Caffe solver prototxt configuration which contains parameters such as learning rate, weight decay, solver method and the network to use for training.
	\item The padding size ($v$), network output (patch) size ($w$), network input ($f_{\text{in}}$) and output ($f_{\text{out}}$) feature map count.
	\item A folder with raw images and a folder with the corresponding label images, which are matched by alphabetic order.
	\item The preprocessor configuration block, including histogram equalization settings.
	\item Optionally, a solverstate file to resume a training. The already learned weights and current learning rate will be loaded instead of the initial network configuration.
\end{itemize}

For processing, the following parameters have to be provided:
\begin{itemize}
	\item A Caffe network prototxt configuration to use for processing.
	\item A caffemodel file, containing the trained network weights.
	\item A folder with raw images and the segmentation output options (file type, pixel format and folder).
	\item The padding size ($v$), network output (patch) size ($w$), network input ($f_{\text{in}}$) and output ($f_{\text{out}}$) feature map count.
	\item The preprocessor configuration block (without histogram equalization).
	\item Optionally, all memory blobs in the network can be stored during processing. This feature is called \textit{filter output} and is useful to check if the network learns the right convolution filters.
\end{itemize}

The tool also supports a variety of input and output image formats: Normal JPEG, PNG, TIF and BMP files as well as TIF image stacks (multiple images in one file) and 32 bit floating point TIF instead of integer pixel values.

When benchmarking, the tool re-uses existing training and processing configurations, but only fills the network with random data. It will report memory usage, layer wise forward and backward times and the total processing time of the network. For convolution layers, it will also estimate and store the computational complexity. This was used to generate the results in Chapter \ref{ch:benchmarks}.

\section{Preprocessing}
The preprocessing options available are:
\begin{itemize}
	\item Label consolidation, allowing to combine multiple labels into one. This technique was used on the data set DS1 (see Section \ref{sec:dataset1}). The consolidation is applied after histogram equalization, allowing to balance out difficult and rare labels before consolidating to only background and foreground. This also allows to mark important small, difficult features in the training data.
	\item Rotation of the training patches to a random multiple of \ang{90}.
	\item Random mirroring of the training patches.
	\item Blurring training patches with a Gaussian kernel of any size. The blurring has a zero mean and the variance is picked at random from a normal distribution. The mean and variance of the distribution can be selected.
	\item CLAHE (contrast limited adaptive histogram equalization), with a clipping parameter. The function is integrated with OpenCV.
	\item Patch normalization to $[-1.0,1.0]$ in floating point before feeding the neural network.
\end{itemize}

There are a few reasons why arbitrary rotation is not available:
\begin{itemize}
	\item Interpolating to any angle that is not a multiple of \ang{90} causes aliasing of training patches (labels and raw images).
	\item Arbitrary rotations make the training patch smaller due to corner cutting. This means there are less available total patches and the computation of histogram equalization methods gets much more complicated. The assumptions about the patch prior and label posterior distributions do not hold anymore.
	\item It is questionable how useful the additional training data would be. Elastic deformations would have more potential in generating unique new training data for biological images such as neural tissue EM images of DS1 and DS2.
\end{itemize}

\section{Histogram Equalization}
\label{sec:histeq}
Histogram equalization is a technique to balance out the frequency of labels that the network sees during training. As training with patches leads to a set of dependent pixels which are in close proximity on an image (see Figure \ref{fig:pixelpatch}), the training results can be worse than with minibatches. With minibatches (see Figure \ref{fig:pixelsliding}), it is possible to create a database of single pixel labels and the minibatch can draw $n$ independent pixels to train with in each stochastic gradient descent step.

The first equalization approach is a patch prior, which will prefer patches with rare labels. This is done by comparing the label distribution within each patch to the total label frequency in all training images.

\begin{equation}
	r_j = \sum_{i=0}^{n-1} \frac{a_{j,i}}{c_i}
	\label{eq:patchweight}
\end{equation}
\begin{equation}
	\hat{c}_i = \frac{1}{Z_i} \sum_{j=0}^{m-1} r_j \cdot a_{j,i}
	\label{eq:labelposterior}
\end{equation}

For patches $j = 0,\ldots,m-1$ and labels $i = 0,\ldots,n-1$, Equation \ref{eq:patchweight} calculates the weight for each patch ($r_j$) based on the total label distribution $c_i$ and the frequency within each patch ($a_{j,i}$).
Equation \ref{eq:labelposterior} calculates the label posterior distribution $\hat{c}_i$ based on the patch weights $r_j$ and the label frequency within the patch. $Z_i$ is a normalization factor to get $\sum_{i=0}^{n-1}\hat{c}_i = 1$.

\begin{figure}[H]
	\center
	\begin{subfigure}{0.5\textwidth}
		\center
		\includegraphics[scale=0.35]{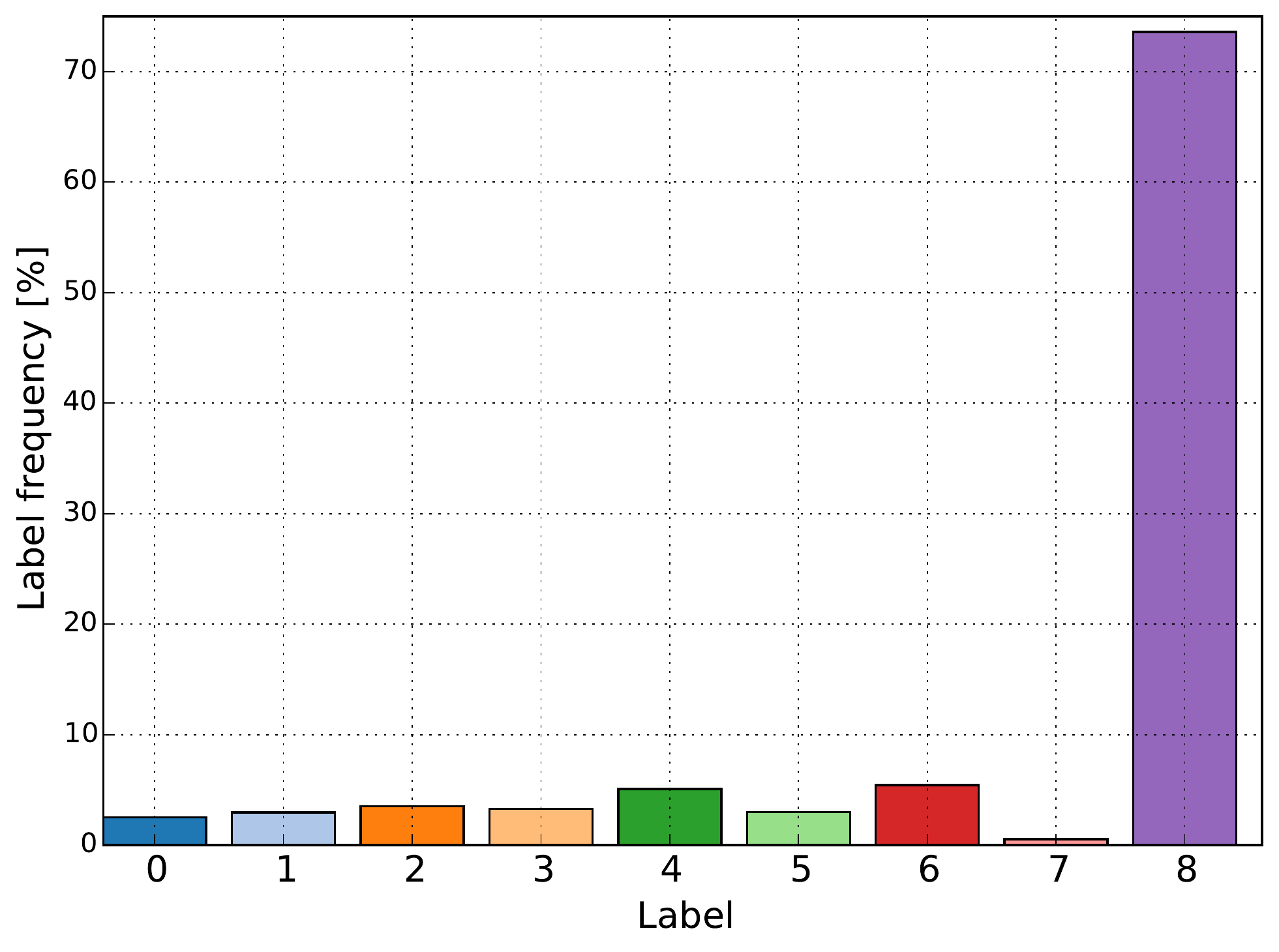}
		\caption{No equalization ($c_i$). The majority of labels is underrepresented.}
	\end{subfigure}%
	\begin{subfigure}{.5\textwidth}
		\center
		\includegraphics[scale=0.35]{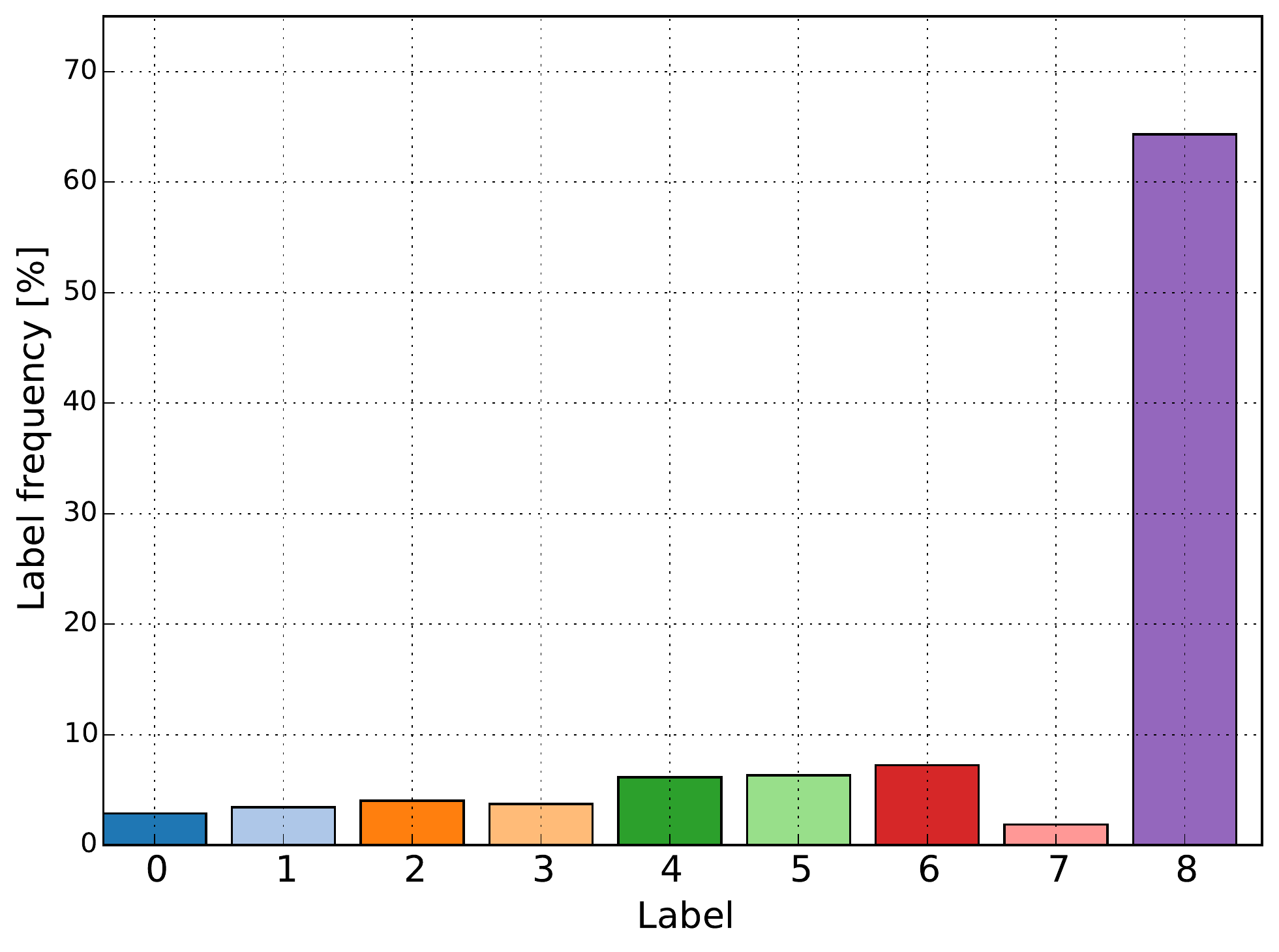}
		\caption{Patch prior ($\hat{c}_i$) on 64 by 64 pixel patches ($w = 64$).}
	\end{subfigure}%
	\caption{Label frequency using the patch prior. Label 8, the cell interior, gets slightly reduced while especially glia cells (5), mitochondria (6) and synapses (7) are represented stronger. Label numbers correspond to DS1 (see Section \ref{sec:dataset1}).}
	\label{fig:histeq}
\end{figure}

The method does help if there are patches that have rare labels, because those patches will be drawn at random with a higher probability than others. An example is the synapse label (number 7) (see Figure \ref{fig:histeq} and Table \ref{tab:histeq}). It cannot balance the labels which have a similar distribution in every patch - for example cell membranes versus cell interior.

When the patch size gets bigger and approaches the size of the training images, the label distribution after the patch prior approaches the original label distribution. Thus the patch prior only works with small training patches. It can also completely equalize the histogram when using a single pixel as patch size ($w = 1$), as this is the same situation as with independent pixels.

When calculating label frequencies $c_i$, it is taken into account that pixel labels closer to the border of the image are covered by less patches than those in the center of the image. A patch can start and end at any offset inside the image. Corner pixels for example are only covered by the one patch that starts in that corner.

After applying the patch prior, the new label distribution $\hat{c}_i$ is depicted in Figure \ref{fig:histeq} with values from Table \ref{tab:histeq}.

\begin{table}[H]
\begin{center}
	\begin{tabular}{|l|r|r|r|r|r|r|r|r|r|}
		\hline
		\textbf{Method} & \textbf{0} & \textbf{1} & \textbf{2} & \textbf{3} & \textbf{4} & \textbf{5} & \textbf{6} & \textbf{7} & \textbf{8}\\\hline
		None ($c_i$) (\%) & 2.5 & 3.0 & 3.5 & 3.3 & 5.1 & 3.0 & 5.5 & 0.6 & 73.6 \\\hline
		Patch prior ($\hat{c}_i$) (\%) & 2.9 & 3.4 & 4.0 & 3.7 & 6.1 & 6.3 & 7.3 & 1.9 & 64.3\\\hline
		Masking (\%) & 11.1 & 11.1 & 11.1 & 11.1 & 11.1 & 11.1 & 11.1 & 11.1 & 11.1\\\hline
	\end{tabular}
	\captionof{table}{Label frequency using different histogram equalization techniques.}
	\label{tab:histeq}
\end{center}
\end{table}

The second method masks pixels randomly in each patch, where the random function is thresholded by the inverse frequency of the label. This means a pixel of label type $i$ gets masked (removed) from the error map if:

\begin{equation}
	c_i^{-1} \cdot (\min_{j} c_j)^{-1} < p
	\label{eq:masking}
\end{equation}

In Equation \ref{eq:masking}, $p \in [0,1]$ is a random value picked at uniform and $c_i \in [0,1]$ the label frequency for label $i = 0,\ldots,n-1$.
Less frequent labels are less likely to be masked out, while membrane and cell interior labels, which are very common, get masked with a high probability. The result is a completely balanced label histogram (see Table \ref{tab:histeq}). The least frequent label consequently never gets masked.

The best solution to train on an exact label distribution as desired remains to do minibatch training. This is theoretically also possible with SK-, USK- and U-Net, but very inefficient in terms of training speed.

Minibatch training is the standard for most networks, also for SW-Net. The network weights are updated every time after a patch or minibatch has run through forwarding and backwarding. The gradients get accumulated over all error pixels in the Softmax loss layer and are then normalized for stable training.

Having a balanced label distribution is mostly important with Softmax and multi label classification. When doing background-foreground separation with Malis, the patch prior has little to no effect and masking can not be used at all. Malis does already focus the error on problematic zones globally over the whole patch that runs through the network (see Section \ref{sec:malisloss}) and therefore the label distribution during training does not matter.
\addtocontents{toc}{\protect\newpage}
\chapter{Caffe Library}
\label{ch:caffelibrary}
\section{Introduction}
Adapting the Caffe library for efficient pixelwise classification on heterogeneous hardware contains the most programming work in the scope of the project, changing 20'000 lines of code compared to the BVLC master branch \cite{Caffe}, \cite{BVLCCaffe}.\\\newline
The changes can be grouped into adaptions on different levels:
\begin{itemize}
	\item Modified solver and network code to support the Caffe Neural Tool C++ interface.
	\item Modified existing layers to fit the pixelwise inputs and outputs.
	\item Additional layers for SK, U and USK architectures.
	\item Additional layers for the Malis loss.
	\item Redesign of N-dimensional layers to support up to 6D convolutions and max pooling with strided kernels.
	\item OpenCL and OpenCL hybrid code for supporting a wide range of GPUs and CPUs.
	\item Backend adaptions to allow dynamic backend dispatching at run- and compile-time.
	\item Adapted GNU Makefile and CMake build infrastructures to support the new OpenCL backend.
\end{itemize}

The changes are implemented in a way that does not break backward compability to the original library. All existing network models and trained networks can still be used. The source code remains highly maintainable and was ahead of the Caffe BVLC branch \cite{BVLCCaffe} during the whole scope of the project.

\section{Modified Layers}

\subsection{SK Layers}
SK (strided kernel) layers are layers with a kernel size $k > 1$ and an inner stride in the kernel ($d > 1$). The result is a kernel that looks at a feature map in a context of $(k - 1)\cdot d + 1$ pixels, which is also called the external kernel size.\\
The motivation to have such kernels is to be able to convert single pixel prediction networks (sliding window (SW) networks), to patch prediction networks (SK). How this works is visible in the Figures \ref{fig:pixelsliding} and \ref{fig:pixelpatch}, as well as the strided representation in Figure \ref{fig:sks}.

For SK convolutions, the matrix multiplication (see Section \ref{sec:convmethods}) stays the same as with normal convolutions. Only the \textit{im2col} and \textit{col2im} memory copy kernels have to be adapted. For this, I used the existing kernel codes provided by Hongsheng Li \etal \cite{2014arXiv1412.4526L}.

Changing existing \textit{im2col} and \textit{col2im} kernels to support $d > 1$ is trivial:\\
The kernels just read the data by iterating over all input feature maps and dimensions. Within each dimension, the iteration goes over the kernel size $k$, copying data into the convolution buffer.\\
With strided kernels, when iterating over the kernel, the reading pointer has to be increased by the dimension stride multiplied by the kernel stride $d$ instead of just adding the dimension stride. The dimension stride is $(w^{(i-1)})^j$, starting at $j=0$ for the first dimension and $w^{(i-1)}$ denoting the input feature map size.

The same iteration scheme applies for pooling operation kernels. All other layers do not have to be adapted and work together with SK layers as long as they have a kernel size of $k = 1$ (see Algorithm \ref{alg:swtosk}).

Strided kernels are only implemented in CUDA and OpenCL and do not run as native CPU code.

\subsection{N-Dimensional Layers}

The Caffe library is able to specify an arbitrary amount of dimensions for the blob memory infrastructure used to pass data between layers. However, not all layers automatically work in higher dimensions. For most element-wise kernels, nothing has to be changed. Convolutions and pooling operations require a slight redesign.

For convolutions, the matrix multiplication stays again the same as with normal and strided kernel convolutions (see Section \ref{sec:convmethods}). 
There were existing kernels for normal N-dimensional convolutions by Jeff Donahue \cite{ndconv}. However, those kernels only support the default kernel stride $d = 1$. 

In the scope of this research project, at HHMI Janelia, I combined the existing code of the strided kernel and N dimensional convolutions to get the most generalized form of convolutions, supporting up to 6 dimensions and kernel strides. The 6 dimension limit exists because allocating local arrays of dynamic sizes is not allowed in OpenCL and CUDA. The arrays are required to store temporary variables such as iterators for each dimension.

Derived from the convolution code, I also implemented the max pooling function as ND-SK kernel.

In the case of 1D or 2D, the SK and normal convolution layers should be considered, as looping over two dimensions is more efficient than having an outer loop over dimensions and an inner loop which processes one dimension at a time. The ND layers should be used from 3D to 6D.

This report does not analyze any networks with a dimension higher than two, but it is possible to replace \textit{ConvolutionSK} layers by \textit{ConvolutionND} layers in the SK-Net to get an arbitrary dimensional network. It may be required to reduce the network output size and feature map count in order to meet memory constraints.

N-dimensional kernels are only implemented in CUDA and OpenCL and do not run as native CPU code.

\section{New Layers}
\subsection{Merge Crop}

The \textit{MergeCrop} layer is required in U- and USK-Net architectures (see Sections \ref{sec:unet} and \ref{sec:usknet}).
The layer accepts two input blobs:
\begin{itemize}
	\item Blob $A$ of size $w^{(i-1)}_A$.
	\item Blob $B$ of size $w^{(i-1)}_B \geq w^{(i-1)}_A$.
\end{itemize}

The layer outputs a blob containing all feature maps of $A$ and $B$, which can have a different amount of feature maps. Input $B$ has to be cropped to the size of $A$. The output feature maps are of size $w^{(i)} = w^{(i-1)}_A$.

During backpropagation, the error maps are propagated through by copying them in the inverse direction. For U- and USK-Net, backwarding is only enabled for the input $A$, as input $B$ gets the differential data on a different path (from the down sampling pooling layer) in the neural network. Copying back $B$ would overwrite the gradients and interfere with the intended training.

\textit{MergeCrop} is only implemented in CUDA and OpenCL and does not run as native CPU code.

\subsection{Malis Loss}
\label{sec:malisloss}

MALIS stands for maximum affinity learning of image segmentation. The implementation in Caffe \cite{Caffe} that I provide is based on existing code to compute the Malis criterion for Matlab \cite{malismatlab} and Torch \cite{malistorch} by Srinivas Turaga \etal \cite{2009arXiv0911.5372T}.

Additional layer forward and backward functions (for interaction with Caffe), memory management and the two additional layers \textit{ConnectedComponent} and \textit{Affinity} are new contributions. The layers are only implemented as CPU code and do not run on OpenCL or CUDA.

\begin{figure}[H]
	\center
	\includegraphics[width=0.9\textwidth]{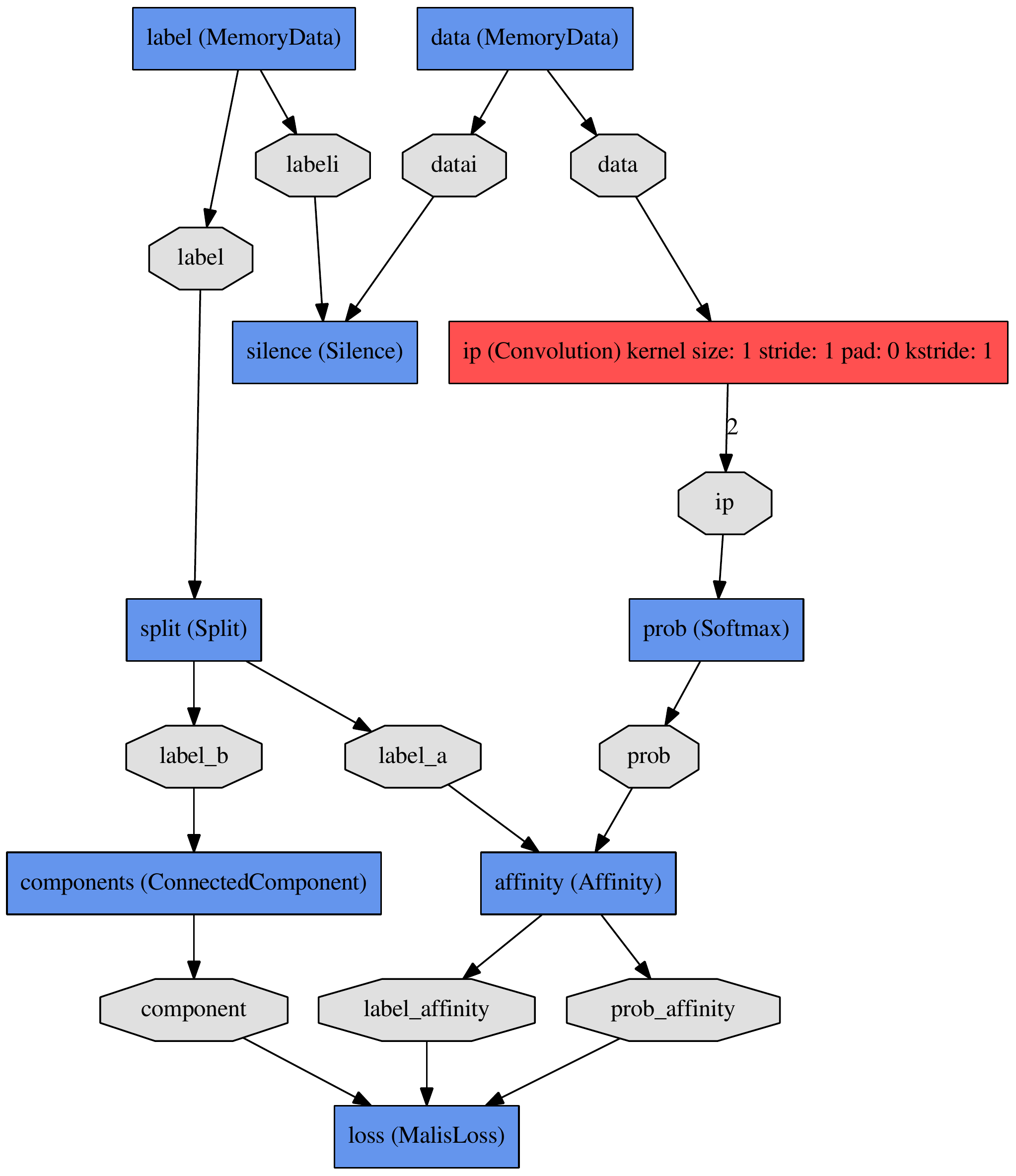}
	\caption{Malis loss setup DAG. In place of the single \textit{ip} convolution layer, there would be a whole neural network (see Appendix \ref{app:net}).}
	\label{fig:malissetup}
\end{figure}

Figure \ref{fig:malissetup} describes how the Malis criterion loss is used together with the network models presented in this report (see Chapter \ref{ch:models}). In place of the Softmax activation for two labels, a rectified linear unit or other activation could also be used. The \textit{split} layer is required to feed the ground truth \textit{label} blob into the two following \textit{components} and \textit{affinity} layers.

The layer structure is separable so that the Malis loss can be used with feeding in external connected component and label affinity maps instead of computing them indirectly. Likewise, the neural network can directly learn affinity graphs instead of pixelwise labels.

As the Malis criterion is calculating minimum spanning trees between pixel pairs on affinity graphs and is only selecting the minimum edge of the tree as error edge, if the position is indeed an error in the prediction, the actual Malis function is called twice on two affinity maps $A^+$ and $A^-$.

Having the predicted affinity graph $A_{\text{pred}}$ (=\textit{prob\_affinity}) and the ground truth to it ($A_\text{label}$ (=\textit{label\_affinity})), two new affinity maps are generated:
\begin{align}
A^+ &= \min(A_{\text{pred}},A_{\text{label}})\label{eq:malisaplus}\\
A^- &= \max(A_{\text{pred}},A_{\text{label}})\label{eq:malisaminus}
\end{align}

Then, $A^+$ will not contain any errors on the background prediction while $A^-$ has no errors in the foreground. The result is that the Malis criterion is able to isolate errors in the background prediction (membranes with gaps) on $A^-$ and find errors on the cell interior with $A^+$. The resulting error map for backpropagation is $\Delta A = \Delta A^- + \Delta A^+$.

In Equations \ref{eq:malisaplus} and \ref{eq:malisaminus} the affinity maps $A$ denote the combination of vertical ($A^{(y)}$) and horizontal ($A^{(x)}$) affinity maps. The affinity graphs contain one value per edge between two neighboring pixels. A value close to 1 stands for high affinity while 0 means unconnected.

Details on how the Malis criterion works internally can be found in the original paper on Malis by Srinivas Turaga \etal \cite{2009arXiv0911.5372T}. The implementation is quite efficient and does not contribute massively to the training time.

\begin{figure}[H]
	\center
	\begin{subfigure}{0.5\textwidth}
		\center
		\includegraphics[scale=0.37]{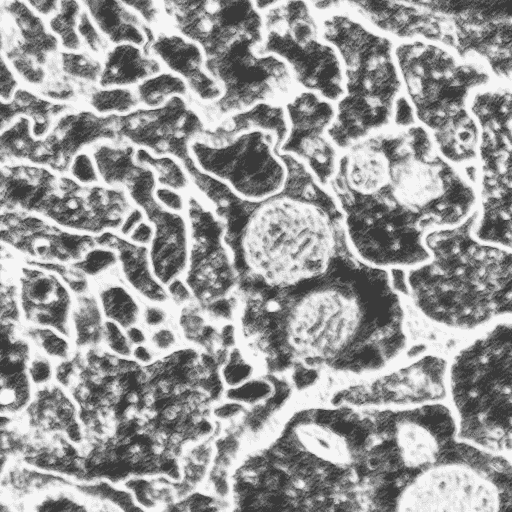}
		\caption{Background prediction map ($I^-$).}
	\end{subfigure}%
	\begin{subfigure}{.5\textwidth}
		\center
		\includegraphics[scale=0.37]{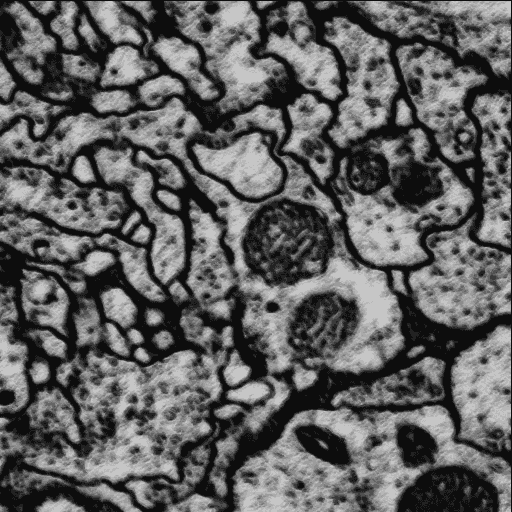}
		\caption{Foreground prediction map ($I^+$).}
	\end{subfigure}%
	\caption{Network predictions after a few training steps with Malis loss (USK-Net, Section \ref{sec:usknet}). The network can not separate membranes and mitochondria yet. The images correspond to the data (forward) component of the memory blob \textit{prob} in Figure \ref{fig:malissetup}.}
	\label{fig:malispred}
\end{figure}

\begin{figure}[H]
	\center
	\begin{subfigure}{0.5\textwidth}
		\center
		\includegraphics[scale=0.37]{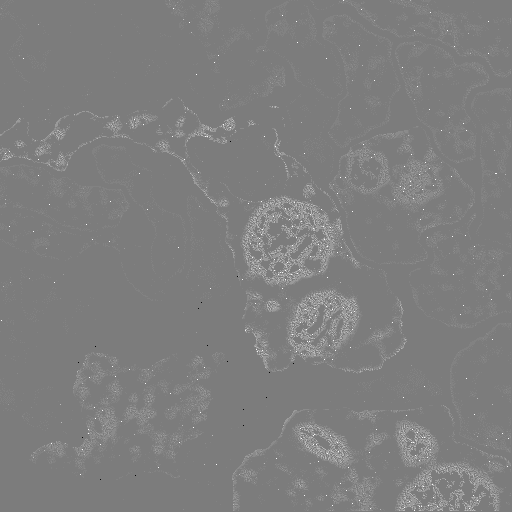}
		\caption{Original error map ($\Delta I^+$). Membrane errors are sparse and barely visible.}
	\end{subfigure}%
	\begin{subfigure}{.5\textwidth}
		\center
		\includegraphics[scale=0.37]{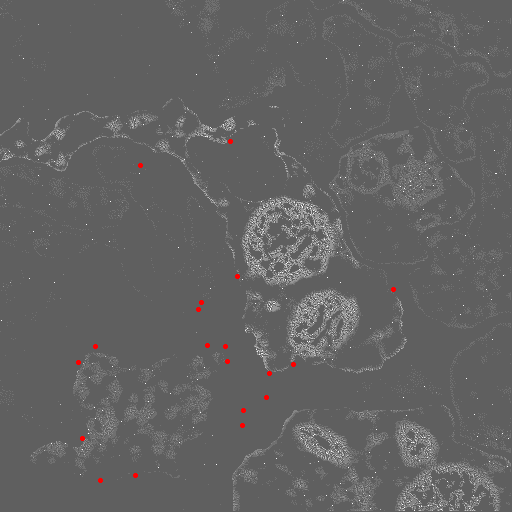}
		\caption{Contrast enhanced error map. Red marks denote membrane errors.}
	\end{subfigure}%
	\caption{The error map generated by the Malis loss. The image corresponds to the diff (backward) component of the memory blob \textit{prob} in Figure \ref{fig:malissetup}. The gray background color is zero (no) error, while black spots are negative errors from $\Delta A^-$ and light gray areas are positive errors from $\Delta A^+$.}
	\label{fig:malisdiff}
\end{figure}

The error map $\Delta I^+$, which is derived from $\Delta A$ using the \textit{affinity} layer during backpropagation, is depicted in Figure \ref{fig:malisdiff}.
On a first look, it seems like only cell interior errors, the light gray areas, exist in the picture.\\
When enhancing the contrast and marking errors in the membrane with red spots, it becomes more clear what happens: The Malis criterion calculates minimum spanning trees between pixels of the cell interior of two merged cells that should be separated. The edge that will be corrected is almost always the same between two cells for all pixel pairs within those cells, due to shared edges in the spanning tree. This edge will therefore accumulate a high error value in $\Delta A^-$, but there are only a few such spots in every training patch.

On the cell interior errors ($\Delta A^+$), the edges to be corrected occur on the border of cell interior areas that are mislabeled as cell membrane. This leads to a denser distribution of the error, but less error intensity per edge.

Malis loss and the U/USK-Net models match perfectly: As the networks can output very large patches up to 512 by 512 pixels without reaching memory limits of current GPUs (see Section \ref{sec:devicememory}), connected components and the Malis loss have a very large context to work on.\\
As soon as the patch size is so small that it barely covers a cell, Malis becomes useless. With sliding window networks for example, it would be necessary to compute many forwarding iterations first before being able to run Malis and generate an error map to backpropagate. It would also be necessary to retain all blobs of all forwarding iterations. The whole process would not be efficient in terms of memory consumption and training times.

\subsection{Affinity}
The affinity layer is an additional layer that has to be used in conjunction with Malis loss (see Figure \ref{fig:malissetup}).

The idea is that not only affinity graphs but also pixelwise classifications can be learned with the Malis loss. During forwarding, the affinity layer has to look at  neighboring pixels in the horizontal and vertical direction of the input and compute their connectivity (affinity).

\begin{align}
A^{(x)}_{x,y} &= \min_{z=x,x+1} I^+_{z,y}\label{eq:affforstart}\\
A^{(y)}_{x,y} &= \min_{z=y,y+1} I^+_{x,z}\\
M^{(x)}_{x,y} &= \argmin_{z=x,x+1} I^+_{z,y}\\
M^{(y)}_{x,y} &= \argmin_{z=y,y+1} I^+_{x,z}\label{eq:affforend}
\end{align}

In Equations \ref{eq:affforstart} to \ref{eq:affforend}, $A^{(x)}$ is the horizontal oriented affinity map, $A^{(y)}$ the vertical one. The affinity maps have the same size as the input image $I$. Additionally, the minimum index maps $M^{(x)}$ and $M^{(y)}$ have to be stored so that the loss can be distribute accordingly during backpropagation.

The Softmax layer (blob \textit{prob}) in Figure \ref{fig:malissetup} and the ground truth labels (blob \textit{label\_a}) actually store both the foreground and background (see Figure \ref{fig:malispred}). For the image $I$ used to produce the affinity map, only the foreground prediction map ($I^+$) is considered because it stores 1 for foreground (connected) and 0 for background (disconnected). The resulting affinity graph will correctly have higher values for connected pixels than disconnected ones.

The affinity graph is computed twice: Once for the ground truth (blob \textit{label\_affinity} in Figure \ref{fig:malissetup}) and once for the current network prediction (blob \textit{prob\_affinity}).

During backpropagation, the affinity loss has to be attributed to single pixels again, as the Malis criterion will attribute the error map to an affinity graph. Now, both the foregound ($\Delta I^+$) and background ($\Delta I^-$) have to get an error map to balance out the Softmax function. The loss is attributed symmetrically, $\Delta I^+ = -\Delta I^-$.
\begin{align}
\intertext{Initialization: } 	\Delta I^+_{(x,y)} = \Delta I^-_{(x,y)}\qquad \phantom{+}&= 0\\
\label{eq:affbackstart}
\intertext{Update: }			\Delta I^+_{M^{(x)}_{x,y},y}\qquad +&= \Delta A^{(x)}_{x,y}\\
\Delta I^-_{M^{(x)}_{x,y},y}\qquad -&= \Delta A^{(x)}_{x,y}\\
\Delta I^+_{x,M^{(y)}_{x,y}}\qquad +&= \Delta A^{(y)}_{x,y}\\
\Delta I^-_{x,M^{(y)}_{x,y}}\qquad -&= \Delta A^{(y)}_{x,y}
\label{eq:affbackend}
\end{align}

Equations \ref{eq:affbackstart} to \ref{eq:affbackend} describe how to attribute the affinity loss back to pixel loss, given the minimum index maps $M^{(x)}$ and $M^{(y)}$ computed in the forward processing step.

There are also other ways to compute an estimation to an affinity graph, such as averaging neighboring pixels. The choice for the minimum selection worked particularly well because there is no loss of resolution or aliasing when computing it this way. The Malis criterion selects the minimum edge of the affinity graph for creating the loss maps $\Delta A^{(x)}$ and $\Delta A^{(y)}$. Thus using the minimum valued pixel through $M^{(x)}$ and $M^{(y)}$ for attributing the pixel loss makes sense. Both objectives minimize the same error by either increasing or decreasing the affinity of the neighboring pixels.

\subsection{Connected Components}

\begin{figure}[H]
	\center
	\includegraphics[scale=0.37]{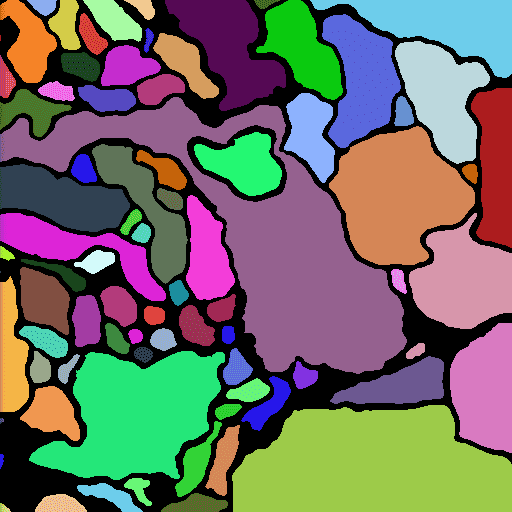}
	\caption{Connected components belonging to Figure \ref{fig:malispred}. The feature map corresponds to the memory blob \textit{component} in Figure \ref{fig:malissetup}.}
	\label{fig:connectedcomponents}
\end{figure}

The connected components layer is a small layer based on the OpenCV flood-filling algorithm and outputs separated connected components from a foreground-background labeled ground truth (Figure \ref{fig:connectedcomponents}). Based on this map, the Malis loss knows which areas have to be separated and which are connected. Here, the cell membrane, which is considered background, is not assigned to any component.

\section{OpenCL Backend}
\label{sec:openclbackend}
\subsection{Implementation}
In my version of the Caffe library, an additional versatile backend for various compute devices, based on OpenCL and ViennaCL \cite{Rupp:ViennaCL}, is available. The backend is called \textit{Greentea} and is part of the \textit{Project Greentea} consisting of frontend, models and modified Caffe library (see Figure \ref{fig:greenteaoverview}). In this section, an overview of interesting aspects how the backend works and how the Caffe library had to be changed is given. Further details and a full documentation is available within the source code, which is available for download \cite{Caffe}.

A key feature is that the OpenCL backend is feature equivalent to the CUDA backend. All GPU layers can be used on both backends. The OpenCL backend is also unit test verified and passes all test cases of the original Caffe library. The tests can be invoked by executing ``make runtest'' on the source code folder.

It remains possible to compile the library with support for all backends at once. The compute kernel and BLAS calls can be dispatched dynamically at runtime, depending on what kind of device is selected. Every device available is registered in a new \textit{DeviceContext} object that stores the device and backend type.

The following aspects of the library had to be changed:
\begin{itemize}
	\item The Caffe library enumerates all devices on all enabled backends, starting with CUDA devices. The selected GPU number determines which \textit{DeviceContext} is set as the default.
	\item The \textit{SyncedMem} class that is used to manage the device memory can now store either a CUDA GPU pointer or an OpenCL \textit{cl\_mem} memory object, depending on which device and backend the memory object belongs to.
	\item The \textit{Forward\_gpu} and \textit{Backward\_gpu} functions now contain both OpenCL and CUDA code to call compute kernels and BLAS functions.
	\item Network layers, \textit{SyncedMem} and blob objects carry a pointer to a \textit{DeviceContext}, which allows to express neural network object to device relationships. This is a feature for allowing future multi-device networks, where keeping track of which memory blob is on which device is essential.
	\item All CUDA compute kernels are translated to OpenCL code. This is trivial for the most part, as the syntax is very similar.
	\item The OpenCL backend can dispatch BLAS calls to ViennaCL-BLAS or clBLAS on GPUs. ViennaCL-BLAS is header-only and therefore easier to use, while clBLAS is optimized for certain AMD GPUs but has to be compiled separately.
	\item On CPU devices, the \textit{Greentea} backend mixes native CPU code with OpenCL code to achieve an optimal performance (see Section \ref{sec:openclhybrid}).
\end{itemize}

\subsection{OpenCL Hybrid}
\label{sec:openclhybrid}
The OpenCL hybrid implementation describes how the OpenCL backend is used when selecting a CPU device instead of a GPU device. The two fundamental differences are memory allocation and BLAS library calls.

When a \textit{SyncedMem} object is instantiated, the memory is allocated as host memory rather than device memory. Differently than with the CPU backend, the memory is allocated through OpenCL. This allows to access the underlying memory pointer of OpenCL memory objects while also being able to use the memory in OpenCL kernels.
\newpage
For BLAS calls, the following steps are executed:
\begin{enumerate}
	\item For all involved \textit{cl\_mem} memory objects, the underlying host pointer is recovered and mapped to a new CPU pointer. At this point, the OpenCL backend ensures all compute kernels accessing the memory concurrently are done executing so that it is safe to access the memory over CPU pointers.
	\item The BLAS call is dispatched to a cBLAS library (Intel MKL, ATLAS or OpenBLAS). Here, the most optimized BLAS for the CPU device can be selected. Optimally, it should be a BLAS that is fully parallelized and uses all CPU cores. NUMA issues (see Section \ref{sec:numaissues}) might occur.
	\item As soon as the BLAS call returns, all CPU pointers are unmapped. This signals to the OpenCL backend that it is safe again to start OpenCL kernels on the \textit{cl\_mem} objects involved.
\end{enumerate}

When mapping \textit{cl\_mem} objects, it can be specified that the access is read-only. In this case, OpenCL kernels that also only use the object in read-only mode can continue to run during the BLAS call.

Using the OpenCL backend on CPUs is a design decision. An alternative would be to parallelize the existing CPU backend with OpenMP pragmas. However, as most of the computational complexity resides with the BLAS calls (see Section \ref{sec:convmethods}) and the OpenCL kernels are not using local memory extensively, they run very well also on CPUs. Only the BLAS, which is very device specific, and needs to be optimized for the memory architecture (see Section \ref{sec:benchsoftware}), needs to be different from the GPU version of the OpenCL backend.

\section{Convolution Methods}
\label{sec:convmethods}

Convolutions are usually computed using three different methods:
\begin{enumerate}
	\item GEMM (matrix multiplication) convolutions, requiring a reshape of the input to fit the BLAS SGEMM scheme.
	\item Direct convolutions, shifting the convolution kernel directly over the input.
	\item FFT domain convolutions, requiring to compute at least two Fourier transforms and one inverse Fourier.
\end{enumerate}

Caffe implements GEMM convolutions. The advantage is that highly efficient BLAS libraries are available specifically for various devices, such as clBLAS, cuBLAS and OpenBLAS. It is very hard to implement convolutions more efficiently using direct convolution. The performance of such implementations is not portable for different kernel sizes and hardware types. In my own preliminary experiments, not even \SI{10}{\%} efficiency could be reached in the \textit{ip1} layer of SK-Net, compared to up to \SI{90}{\%} using GEMM convolution, including the time for input reshaping (see Section \ref{sec:skbenchmark}, Table \ref{tab:skefficiency} and Figure \ref{fig:skefficiency}).

It is particularly difficult to get good local and global memory access patterns when programming kernels for direct convolution. BLAS libraries have already been optimized to use GPU local memory and CPU caches optimally.

GEMM convolutions also simplify the implementation of higher dimension and strided kernel convolutions, as only the code for reshaping the input has to be adapted. In Caffe, these functions are called \textit{im2col} and \textit{col2im}.

A huge disadvantage with GEMM convolution is the memory requirement for the convolution buffer, discussed in Section \ref{sec:devicememory}. Assuming square sized kernels and output images, Equation \ref{eq:convbuffer} gives the buffer size in float elements. The kernel and output dimension is denoted by $x$. The network architectures in this report all use $x = 2$ (2D).
\begin{equation}
M_{\text{buffer}} = f_{\text{in}}\cdot k^x\cdot w^x = K \cdot N
\label{eq:convbuffer}
\end{equation}

A matrix multiplication consists of three matrices, $A \in \R^{M \times K}$, $B \in \R^{K \times N}$ and $C \in \R^{M \times N}$. In Caffe, $A$ is the weight matrix, $B$ the column data after \textit{im2col} and $C = A\cdot B$ the layer output.

The dimensions are $M = f_{\text{out}}$, $N = w^x$ and $K = k^x \cdot f_{\text{in}}$. The resulting computational complexity is $\mathcal{O}(M\cdot N\cdot (2\cdot K - 1)) = \mathcal{O}(f_{\text{out}}\cdot w^x \cdot (2\cdot f_{\text{in}}\cdot k^x-1))$.

{
	
	\tikzstyle{sred} = [rectangle, draw, top color=white, bottom color=red!30, draw=red!50!black!100, drop shadow, minimum width=3cm, text centered, minimum height=3cm]
	\tikzstyle{sgreen} = [rectangle, draw, top color=white, bottom color=green!30, draw=green!50!black!100, drop shadow, minimum width=3cm, text centered, minimum height=3cm]
	\tikzstyle{sblue} = [rectangle, draw, top color=white, bottom color=blue!30, draw=blue!50!black!100, drop shadow, minimum width=3cm, text centered, minimum height=3cm]
	
	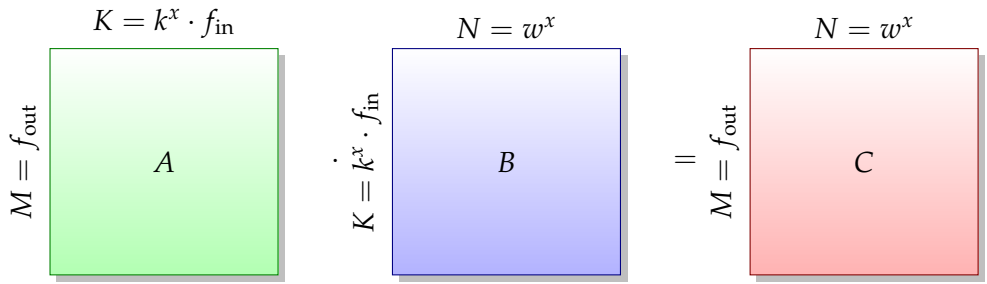
\begin{figure}[H]
		\centering
		\scalebox{1.0}{
			\begin{tikzpicture}[node distance = 0.55cm and 0.55cm, auto]
			\node[sgreen](A){$A$};
			\node[rotate=0, anchor=south] at (A.north) {$K = k^x \cdot f_{\text{in}}$};
			\node[rotate=90, anchor=south] at (A.west) {$M = f_{\text{out}}$};
			
			\node[right=of A](dot){$\cdot$};
			
			\node[sblue, right=of dot](B){$B$};
			\node[rotate=0, anchor=south] at (B.north) {$N = w^x$};
			\node[rotate=90, anchor=south] at (B.west) {$K = k^x \cdot f_{\text{in}}$};
			
			\node[right=of B](eq){$=$};
			
			\node[sred, right=of eq](C){$C$};
			\node[rotate=0, anchor=south] at (C.north) {$N = w^x$};
			\node[rotate=90, anchor=south] at (C.west) {$M = f_{\text{out}}$};
			\end{tikzpicture}}
		\caption{Matrix-matrix multiplication in Caffe.}
		\label{fig:sgemm}
	\end{figure}
}

In the BLAS libraries, row-major NN-SGEMM has to be used. Row-major means the leading dimension in memory is the matrix row. NN means that the matrices A and B are both not transposed. SGEMM stands for single precision general matrix matrix multiplication.

In the scope of this project, FFT convolutions have not been considered. There were no suitable FFT libraries available for all devices that needed to be supported. Normal FFT convolution also uses a lot of additional memory, as all kernels are stretched to the size of the input in Fourier space. The device memory is typically insufficient (see Section \ref{sec:benchhardware}) to store all FFT kernels for reuse, and recomputing every kernel is too time intensive.

The cuDNN library also implements a modified form of GEMM convolutions, and they also evaluated FFT and direct convolution as options \cite{2014arXiv1410.0759C}. A recent paper about using fast FFT convolutions exists (Vasilache \etal \cite{1412.7580}), but managing the device memory remains difficult.
\chapter{Benchmarks}
\label{ch:benchmarks}
\section{Introduction}
This chapter assesses the performance (or efficiency) of different models across a variety of hardware devices. This is essential for speeding up neural networks by finding and eliminating computation and memory bottlenecks.\\
All theoretical computations in this chapter assume two dimensional square sized compute kernels and feature maps. This is in accordance to how the network architectures are set up (see Chapter \ref{ch:models}).

\section{Hardware}
\label{sec:benchhardware}
\begin{table}[H]
\begin{center}
	\centering
	\begin{tabular}{|l|l|l|l|l|}
		\hline
		\textbf{Vendor} &  AMD & nVidia & Intel & Intel\\\hline
		\textbf{Device} & W9100 & GTX 980 & i7-4790K & 2x E5-2697v3\\\hline
		\textbf{Compute Units} & 44 (2816) & 16 (2048) & 4 (8) & 28 (56)\\
		\textbf{(Shaders) / (Threads)} &&&& \\\hline
		\textbf{Memory (GiB)} & 16 & 4 & 16 & \textgreater 16 \\\hline
		\textbf{Clock Frequency (MHz)} & 930 & 1126 & 4000 & 2600 \\\hline
		\textbf{Performance (GFLOP/s)} & 5240.0 & 4612.0 & 512.0 & 2329.6 \\\hline
		\textbf{Memory Speed (GiB/s)} & 320 & 224 & 25.6 & 2x 68 \\\hline
	\end{tabular}
	\captionof{table}{Hardware used for benchmarking.}
	\label{tab:benchhardware}
\end{center}
\end{table}

The device specifications in Table \ref{tab:benchhardware} are taken from the official white papers (AMD \cite{W9100Whitepaper}, nVidia \cite{GTX980Whitepaper}, Intel ARK \cite{intelark}). The FLOP/s performances indicated assume the fused-multiply-add (FMA) operation.

The two AMD W9100 graphics cards have kindly been sponsored by AMD \cite{AMD}, as noted in the acknowledgments. The card has special features such as high 64 bit precision performance and error correcting memory. Those were not used for Caffe, only the OpenCL 2.0 driver and the large amount of memory was of importance. The W9100 is a workstation card, but consumer cards from both nVidia and AMD can also be used without restrictions as long as there is enough device memory (R9 290X, R9 390X, Titan, Titan X). This is to be considered when building low-cost, high-throughput systems for neural networks.

The i7-4790K CPU and GTX 980 GPU are devices of my personal workstation, specially acquired to test with up to date hardware (as of 2015).\\
Extended testing with the E5-2697v3 processors and its issues (see Section \ref{sec:numaissues}) was not possible as this device was a workstation that I could only access briefly during my time at HHMI Janelia.

For all benchmarks in this chapter, the W9100 GPUs have been used, because they were the only available GPUs capable of running all models in forward- and backward-mode on a wide range of output sizes, due to memory requirements. The only exceptions to this are the hardware comparison benchmarks and where indicated explicitly.

Unless otherwise noted, \textit{Intel} is used as an alias for the i7-4790K processor, \textit{nVidia} for the GTX 980 GPU and \textit{AMD} for the W9100 GPU in this chapter.

\section{Software}
\label{sec:benchsoftware}
The modified version of Caffe \cite{Caffe} in \textit{Project Greentea} supports a variety of configurations that perform differently depending on the compute device. Table \ref{tab:benchsoftware} represents the setup used for benchmarking. It is the best performing setup possible for each combination of backend and device for the models in Chapter \ref{ch:models}.

\begin{table}[H]
\begin{center}
	\begin{tabular}{|l|l|l|l|l|}
		\hline
		\textbf{Device} & \textbf{Backend} & \textbf{Memory Allocation} & \textbf{BLAS} & \textbf{Compute Kernels}\\\hline
		Intel & CPU & Host (native) & OpenBLAS & Caffe (CPU native)\\\hline
		Intel & OpenCL & Host (OpenCL) & OpenBLAS & Greentea (OpenCL)\\\hline
		AMD & OpenCL & Device (OpenCL) & clBLAS & Greentea (OpenCL)\\\hline
		nVidia & OpenCL & Device (OpenCL) & clBLAS & Greentea (OpenCL)\\\hline
		nVidia & CUDA & Device (CUDA) & cuBLAS & Caffe (CUDA) \\\hline
	\end{tabular}
	\captionof{table}{Software configuration used for benchmarking.}
	\label{tab:benchsoftware}
\end{center}
\end{table}

OpenBLAS is compiled to use all CPU cores through OpenMP and supports all vector extensions available on the CPUs used. Alternatively, the cBLAS header interface also supports Intel MKL and ATLAS as replacements for OpenBLAS.\\
The CPU could also be used with clBLAS. This is not advisable, as clBLAS is optimized for GPUs, which have a different memory architecture than CPUs. While there is a cache hierarchy on the CPU, the GPU needs to use fast local memory to buffer blocks of the matrix (from global device memory) temporarily. Local memory does not exist on the CPU, therefore it would result in copying data needlessly on the host memory. This results in low efficiency, because CPUs already have a slow memory interface compared to GPUs (see Table \ref{tab:benchhardware}).\\
As ViennaCL \cite{Rupp:ViennaCL} was used as a part of the OpenCL backend, ViennaCL-BLAS is also available as an alternative to clBLAS. It is slower than clBLAS, but more convenient to use as it does not need to be compiled separately and only consists of C++ header files.

\section{Device Memory}

As all networks presented can be run with almost any size of output (up to restrictions given by layers such as even divisibility of the input feature maps, see Chapter \ref{ch:models}), the networks can be set up so that they fit the memory and computational restrictions given by each device. The networks do not have to be re-trained in this case and results are numerically identical.\\
A second objective may be to use processing output sizes matching the dataset: Non-square outputs such as 256 by 32 pixels are also possible. Then, the sizes can be set so that the image sizes to process are divisible by the output size of the network. Like that, no computations are wasted.

It is also important to note that there is no easy scaling rule as to how the memory requirements will change with different output sizes as this depends on the number of feature maps and their sizes on all layers as well as the maximum convolution buffer size ($M_\text{buffer}$).

An upper bound estimation for $w \gg v$ is that the memory usage increases proportional to $w^2$, where $w$ is the output size and $v$ the total padding size as in Table \ref{tab:forwardsetup}.\\
It follows that the network gets more efficient with a bigger ratio $\frac{w}{v}$, removing more overlapping computations. Using $w=1$ on the SK network for example results in having the same efficiency as a minibatch sliding window (SW) network, which is about 50 times slower (see Table \ref{tab:labelthroughput}) than the corresponding SK network with 128 by 128 output.

On 3D networks, the memory allocation would scale in the order of $w^3$, limiting the output patch-cube size very quickly.

\begin{figure}[H]
	\centering
	\includegraphics[width=0.8\textwidth]{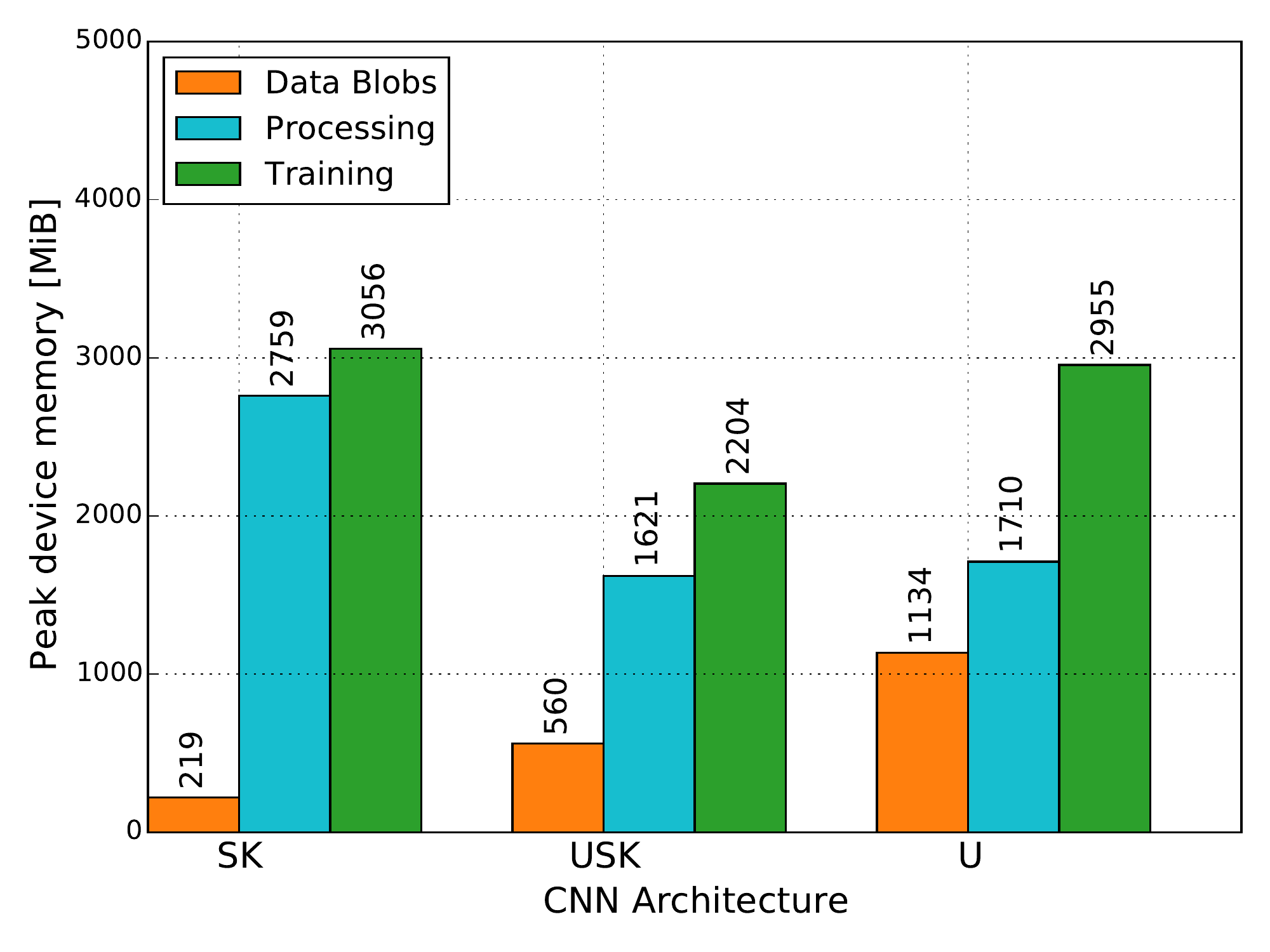}
	\caption{Peak device memory usage when using the processing output sizes denoted in Table \ref{tab:forwardsetup} for both processing and training. The comparison reveals that SK networks would use a lot more memory than USK and U with the same output size, while USK and U are roughly comparable.}
	\label{fig:peakmemory}
\end{figure}

The smallest memory available in the test hardware was \SI{4}{GiB} (see Table \ref{tab:benchhardware}), thus the networks have been set up as in Table \ref{tab:forwardsetup} for forward processing. The configurations are the same as for training and as described in Chapter \ref{ch:models}, except for the USK net, where half the size was used (256 instead of 512). The resulting memory requirements are visible in Figure \ref{fig:peakmemory} and Table \ref{tab:peakmemory}.

Training the USK model with $512 \times 512$ pixels, which improves training with Malis loss, requires up to \SI{8}{GiB} of device memory, which is close to what scaling proportional to $w^2$ predicts.

\begin{table}[H]
	\begin{center}
		\begin{tabular}{|l|r|r|r|}
			\hline
			\textbf{Network} & \textbf{SK} & \textbf{USK} & \textbf{U}\\\hline
			\textbf{Processing output size ($w\times w$)} & $128 \times 128$ & $256 \times 256$ & $388 \times 388$\\\hline
			\textbf{Training output size ($w\times w$)} & $128 \times 128$ & $512 \times 512$ & $388 \times 388$\\\hline
			\textbf{Total padding size ($v$)} & $102$ & $180$ & $184$\\\hline
		\end{tabular}
		\captionof{table}{Output and padding configurations in pixels. Output sizes are mostly flexible, but the padding size is a fixed network characteristic.}
		\label{tab:forwardsetup}
	\end{center}
\end{table}

\begin{table}[H]
	\begin{center}
		\begin{tabular}{|l|r|r|r|}
			\hline
			\textbf{Network} & \textbf{SK} & \textbf{USK} & \textbf{U}\\\hline
			\textbf{Data Blobs (MiB)} & 219 & 560 & 1134 \\\hline
			\textbf{Processing (MiB)} & 2759 & 1621  & 1710 \\\hline
			\textbf{Training (MiB)} & 3056 & 2204  & 2955 \\\hline
		\end{tabular}
		\captionof{table}{Peak device memory usage when using the processing output sizes denoted in Table \ref{tab:forwardsetup} for both processing and training.}
		\label{tab:peakmemory}
	\end{center}
\end{table}

\label{sec:devicememory}

The memory usage during training is always higher than the processing requirements. This is because the error / difference is stored during back propagation. The data blobs therefore have to be stored twice (data and difference) for all layers that are back propagated - making the difference between training and processing roughly the size of the data blobs, as can be seen in Figure \ref{fig:peakmemory}.

With SK and USK models, data, difference and weights are the smallest contributors to memory usage. The biggest consumption comes from the temporary buffer needed to compute convolutions with matrix multiplications. The effect is much smaller with U-Net architectures.\\
To assess this, the size of the buffer can be estimated:
\begin{equation}
	M_{\text{buffer}} = \max_{l^{(i)} \in L_{\text{conv.}}}f_{\text{in}}^{(i)}(k^{(i)})^2(w^{(i)})^2
	\label{eq:maxbuffer}
\end{equation}
The size of $M_{\text{buffer}}$ is in floating point elements. To get an estimate in bytes, it has to be multiplied by the GPUs floating point precision size, which is typically 4 bytes.

The maximum buffer allocation that worked with OpenCL and CUDA was \SI{4}{GiB} for a single memory block. This is the upper limit for $M_{\text{buffer}}$. With the most expensive convolution in the \textit{ip1} layer of SK-Net (see Section \ref{sec:skbenchmark}), it is possible to compute the limits for the network output size $w$, according to Equation \ref{eq:maxbuffer}.\\
$f_{\text{in}} = 192$, $f_{\text{out}} = 1024$ and $k = 10$ are fixed network parameters. Thus, the maximum is $w = \sqrt{\frac{4\, GiB}{4\,B} \cdot (192 \cdot 10^2)^{-1}} \approx 236$ pixels.

For U-Net, this buffer is not the dominant factor that limits the network. In the USK-Net design, the \textit{ip1} layer is of less importance and has less input feature maps and a smaller kernel size. This reduces the $M_{\text{buffer}}$ limitation and allows bigger output patches, making the network more efficient.

With the OpenCL backend (Section \ref{sec:openclbackend}), the memory overhead is up to $\min(n,q)\cdot M_{\text{buffer}}$, where $q$ is the number of parallel work queues and $n$ the minibatch size. This helps to speed up the many small convolutions that occur in SW networks by starting up to $q$ convolutions in parallel. This feature can not be used with the CUDA and CPU backend. For CUDA, the solution in this case is to use cuDNN, which streams the convolutions in batches to be more efficient \cite{2014arXiv1410.0759C}. Both parallel queues and cuDNN are not of importance here, because the SK, U and USK architectures all work efficiently with having $n=1$.

Reusing the convolution buffer is also a new feature in the improved Caffe library \cite{Caffe}. With the original Caffe library \cite{BVLCCaffe}, the memory overhead would have been much higher, allocating one buffer per convolution layer:
\begin{equation}
	M_{\text{buffer}} = \sum_{l^{(i)} \in L_{\text{conv.}}}f_{\text{in}}^{(i)}(k^{(i)})^2(w^{(i)})^2
\end{equation}
Those buffers are in both cases persistent, because freeing them and re-allocating would cause too much run time overhead. With re-using the buffer, it would also not decrease the peak allocation any further. Alternative ways to implement convolutions are discussed in Section \ref{sec:convmethods}.

The memory consumption can not be decreased further during training, because the data blobs have to persist during forward- and backward-computation to calculate the difference and update the weights in training. However, during processing, a lower bound estimate is given by:
\begin{equation}
	M_{\text{total}} = \min(n,q)\cdot M_{\text{buffer}}+n\cdot\max_{b^{(i-1)}\in B,\; b^{(i)}\in B}[f^{(i)}_{\text{in}}(w^{(i-1)})^2+f^{(i)}_{\text{out}}(w^{(i)})^2]
\end{equation}
This assumes that the blobs $B$ of a network are not persistent and at most the input and output of the most memory consuming layer has to be stored. A network with its blobs can be seen as a directed acyclic graph (DAG). Therefore, the estimate is higher if there are layers / blobs that exist in parallel with each other.\\
Currently (as of August 2015) there is no Caffe implementation that re-uses the blobs in this way. It would be necessary to analyze the DAG when instantiating the network first. Then, blobs would have to be allocated and assigned so that no conflicts exist. The DAG would need to be split up for analysis in the case that there are multiple devices with independent memory working on different parts of the network.

The implication is that lower end devices such as GPUs with less memory and even mobile devices would be capable of classifying with larger networks such as those presented in this report.\\
I chose to not implement this because there was enough GPU memory available and Figure \ref{fig:peakmemory} indicates the reduction would not be very large, as most memory is consumed by matrix-matrix multiplication buffer ($M_{\text{buffer}}$).

\section{Labeling Throughput}
\label{sec:labelthroughput}

\begin{table}[H]
\begin{center}
	\begin{tabular}{|l|l|r|r|r|r|}
		\hline
		\textbf{Device} & \textbf{Backend} & \textbf{SW} & \textbf{SK} & \textbf{USK} & \textbf{U}\\\hline
		Intel & CPU & 73 & 0\textsuperscript{(a)} & 0\textsuperscript{(a)} & 0\textsuperscript{(a)}\\\hline
		Intel & OpenCL & 98 & 3\,504 & 19\,414 & 32\,692\\\hline
		nVidia & OpenCL & 850 & 28\,461 & 267\,478 & 390\,156\\\hline
		AMD & OpenCL & 1\,108 & 59\,513 & 354\,865 & 483\,766\\\hline
		nVidia & CUDA & 1\,488 & 85\,460 & 658\,018 & 1\,058\,125\\\hline
	\end{tabular}
	\captionof{table}{Labeling throughput\protect\linebreak\textsuperscript{(a)} not implemented}
	\label{tab:labelthroughput}
\end{center}
\end{table}

The labeling throughput (Table \ref{tab:labelthroughput}) is an overall performance measure for neural networks. It also shows how different devices and backends perform. Even on the CPU, using the fastest network (U) gives a speedup of $447\times$ compared to what was achievable with Caffe \cite{BVLCCaffe} prior to \textit{Project Greentea} \cite{Caffe}. When using a network (SK) that gives identical results as the original SW network, the CPU speedup is still a factor of $48\times$.

On AMD GPUs, speedups of $54\times$ (SK-Net), $437\times$ (U-Net) and $320\times$ (USK-Net) compared to SW-Net are possible.
The nVidia GPU scales similarly on both the OpenCL (SK: $33\times$, U: $459\times$, USK: $315\times$) and CUDA (SK: $57\times$, U: $711\times$, USK: $442\times$) backend.

SK, USK and U networks can not be executed directly on the legacy CPU backend, as layers such as strided kernel layers and merge crop have no native CPU kernels implemented. They are only available on CUDA and OpenCL. The fact that the OpenCL backend on CPUs is better parallelized (see Section \ref{sec:openclhybrid}) than native CPU execution in Caffe and that the speedups between networks are similar on CPUs and GPUs justifies not implementing the CPU kernels. 

\begin{figure}[H]
	\centering
	\includegraphics[width=0.8\textwidth]{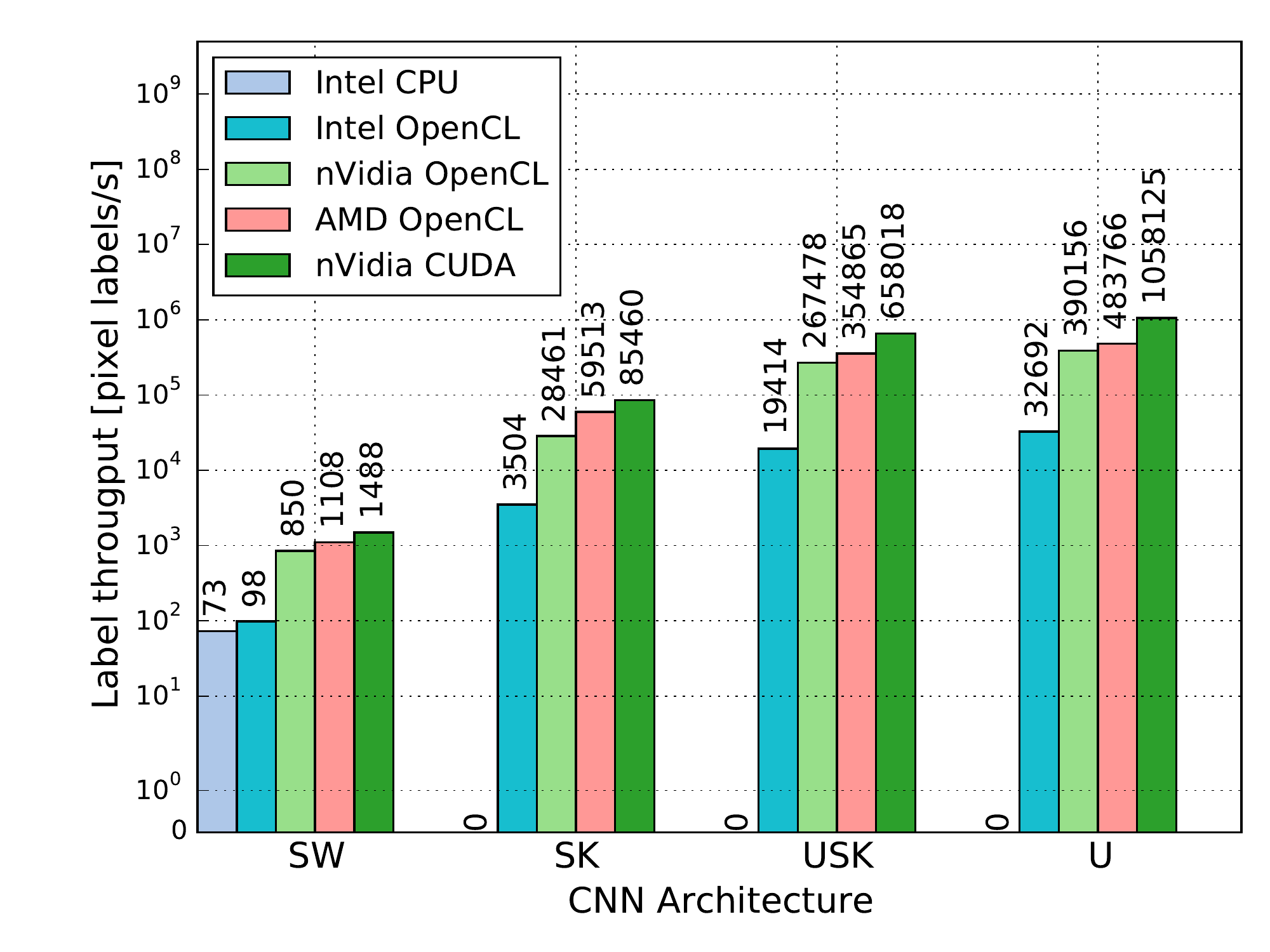}
	\caption{Log scaled labeling throughput}
	\label{fig:loglabelthroughput}
\end{figure}

\begin{figure}[H]
	\centering
	\includegraphics[width=0.8\textwidth]{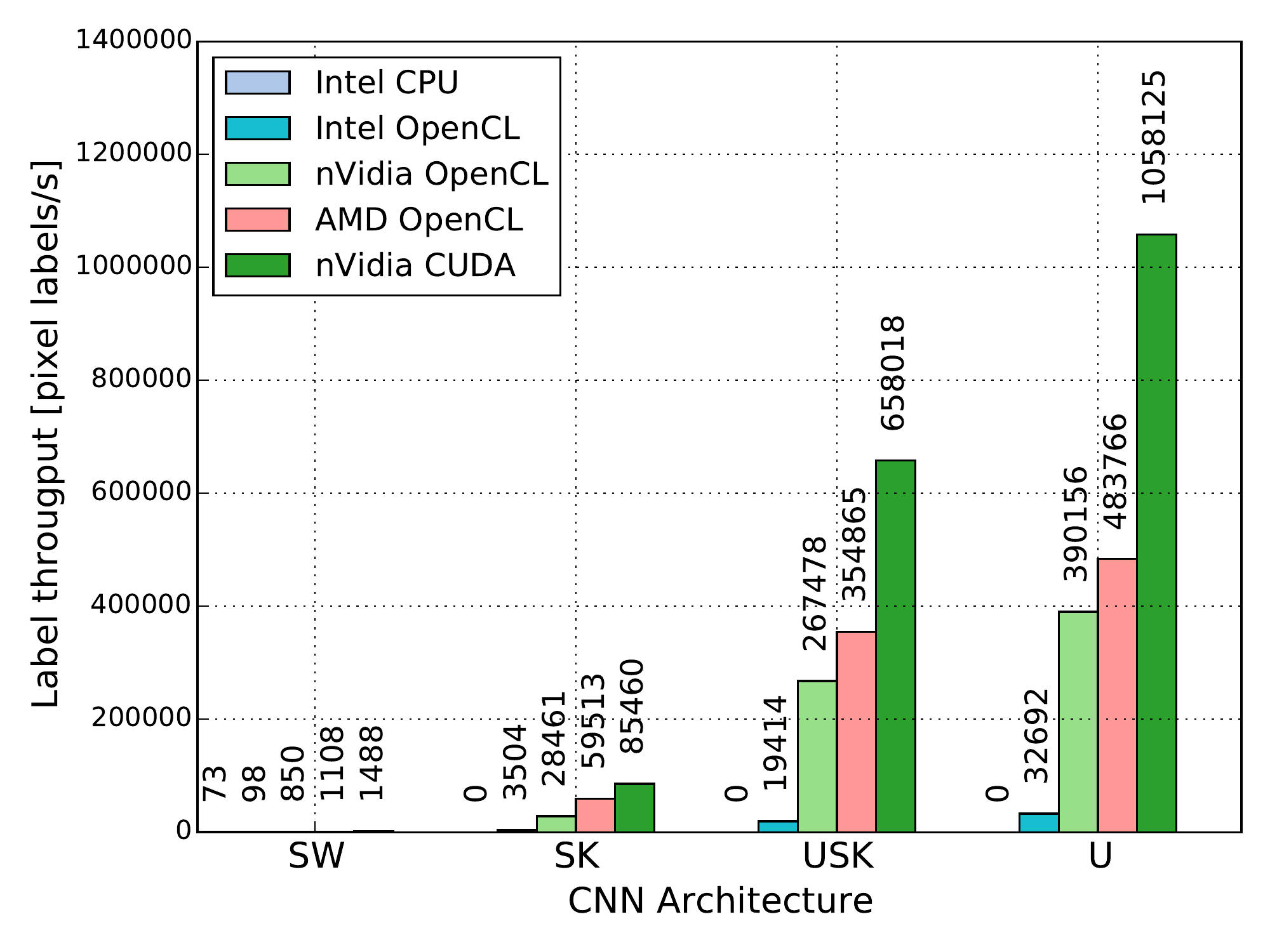}
	\caption{Linear scaled labeling throughput}
	\label{fig:labelthroughput}
\end{figure}

Figures \ref{fig:loglabelthroughput} and \ref{fig:labelthroughput} represent the values of Table \ref{tab:labelthroughput} in both logarithmic and linear scale.

While the CUDA backend is usually the fastest, the nVidia GPU performs slower, as expected by the FLOP values in Table \ref{tab:benchhardware}, relative to the AMD GPU when using OpenCL on both of them. Explaining this behavior needs insight into the individual network layer performance (Section \ref{sec:layerperformance}) and the convolution operations (Section \ref{sec:convmethods}).

The network performance during training was not assessed in this report, as training the models on the available data was not a limiting objective. Back propagation is usually slower than forwarding, as the differential maps, gradients and weight updates have to be computed.

\section{Layer Performance Analysis}
\label{sec:layerperformance}

This section takes apart the neural networks down to individual layers to assess why certain networks are faster than others and find potential to optimize models.

\begin{figure}[H]
	\centering
	\includegraphics[width=0.2\textwidth]{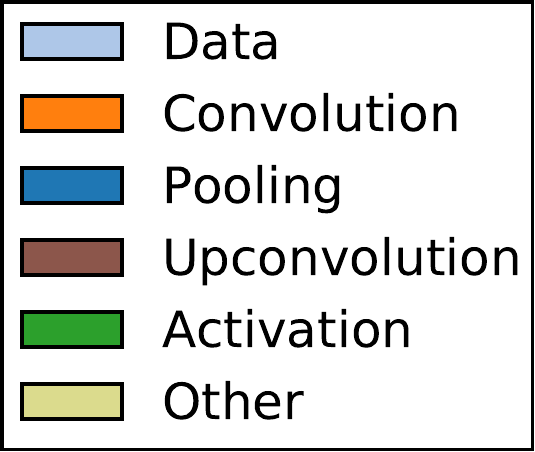}
	\caption{Layer-wise analysis legend.}
	\label{fig:layerwiselegend}
\end{figure}

\subsection{SK-Net}
\label{sec:skbenchmark}
\begin{figure}[H]
	\centering
	\includegraphics[width=0.9\textwidth]{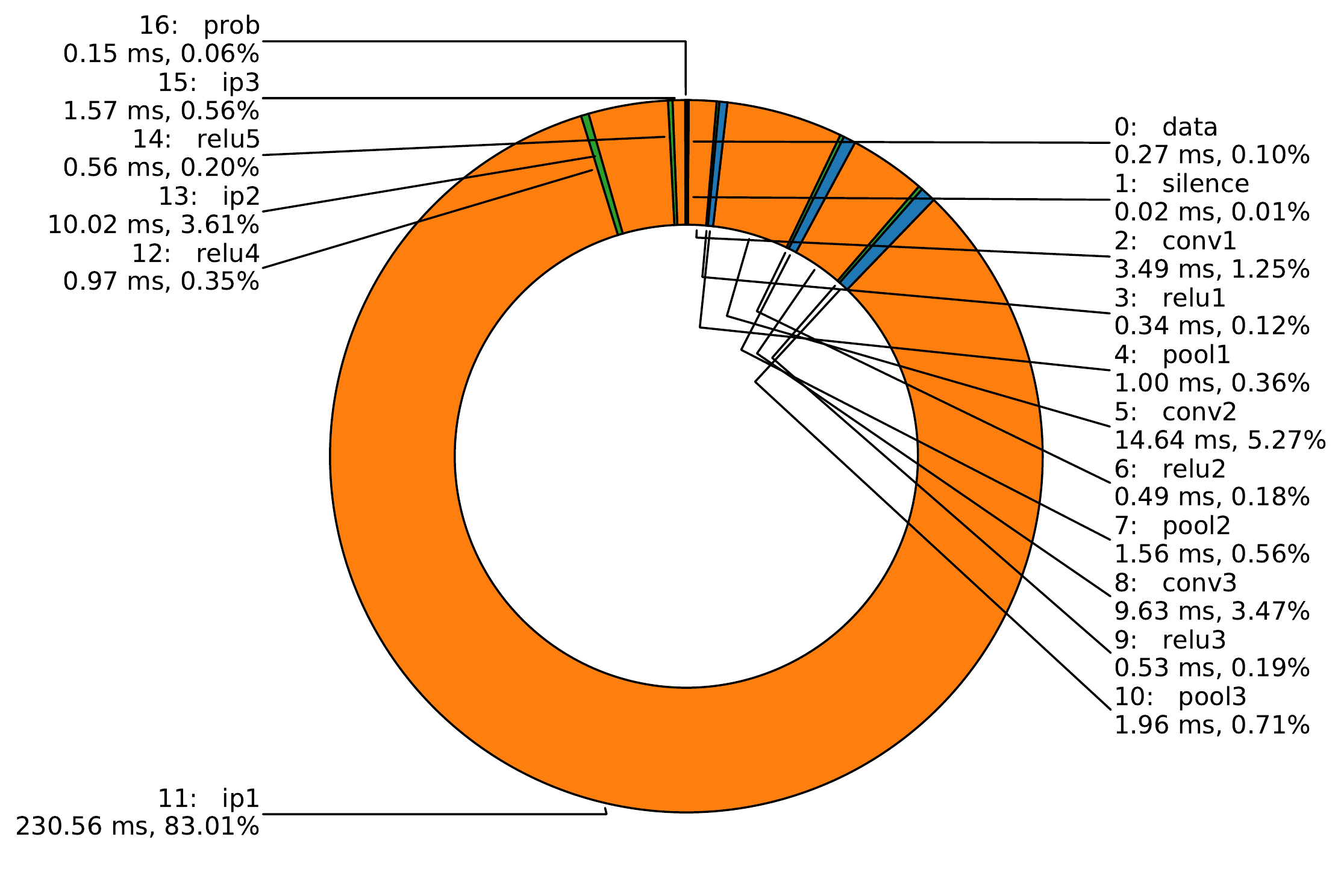}
	\caption{Layer-wise timings of SK-Net.}
	\label{fig:skfwlayerwise}
\end{figure}

Looking at Figure \ref{fig:skfwlayerwise}, the convolution layers and especially the \textit{ip1} layer (\SI{83}{\%}) are responsible for the overall performance. There are many input (192) and output (1024) feature maps (see Section \ref{sec:sknet}) and therefore a high computational complexity ($\mathcal{O}(f_{\text{out}}\cdot w^2 \cdot (2\cdot f_{\text{in}}\cdot k^2-1))$) (see Section \ref{sec:convmethods}) on the \textit{ip1} layer.

\begin{figure}[H]
	\centering
	\includegraphics[width=0.9\textwidth]{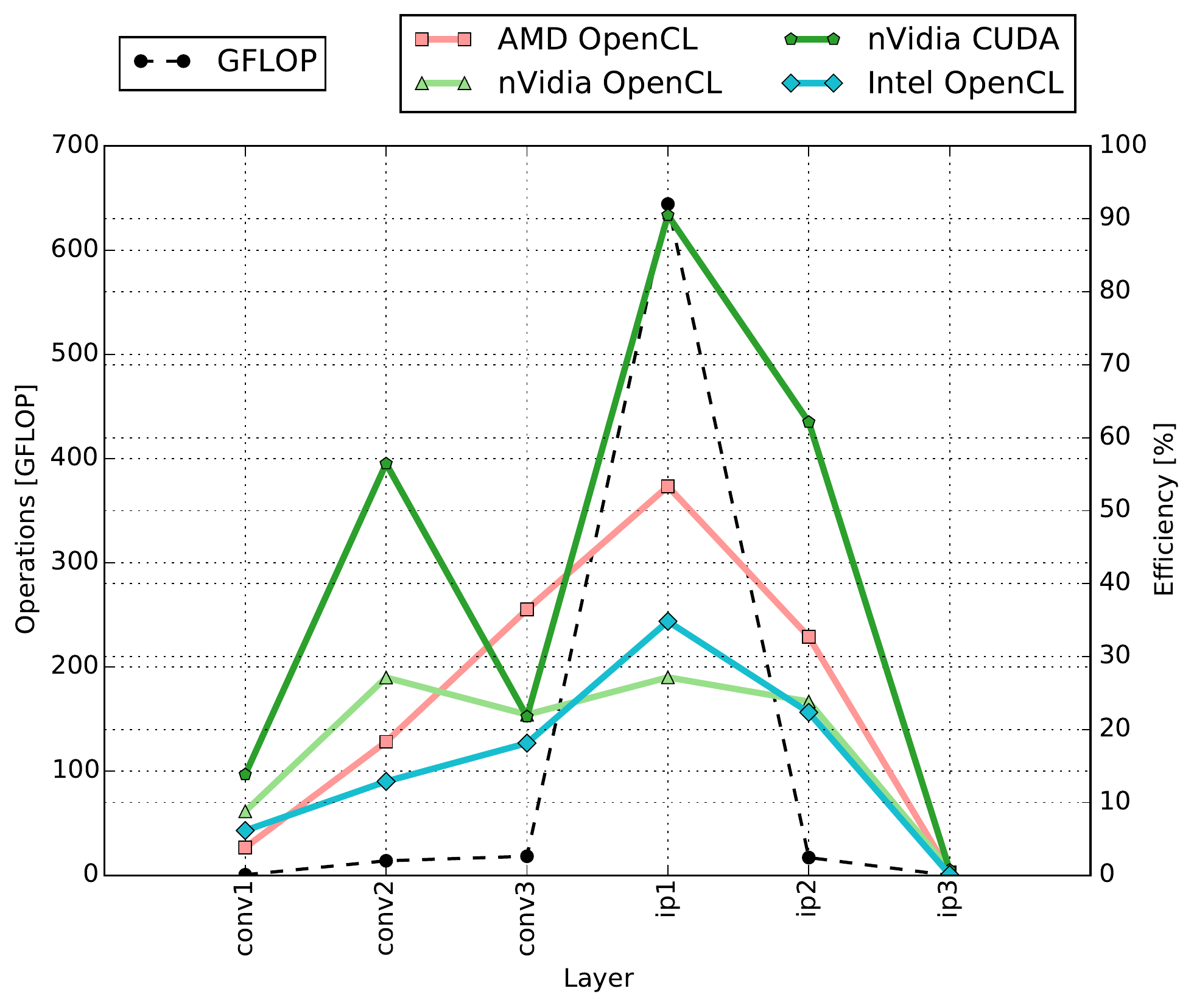}
	\caption{FLOP efficiency of the convolution layers in the SK-Net.}
	\label{fig:skefficiency}
\end{figure}

As the convolution layers only consist of a fast memory copy operation to arrange the data, so that matrix-matrix multiplications through an optimized BLAS become possible, and the SGEMM (single precision general matrix-matrix multiplication) call itself, the efficiency values in Table \ref{tab:skefficiency} and Figure \ref{fig:skefficiency} are direct proxies for how efficient the BLAS works on the devices.

Layers that operate with complexity $\mathcal{O}(f_{\text{in}}\cdot w^2)$, which includes all other layers used in the SK, USK and U networks, only contribute a small fraction to the total forwarding time (less than \SI{1}{\%} per layer).

\begin{table}[H]
\begin{center}
	\begin{tabular}{|l|r|r|r|r|r|}
		\hline
		\textbf{Layer} & \textbf{GFLOP} & \textbf{AMD OCL} & \textbf{nV OCL} & \textbf{nV CUDA} & \textbf{Intel OCL}\\\hline
		conv1 & 0.70 & \SI{3.83}{\%} & \SI{8.79}{\%} & \SI{13.84}{\%} & \SI{6.16}{\%}\\\hline
		conv2 & 14.06 & \SI{18.33}{\%} & \SI{27.13}{\%} & \SI{56.45}{\%} & \SI{12.89}{\%}\\\hline
		conv3 & 18.40 & \SI{36.46}{\%} & \SI{22.05}{\%} & \SI{21.78}{\%} & \SI{18.14}{\%}\\\hline
		ip1 & 644.23 & \SI{53.32}{\%} & \SI{27.15}{\%} & \SI{90.50}{\%} & \SI{34.83}{\%}\\\hline
		ip2 & 17.17 & \SI{32.72}{\%} & \SI{23.83}{\%} & \SI{62.15}{\%} & \SI{22.35}{\%}\\\hline
		ip3 & 0.03 & \SI{0.41}{\%} & \SI{0.75}{\%} & \SI{0.75}{\%} & \SI{0.12}{\%}\\\hline
	\end{tabular}
	\captionof{table}{FLOP efficiency for the convolution layers in the SK-Net. The values are relative to Table \ref{tab:benchhardware} FLOP performances.}
	\label{tab:skefficiency}
\end{center}
\end{table}

Even though \textit{ip1} is the most expensive layer, it also is the most efficient (see Figure \ref{fig:skefficiency}) on all devices, when using the BLAS which is most optimized. This means clBLAS for AMD, cuBLAS for nVidia and OpenBLAS for Intel.

It is important to note that clBLAS is not yet as optimized as cuBLAS for this type of matrix-matrix multiplications (see Section \ref{sec:convmethods}). Both GPUs currently perform worse than expected with the OpenCL backend. As both clBLAS and the \textit{Project Greentea} are still quite recent projects, the performance is expected to increase with further optimizations in the future.

Other convolution layers are over 30 times less expensive and down to half as efficient. With small convolutions, the memory copy and kernel launch overhead lower the efficiency. Very small convolutions such as \textit{ip3} and \textit{conv1} may also not fully utilize the many threads available on GPUs (see Table \ref{tab:benchhardware}). As the inefficient layers have short computation times here, they are not very important in optimization.

Only if the network mostly consists of such inefficient layers (which is not the case with SK-Net, but does apply to SW-Net), the number of feature maps and convolution sizes should be increased, if it improves the segmentation. Using minibatches ($n > 1$) can increase the GPU utilization on OpenCL when using multiple queues ($q > 1$) and the efficiency on CUDA when using cuDNN \cite{2014arXiv1410.0759C}.

\subsection{U-Net}
\label{sec:ubenchmark}
\begin{figure}[H]
	\centering
	\includegraphics[width=0.9\textwidth]{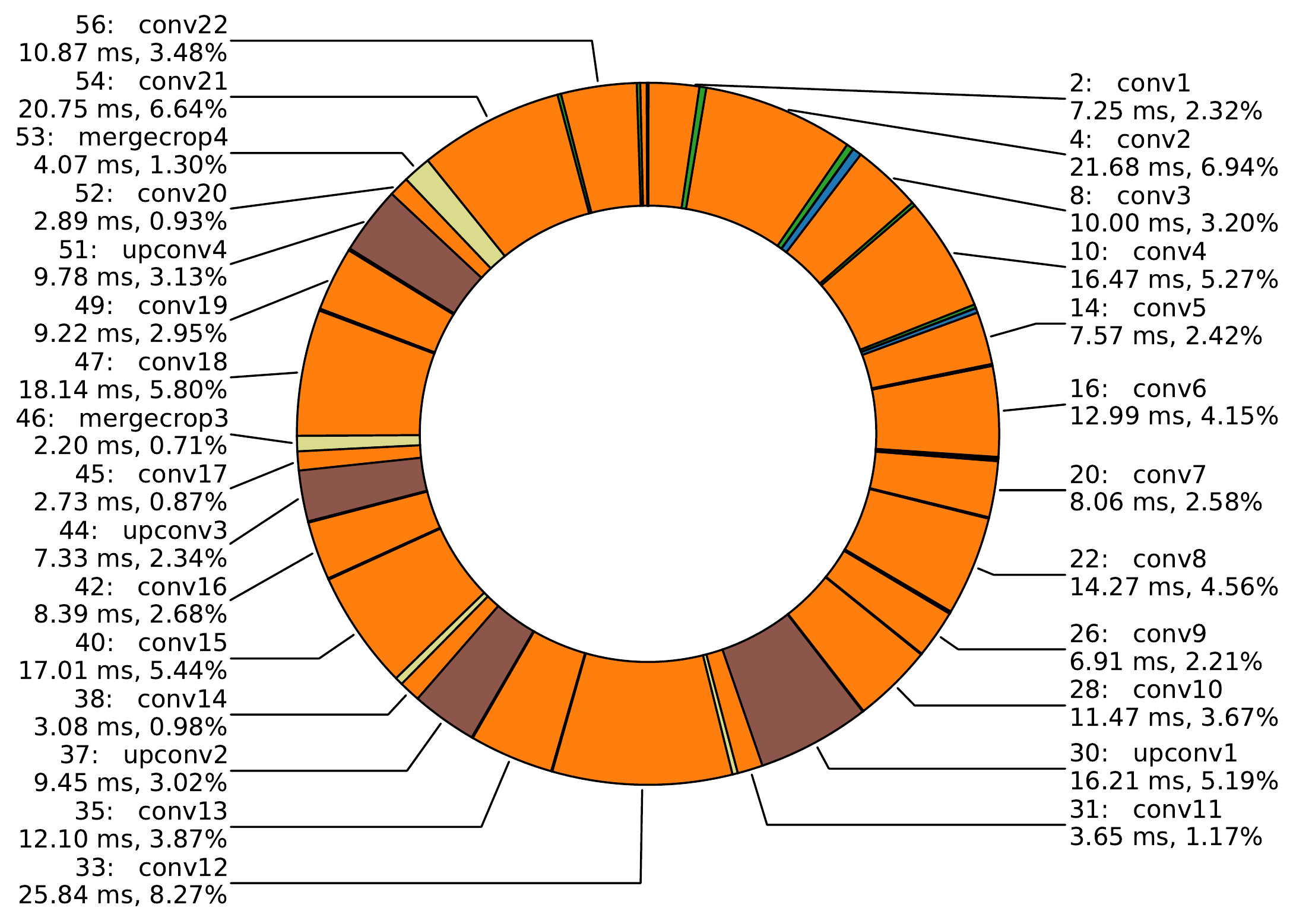}
	\caption{Layer-wise timings of U-Net, excluding layers that contribute less than \SI{0.5}{\%} of the total forwarding time.}
	\label{fig:ufwlayerwise}
\end{figure}

\begin{table}[H]
\begin{center}
	\begin{tabular}{|l|r|r|r|r|r|}
		\hline
		\textbf{Layer} & \textbf{GFLOP} & \textbf{AMD OCL} & \textbf{nV OCL} & \textbf{nV CUDA} & \textbf{Intel OCL}\\\hline
		conv1 & 1.10 & \SI{2.90}{\%} & \SI{6.58}{\%} & \SI{7.69}{\%} & \SI{5.50}{\%}\\\hline
		conv2 & 23.77 & \SI{20.92}{\%} & \SI{19.54}{\%} & \SI{43.92}{\%} & \SI{7.60}{\%}\\\hline
		conv3 & 11.72 & \SI{22.35}{\%} & \SI{26.10}{\%} & \SI{53.21}{\%} & \SI{19.15}{\%}\\\hline
		conv4 & 23.11 & \SI{26.78}{\%} & \SI{22.90}{\%} & \SI{58.79}{\%} & \SI{13.35}{\%}\\\hline
		conv5 & 11.23 & \SI{28.31}{\%} & \SI{29.78}{\%} & \SI{63.92}{\%} & \SI{23.50}{\%}\\\hline
		conv6 & 21.81 & \SI{32.06}{\%} & \SI{24.70}{\%} & \SI{73.71}{\%} & \SI{24.66}{\%}\\\hline
		conv7 & 10.27 & \SI{24.32}{\%} & \SI{32.31}{\%} & \SI{77.68}{\%} & \SI{40.24}{\%}\\\hline
		conv8 & 19.33 & \SI{25.85}{\%} & \SI{24.91}{\%} & \SI{88.07}{\%} & \SI{42.55}{\%}\\\hline
		conv9 & 8.49 & \SI{23.44}{\%} & \SI{32.93}{\%} & \SI{76.82}{\%} & \SI{54.12}{\%}\\\hline
		conv10 & 14.80 & \SI{24.62}{\%} & \SI{25.05}{\%} & \SI{84.03}{\%} & \SI{54.30}{\%}\\\hline
		conv11 & 3.29 & \SI{17.17}{\%} & \SI{26.57}{\%} & \SI{78.12}{\%} & \SI{38.53}{\%}\\\hline
		conv12 & 27.52 & \SI{20.33}{\%} & \SI{32.39}{\%} & \SI{67.04}{\%} & \SI{40.50}{\%}\\\hline
		conv13 & 12.76 & \SI{20.12}{\%} & \SI{23.86}{\%} & \SI{75.78}{\%} & \SI{43.25}{\%}\\\hline
		conv14 & 2.83 & \SI{17.56}{\%} & \SI{24.03}{\%} & \SI{71.13}{\%} & \SI{28.54}{\%}\\\hline
		conv15 & 24.54 & \SI{27.54}{\%} & \SI{29.25}{\%} & \SI{71.21}{\%} & \SI{22.74}{\%}\\\hline
		conv16 & 11.79 & \SI{26.83}{\%} & \SI{24.10}{\%} & \SI{75.04}{\%} & \SI{25.90}{\%}\\\hline
		conv17 & 2.62 & \SI{18.29}{\%} & \SI{21.74}{\%} & \SI{61.10}{\%} & \SI{23.22}{\%}\\\hline
		conv18 & 23.12 & \SI{24.32}{\%} & \SI{27.82}{\%} & \SI{62.06}{\%} & \SI{16.63}{\%}\\\hline
		conv19 & 11.32 & \SI{23.43}{\%} & \SI{21.69}{\%} & \SI{59.62}{\%} & \SI{13.79}{\%}\\\hline
		conv20 & 2.51 & \SI{16.54}{\%} & \SI{19.09}{\%} & \SI{39.84}{\%} & \SI{17.82}{\%}\\\hline
		conv21 & 22.42 & \SI{20.62}{\%} & \SI{23.77}{\%} & \SI{45.12}{\%} & \SI{10.44}{\%}\\\hline
		conv22 & 11.09 & \SI{19.47}{\%} & \SI{18.84}{\%} & \SI{43.50}{\%} & \SI{8.09}{\%}\\\hline
		ip1 & 0.04 & \SI{0.79}{\%} & \SI{1.84}{\%} & \SI{2.40}{\%} & \SI{0.35}{\%}\\\hline
	\end{tabular}
	\captionof{table}{FLOP efficiency for the convolution layers in U-Net. The values are relative to Table \ref{tab:benchhardware} FLOP performances.}
	\label{tab:uefficiency}
\end{center}
\end{table}

In the U-Net, most convolutions have the same order of magnitude in complexity (see Table \ref{tab:uefficiency}). The result is a very balanced network in terms of forward timings (see Figure \ref{fig:ufwlayerwise}). The balancing comes from trading feature map size against feature map count towards the middle of the network.

The upconvolution layers contribute each up to \SI{5.2}{\%} of the network forwarding time. This is less efficient than the optimum as only 4-nearest neighbor interpolation with constant weights is computed. This would not be more effort than a memory copy operation during forwarding and accumulating of the four nearest neighbor values during backward computation. Currently, it is implemented using a Caffe deconvolution, which is a reversed convolution layer. This layer already existed in Caffe, while no direct upsampling of feature maps is implemented. The convolution kernels are grouped, meaning each upsampling kernel only considers a single feature map. One advantage is that the deconvolution layer also allows adaptive weights and other interpolations such as bilinear upsampling.

\begin{figure}[H]
	\centering
	\includegraphics[width=0.9\textwidth]{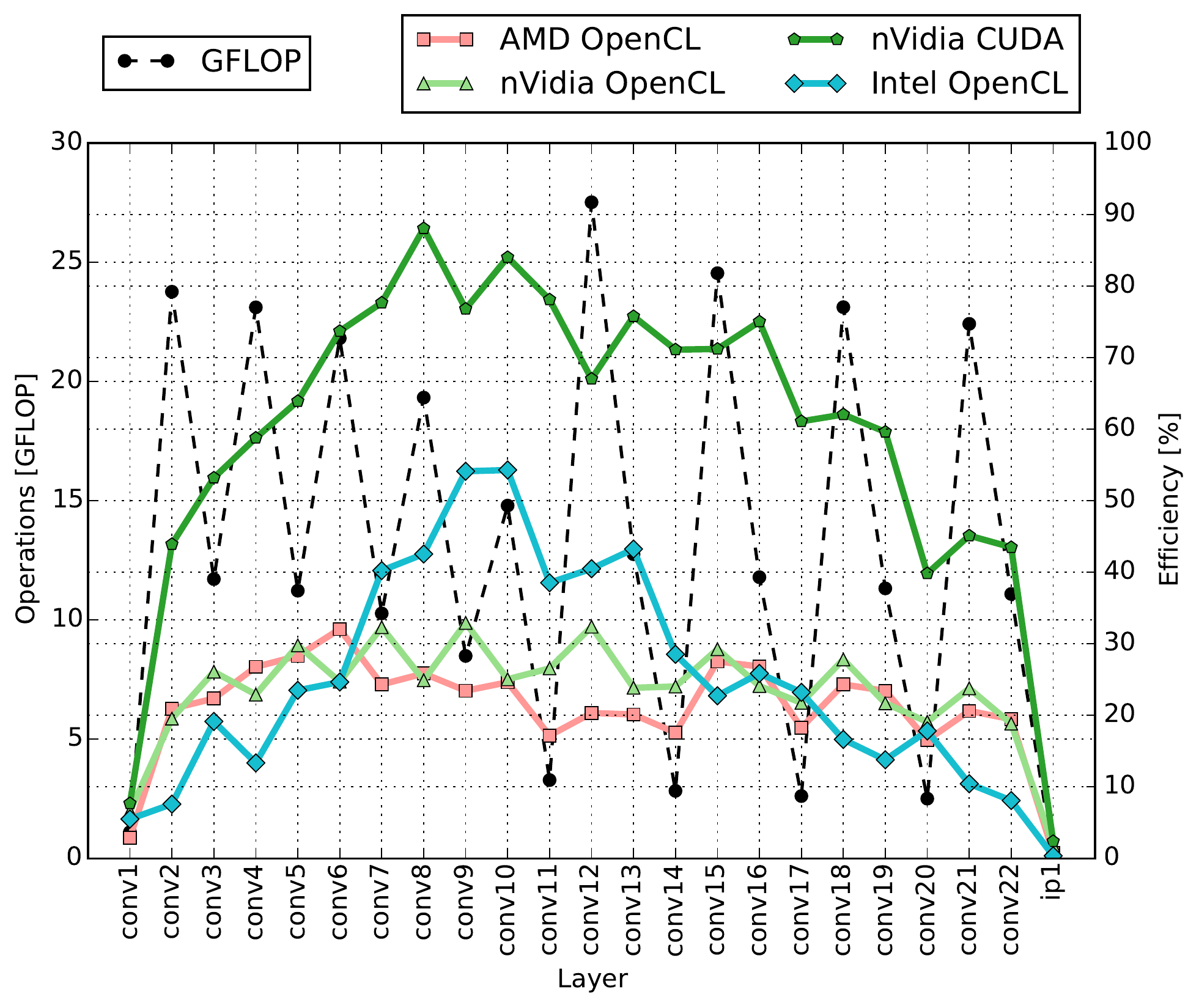}
	\caption{FLOP efficiency of the convolution layers in U-Net.}
	\label{fig:uefficiency}
\end{figure}

\subsection{USK-Net}
\label{sec:uskbenchmark}

\begin{table}[H]
\begin{center}
	\begin{tabular}{|l|r|r|r|r|r|}
		\hline
		\textbf{Layer} & \textbf{GFLOP} & \textbf{AMD OCL} & \textbf{nV OCL} & \textbf{nV CUDA} & \textbf{Intel OCL}\\\hline
		conv1 & 0.64 & \SI{2.51}{\%} & \SI{4.97}{\%} & \SI{7.08}{\%} & \SI{5.43}{\%}\\\hline
		conv2 & 13.75 & \SI{21.55}{\%} & \SI{19.25}{\%} & \SI{42.11}{\%} & \SI{8.04}{\%}\\\hline
		conv3 & 26.25 & \SI{27.06}{\%} & \SI{20.14}{\%} & \SI{62.58}{\%} & \SI{11.73}{\%}\\\hline
		conv4 & 21.81 & \SI{26.99}{\%} & \SI{21.78}{\%} & \SI{62.61}{\%} & \SI{12.57}{\%}\\\hline
		conv5 & 18.92 & \SI{24.93}{\%} & \SI{27.45}{\%} & \SI{63.93}{\%} & \SI{13.06}{\%}\\\hline
		ip1 & 141.76 & \SI{45.93}{\%} & \SI{31.25}{\%} & \SI{86.64}{\%} & \SI{27.82}{\%}\\\hline
		ip2 & 4.43 & \SI{23.53}{\%} & \SI{28.14}{\%} & \SI{72.69}{\%} & \SI{42.59}{\%}\\\hline
		conv6 & 4.42 & \SI{23.56}{\%} & \SI{21.29}{\%} & \SI{63.39}{\%} & \SI{31.44}{\%}\\\hline
		conv7 & 29.44 & \SI{27.31}{\%} & \SI{27.88}{\%} & \SI{61.29}{\%} & \SI{17.26}{\%}\\\hline
		conv8 & 9.66 & \SI{18.44}{\%} & \SI{20.45}{\%} & \SI{46.17}{\%} & \SI{7.34}{\%}\\\hline
		ip3 & 0.02 & \SI{0.53}{\%} & \SI{1.73}{\%} & \SI{1.73}{\%} & \SI{0.36}{\%}\\\hline
	\end{tabular}
	\captionof{table}{FLOP efficiency for the convolution layers in USK-Net. The values are relative to Table \ref{tab:benchhardware} FLOP performances.}
	\label{tab:uskefficiency}
\end{center}
\end{table}

In the USK network, the \textit{ip1} layer is again the most expensive and efficient one, but the network is more balanced as \textit{ip1} only contributes \SI{32}{\%} of the forwarding time, compared to \SI{83}{\%} in the original SK-Net. Efficiency and forwarding times are, as expected, a mixture of the SK- and U-Net.

Except for \textit{ip3} and \textit{conv1}, all layers have a relatively high efficiency. The inefficient layers go from $f_{\text{in}} = 64$ to $f_{\text{out}} = 2$ and $f_{\text{in}} = 3$ to $f_{\text{out}} = 64$ respectively. This will directly result in matrix multiplications with matrices that have strongly non-square shapes and are therefore less efficient (see Section \ref{sec:convmethods}).

\section{NUMA Issues}
\label{sec:numaissues}
An issue that came up testing the OpenCL hybrid backend (see Section \ref{sec:openclhybrid}) was that the performance did not scale as expected with systems that have more than one CPU. Such systems have non-unified memory access (NUMA) because the CPUs share one address space for memory, but every processor has its own cache and memory interface. Accessing data across the other CPU comes with a large performance penalty. Compute kernels, such as the matrix-matrix multiplication in the BLAS library or the custom OpenCL kernels, cause the threads to work on adjacent data. This means a write operation of one CPU is likely to invalidate cache lines across both CPUs. At this point, the synchronization overhead seems to become larger than any speedup of having additional cores working on the algorithms.

\begin{table}[H]
\begin{center}
	\begin{tabular}{|l|r|r|r|}
		\hline
		\textbf{Layer} & \textbf{i7-4790K} & \textbf{2x E5-2697v3} & \textbf{Speedup} \\\hline
		conv1 & \SI{22.19}{ms} & \SI{47.02}{ms} & 0.47\\\hline
		relu1 & \SI{2.68}{ms} & \SI{5.08}{ms} & 0.53\\\hline
		pool1 & \SI{14.65}{ms} & \SI{17.17}{ms} & 0.85\\\hline
		conv2 & \SI{213.16}{ms} & \SI{176.96}{ms} & 1.20\\\hline
		relu2 & \SI{4.94} {ms} & \SI{14.04}{ms} & 0.35\\\hline
		pool2 & \SI{40.06}{ms} & \SI{44.40}{ms} & 0.90\\\hline
		conv3 & \SI{198.15}{ms} & \SI{238.91}{ms} & 0.83\\\hline
		relu3 & \SI{5.85}{ms} & \SI{17.45}{ms} & 0.34\\\hline
		pool3 & \SI{57.08}{ms} & \SI{59.05}{ms} & 0.97\\\hline
		ip1 & \SI{3612.29}{ms} & \SI{4168.21}{ms} & 0.87\\\hline
		relu4 & \SI{13.59}{ms} & \SI{47.62}{ms} & 0.29\\\hline
		ip2 & \SI{150.03}{ms} & \SI{187.05}{ms} & 0.80\\\hline
		relu5 & \SI{5.90}{ms} & \SI{24.10}{ms} & 0.24\\\hline
		ip3 & \SI{56.25}{ms} & \SI{77.51} {ms} & 0.73\\\hline
		prob & \SI{0.48}{ms} & \SI{4.88}{ms} & 0.10\\\hline
	\end{tabular}
	\captionof{table}{4 core i7-4790K versus 2 CPU 28 core (NUMA) E5-2697v3, comparing SK network layerwise forward timings.}
	\label{tab:numaspeedup}
\end{center}
\end{table}

Each CPU has 14 cores, which gives 28 cores for the whole system. The number of threads for the frontend was set to 14, which gave the best performance by keeping at least the BLAS library temporarily tied to one processor by the operating system's scheduler. The OpenCL backend, which also allocates the memory, used 56 threads and allocated memory on both interfaces.\\
Table \ref{tab:numaspeedup} shows the impact of having NUMA issues, taking the SK network as an example. The layers are all sufficiently parallelized, which is evident when looking at Figure \ref{fig:skefficiency}. Differences of the CPU architectures are also not a possible explanation as both processors are based on the same instruction sets and generation. The memory interface should also be fast enough to keep up with the computations (see Table \ref{tab:benchhardware}).

Correcting for processor frequency, the speedup should be up to a factor $4.55\times$ using all 28 cores (ReLU and pooling layers) and up to $2.275\times$ using only one processor (inner product and convolution layers). The effective speedups measured are much slower:
\begin{itemize}
	\item Speedups between $0.1\times$ and $0.97\times$ on element-wise layers with complexity up to $\mathcal{O}(f_{\text{in}}\cdot w^2)$. Cache invalidation between the two processors (28 cores) seems to be dominant. The element-wise kernels run on the OpenCL backend.
	\item Speedups between $0.47\times$ and $1.20\times$ on convolution layers with complexity up to $\mathcal{O}(f_{\text{out}}\cdot w^2 \cdot (2\cdot f_{\text{in}}\cdot k^2-1))$ (matrix-matrix multiplication). The effects of OpenBLAS running on 14 cores and sub optimal memory allocations are dominant. It is a proxy for how the matrix-matrix multiplications used in all convolutions perform.
\end{itemize} 

To get the expected speedup, the two processors need to be presented to the Caffe library as two separate devices. Then the library can be used in two individual instances. As the OpenCL hybrid backend uses two separate parallelization mechanisms (OpenCL kernels and a parallelized BLAS), two solutions would need to be applied:
\begin{itemize}
	\item The Caffe frontend needs to be tied to the cores of one CPU, so that the BLAS library does not show NUMA issues.
	\item The OpenCL backend needs to split up the processor setup into sub-devices using device fission. The splitting rule needs to be that all cores belonging to one processor (tested by cache affinity) are tied to the same sub-device. Only one is then used per Caffe instance. Device fission is an extension to OpenCL that is already available (\textit{cl\_ext\_fission} \cite{KhronosFission}).
	\item The cores used in the frontend and selected sub-device need to be the same.
\end{itemize}

Due to not having permanent access to a system with two processors and OpenCL installed, I did not have time to test out the solutions. Implementing the solutions remains as an open issue at the time of the project.

\newpage
\section{Alexnet}
For comparison how the backends and devices perform on a widely used network for image classification that uses minibatches (with $n = 10$, $w=227$, $f_{\text{in}}=3$) and multiple OpenCL queues ($q = 8$), the Alexnet \cite{NIPS2012_4824} included in the Caffe library was also evaluated.

\begin{table}[H]
\begin{center}
	\begin{tabular}{|l|l|r|r|r|r|}
		\hline
		\textbf{Device} & \textbf{Backend} & \textbf{Forward} & \textbf{Backward} & \textbf{Total} & \textbf{Forward Speedup}\\\hline
		Intel & CPU & \SI{452.807}{ms} & \SI{355.553}{ms} & \SI{808.420}{ms} & 1.00 \\\hline
		Intel & OpenCL & \SI{238.188}{ms} & \SI{152.811}{ms} & \SI{391.160}{ms} & 1.90 \\\hline
		AMD & OpenCL & \SI{52.348}{ms} & \SI{128.009}{ms} & \SI{180.420}{ms} & 8.65 \\\hline
		nVidia & OpenCL & \SI{26.055}{ms} & \SI{47.446}{ms} & \SI{73.540}{ms} & 17.37\\\hline
		nVidia & CUDA & \SI{20.899}{ms} & \SI{16.730}{ms} & \SI{37.718}{ms} & 21.66\\\hline
	\end{tabular}
	\captionof{table}{Alexnet timings, average forward-backward pass over 50 iterations.}
	\label{tab:alexnetbench}
\end{center}
\end{table}

The CUDA backend has an advantage over the OpenCL backend in terms of speed, but is less versatile. The AMD GPU seems to be less efficient with minibatches and smaller matrix-matrix multiplications than the nVidia GPU, which is why the AMD GPU performs worse than the nVidia GPU on the same backend. With the SK-, USK- and U-Net (Section \ref{sec:labelthroughput}), the AMD GPU performs better using the same (OpenCL) backend on both GPUs.

Especially the backward computation is much slower (by a factor of three) using OpenCL instead of CUDA. The algorithms used are the same, therefore this difference is difficult to explain. Possibly, different optimizations need to be applied in the backward step of convolution layers, which can have a sequential bottleneck by adding up the gradients over the minibatch.

Using the OpenCL hybrid backend on the Intel CPU outperforms the (legacy) CPU backend by almost a factor of two. The speedup comes from parallelization of  \textit{Greentea}'s OpenCL compute kernels, which are only single-threaded in the Caffe CPU backend. The BLAS library used for convolutions is multithreaded in both cases.

\chapter{Results}
\label{ch:results}

\section{Introduction}
In this chapter, the results of training the models from Chapter \ref{ch:models} are presented.

The evaluation is based on:
\begin{itemize}
	\item Two data sets (DS1 and DS2: see Chapter \ref{ch:datasets}).
	\item Three models (SK, U, USK: see Chapter \ref{ch:models}).
	\item Two training loss functions (Softmax and Malis).
	\item One processing loss function (Softmax).
	\item Three training configurations (Softmax, Malis and Softmax + Malis).
	\item 10'000 training iterations per configuration, respectively 20'000 for combined training.
	\item Three error objectives (rand, warping and pixel error).
\end{itemize}

The amount of training iterations was chosen so that training of all 18 combinations was feasible during one week on two AMD W9100 GPUs, providing a total of \SI{10}{TFLOP/s} \cite{W9100Whitepaper}.\\
The training data is not very large in both cases (see Chapter \ref{ch:datasets}) and thus the loss always converged in under 10'000 iterations for each training method. It is possible that some trainings did overfit as no early stopping was applied. Technically, the networks did not get to see the same amount of examples during training even though the iterations are the same, as the chosen output sizes of each network were set differently. However, due to gradient accumulation, different learning rates and weight initialization it is hard to estimate the effect of the amount of labels seen. As all networks converged to a stable loss, it should negligible.

\section{Analysis on DS1}
\label{sec:ds1analysis}

\subsection{Training}
The training parameters used on DS1 were set to not use the error masking functionality. Masking usually gives thicker membrane (background) labels when used with Softmax by balancing out the amount of background error against foreground (cell interior) error.

Malis loss was run without using a patch prior for preferring training patches based on their label histogram. Softmax loss on the other hand was used with the patch prior enabled. This is justified by the different characters of the loss functions: Softmax computes a per-pixel error while Malis gives errors at problematic pixels only, which can be very concentrated on a few pixels (see Section \ref{sec:malisloss}).

In the patch pre-processing step, the images were enhanced with CLAHE (contrast limited adaptive histogram equalization) and normalized in the range of $[-1.0,1.0]$.

To get more training data, the images were blurred with a randomly chosen 5 by 5 Gaussian kernel. The training patches were also rotated to multiples of \ang{90} and mirrored randomly (horizontal and vertical).

Details of the pre-processing and label priors are described in Chapter \ref{ch:caffeneuraltool}. The exact training parameters are stored as prototxt configuration in the Caffe Neural Models repository \cite{CaffeNeuralModels}.

Interestingly, it was not possible to start training of the SK network with Malis loss directly. The loss did not converge, and the output feature maps drifted to being classified as only foreground or only background. Therefore, the weights of the network in \textit{Malis only}-training were initialized using 4000 iterations of Softmax training first. This was the lowest number of iterations where the training converged to a small loss afterwards. The other network architectures did not show such a behavior and trained well when starting with Malis directly.\\
This is related to weight initialization as well as the fact that Malis works better on bigger training patch sizes. Training of SK was limited to 128 by 128 pixels output, which is much less context for Malis to work on than with USK and U with 512 by 512 and 388 by 388 pixels respectively.

\subsection{Numerical}

\begin{table}[H]
\begin{center}
	\begin{tabular}{|l|l|l|r|r|r|}
		\hline
		\textbf{Rank} & \textbf{Network} & \textbf{Loss Function} & \textbf{Rand} & \textbf{Warping} & \textbf{Pixel}\\\hline
		1. & USK & Softmax + Malis & \textbf{0.031251499} & 0.000423193 & 0.040650392\\\hline
		2. & SK & Softmax + Malis & 0.043871103 & \textbf{0.000324726} & 0.052724789\\\hline
		3. & USK & Malis & 0.045955948 & 0.000578165 & 0.058407081\\\hline
		4. & U & Softmax + Malis & 0.050799468 & 0.000487328 & 0.044159558\\\hline
		5. & SK & Malis & 0.063852001 & 0.000386953 & 0.055678171\\\hline
		6. & USK & Softmax & 0.067068677 & 0.000418186 & \textbf{0.030634890}\\\hline
		7. & U & Malis & 0.078736573 & 0.000520945 & 0.059127919\\\hline
		8. & U & Softmax & 0.091736036 & 0.000595331 & 0.033451667\\\hline
		9. & SK & Softmax & 0.123334489 & 0.000549316 & 0.034416625\\\hline
	\end{tabular}
	\captionof{table}{DS1 error evaluation (lower is better).}
	\label{tab:ds1ranking}
\end{center}
\end{table}

\newpage
Explanations for the Tables \ref{tab:ds1ranking} and \ref{tab:ds2ranking}:
\begin{itemize}
	\item Rank: The internal ranking, as indicated by the rand error.
	\item Loss Function: Softmax indicates 10'000 iterations with the Softmax loss function. Malis indicates 10'000 iterations with the Malis loss function. When both are indicated, the training was executed with 10'000 Softmax and then 10'000 Malis training iterations.
	\item Rand, warping and pixel: Error metrics, as proposed by Jain \etal \cite{Jain2010} and used in the ISBI 2012 challenge \cite{ISBI2012}. A script for evaluation is available for Fiji \cite{fijiscript} in the Caffe Neural Models repository \cite{CaffeNeuralModels}.
\end{itemize}

The ranking on data set DS1 is as expected: Taking the average rank, USK-Net performs better than SK-Net, which is more precise than U-Net.\\
The same goes for training methods: Using Softmax + Malis minimizes the rand error better than using Malis only, and Softmax outperforms Malis on pixel precision.
On one hand, this is because the Malis criterion improves the rand error by penalizing merge and split errors. On the other hand, training with Malis will also decrease the pixel accuracy, which can be seen when inspecting the result visually.\\
As Softmax training initializes the networks better than Malis, the best training method is to start with Softmax and then transit to Malis for fine tuning.\\
Obviously, the USK network architecture performed best on both pixel accuracy and rand error. Only for the warping error, the much slower SK-Net (see Section \ref{sec:labelthroughput}) performs best.

\subsection{Visual}
\begin{figure}[H]
	\center
	\begin{subfigure}{0.5\textwidth}
		\center
		\includegraphics[scale=0.18]{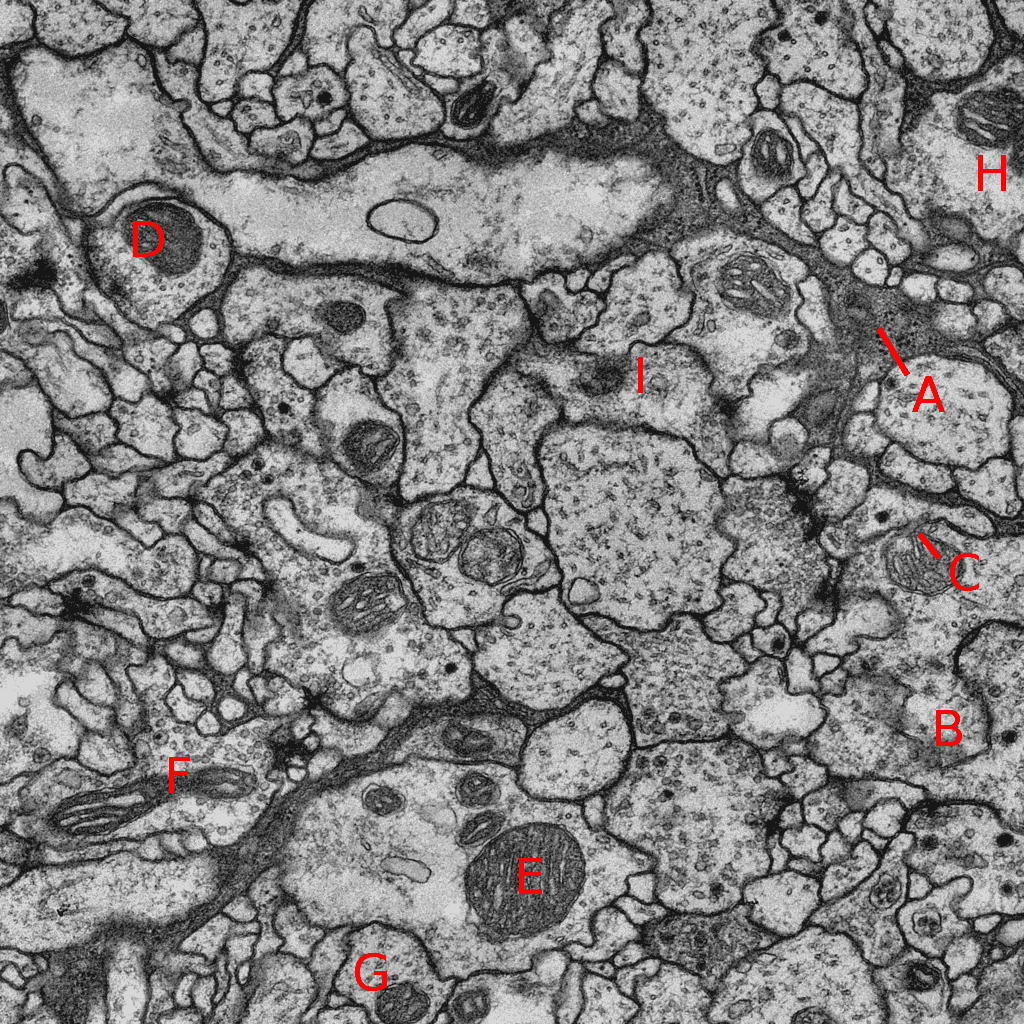}
	\end{subfigure}%
	\begin{subfigure}{.5\textwidth}
		\center
		\includegraphics[scale=0.18]{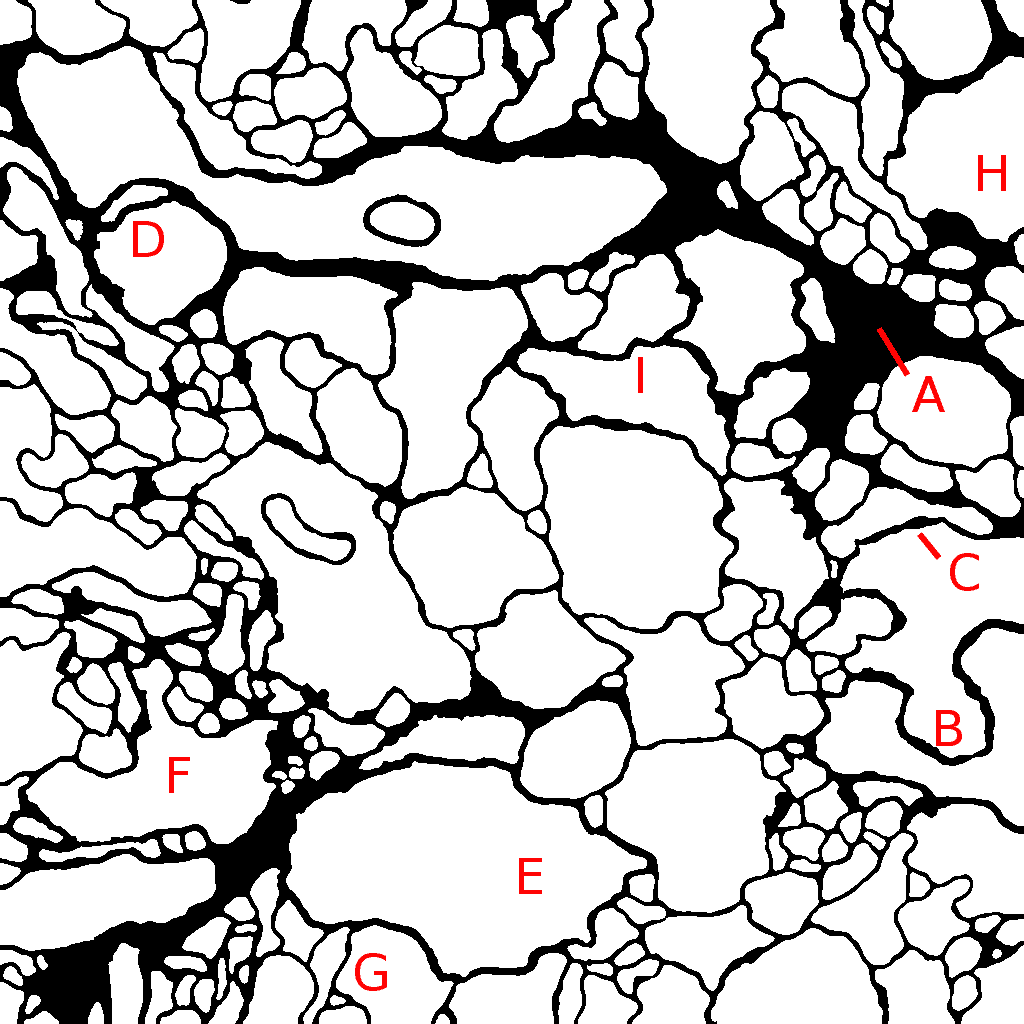}
	\end{subfigure}%
	\caption{DS1 ssTEM raw and corresponding ground truth, 1024 by 1024 pixels (Source: ssTEM \cite{DS1ssTEM}, \cite{CaffeNeuralModels}).}
	\label{fig:ds1araw}
\end{figure}

The visual analysis is based on the image number 2 of the DS1 stack (see Section \ref{sec:dataset1}). It includes a glia cell (Figure \ref{fig:ds1araw}, label A) which is considered as background. The thickness of it makes it harder to label correctly, especially with Malis loss.

\begin{figure}[H]
	\center
	\begin{subfigure}{0.5\textwidth}
		\center
		\includegraphics[scale=0.18]{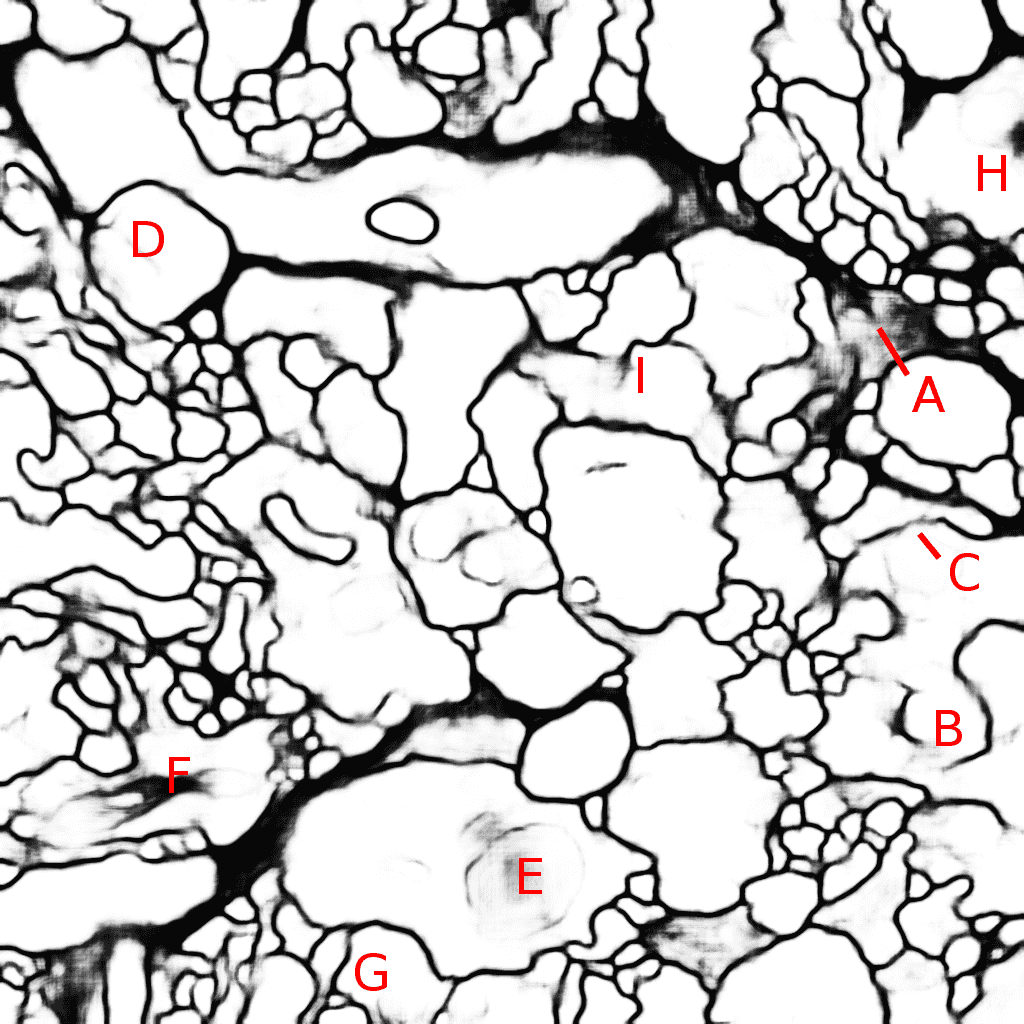}
		\caption{SK network, Softmax.}
	\end{subfigure}%
	\begin{subfigure}{.5\textwidth}
		\center
		\includegraphics[scale=0.18]{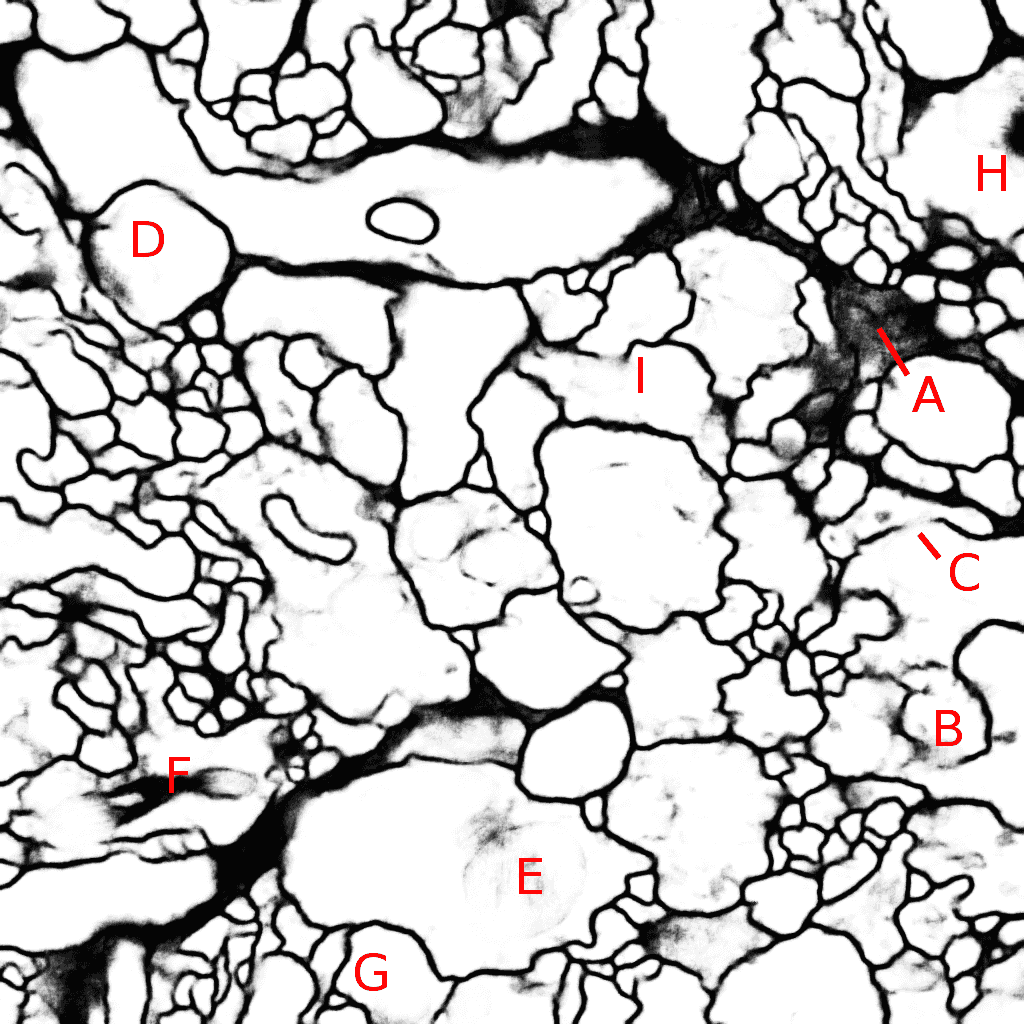}
		\caption{U network, Softmax.}
	\end{subfigure}%
	\newline
	\begin{subfigure}{0.5\textwidth}
		\center
		\includegraphics[scale=0.18]{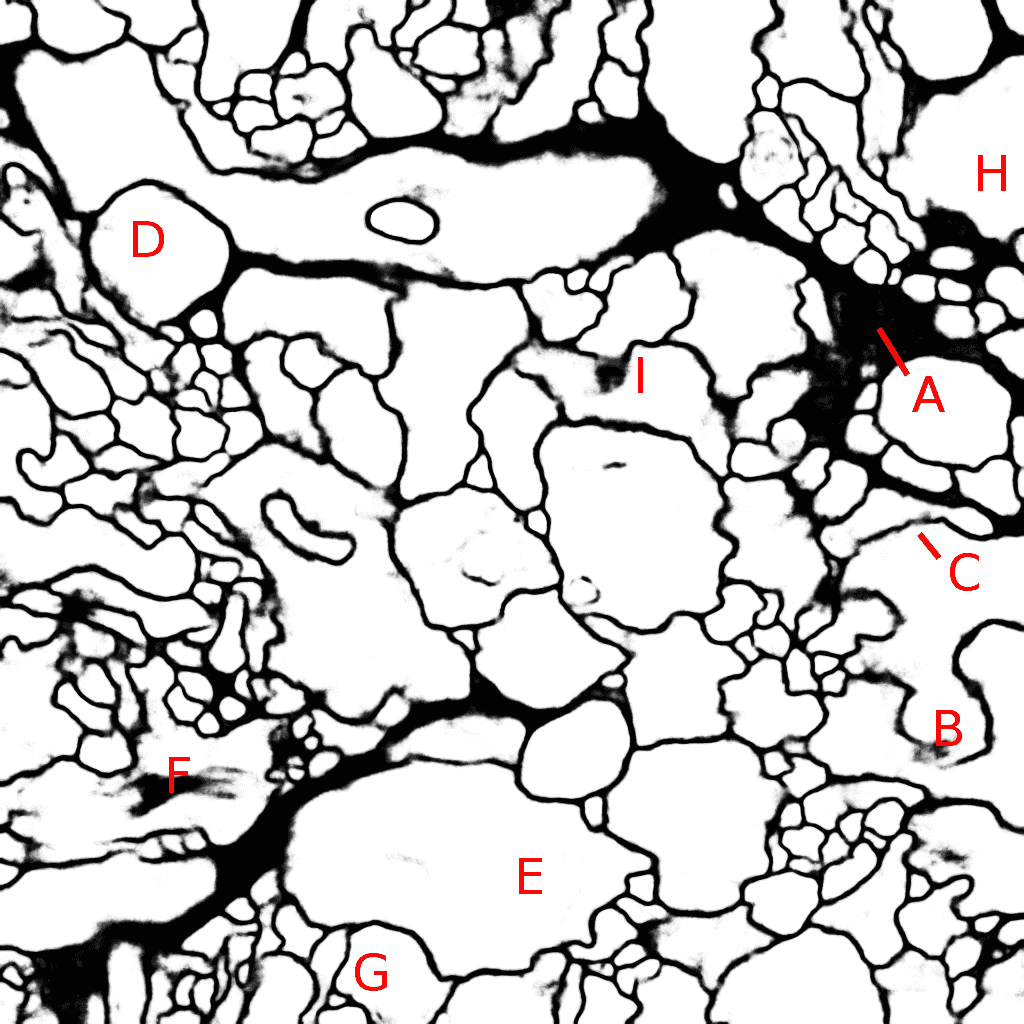}
		\caption{USK network, Softmax.}
	\end{subfigure}%
	\begin{subfigure}{.5\textwidth}
		\center
		\includegraphics[scale=0.18]{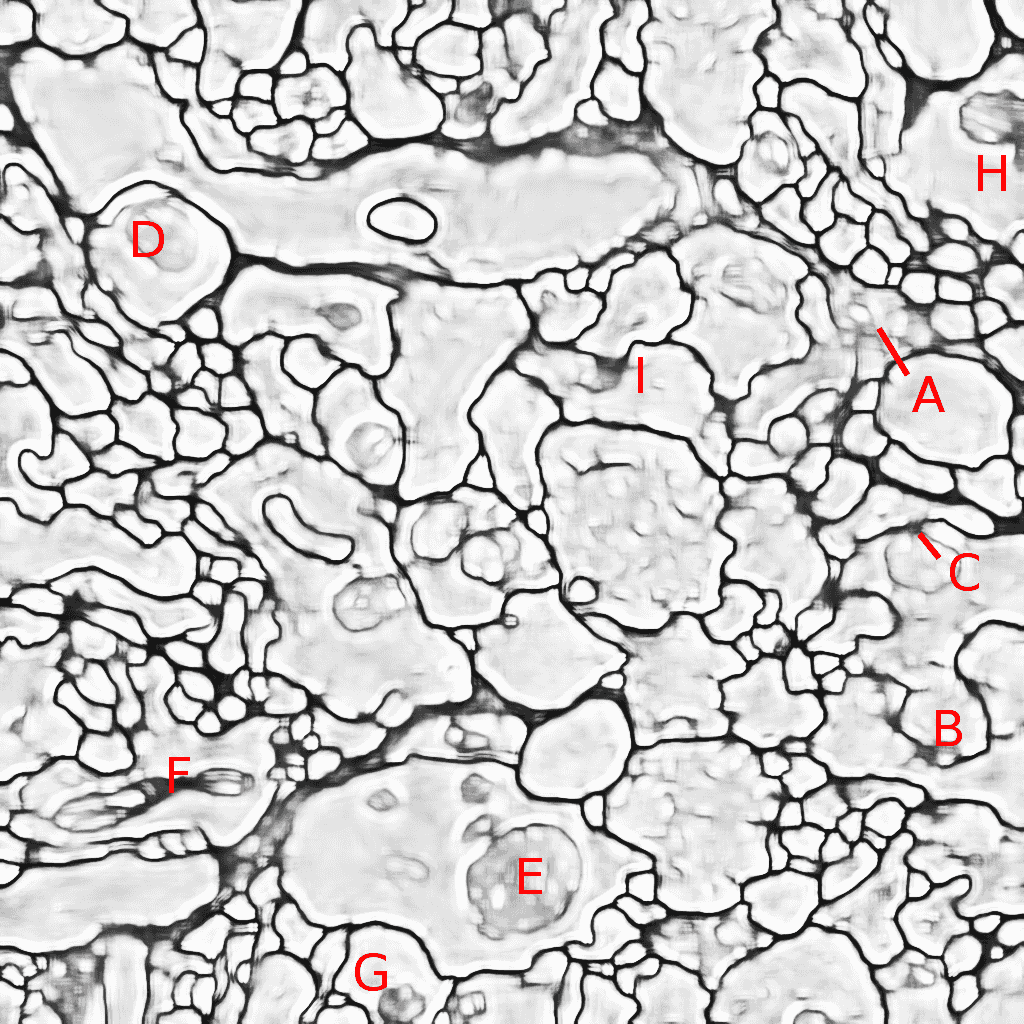}
		\caption{SK network, Malis.}
	\end{subfigure}%
	\newline
	\begin{subfigure}{0.5\textwidth}
		\center
		\includegraphics[scale=0.18]{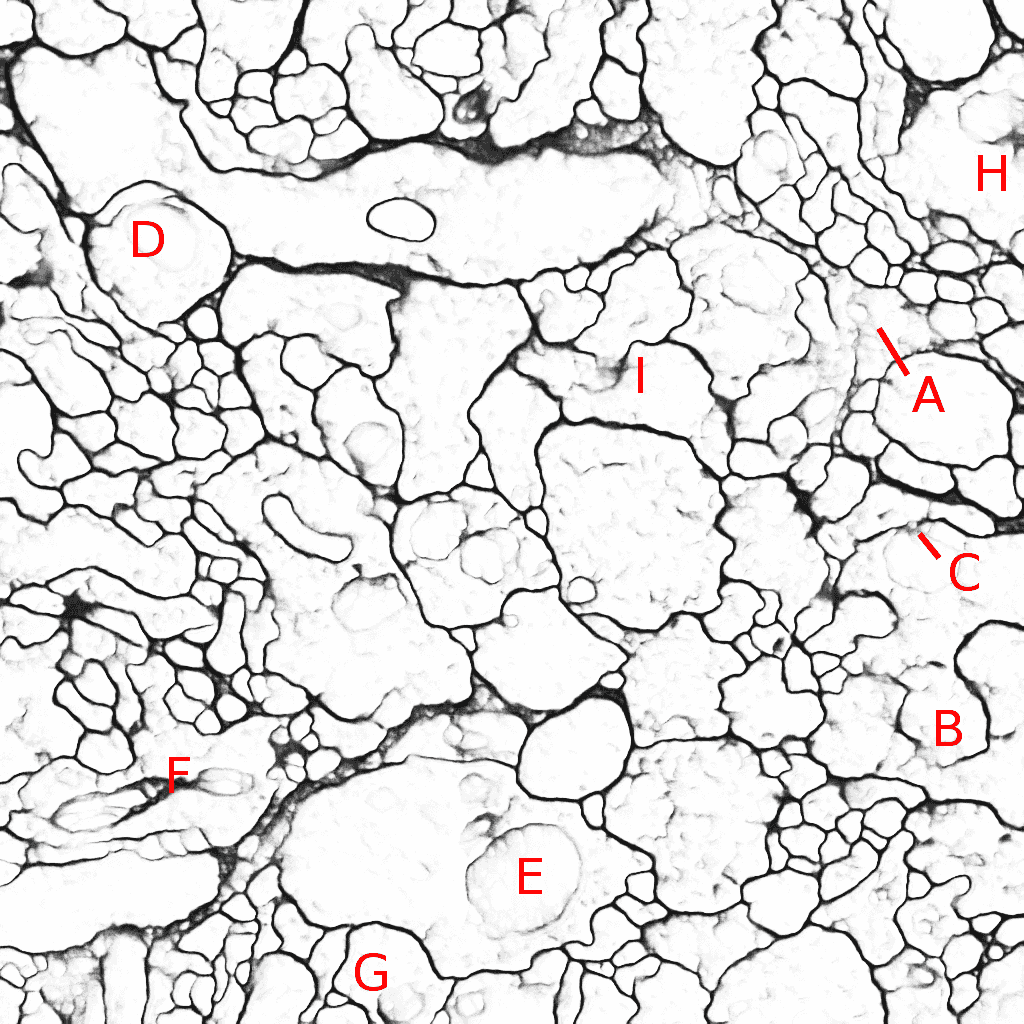}
		\caption{U network, Malis.}
	\end{subfigure}%
	\begin{subfigure}{.5\textwidth}
		\center
		\includegraphics[scale=0.18]{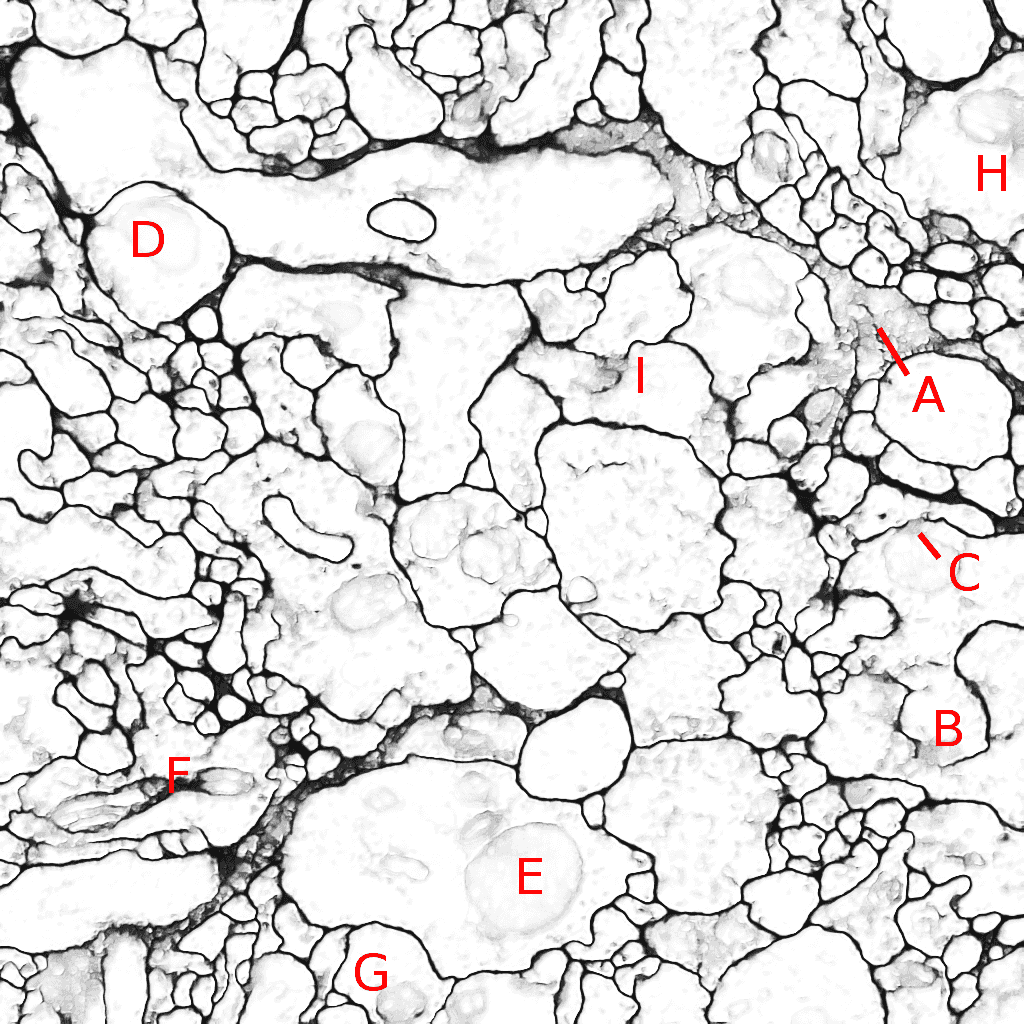}
		\caption{USK network, Malis.}
	\end{subfigure}%
	\newline
	\caption{Visual comparison of all results in Table \ref{tab:ds1ranking}.}
\end{figure}

\begin{figure}[H]
	\ContinuedFloat
	\begin{subfigure}{0.5\textwidth}
		\center
		\includegraphics[scale=0.18]{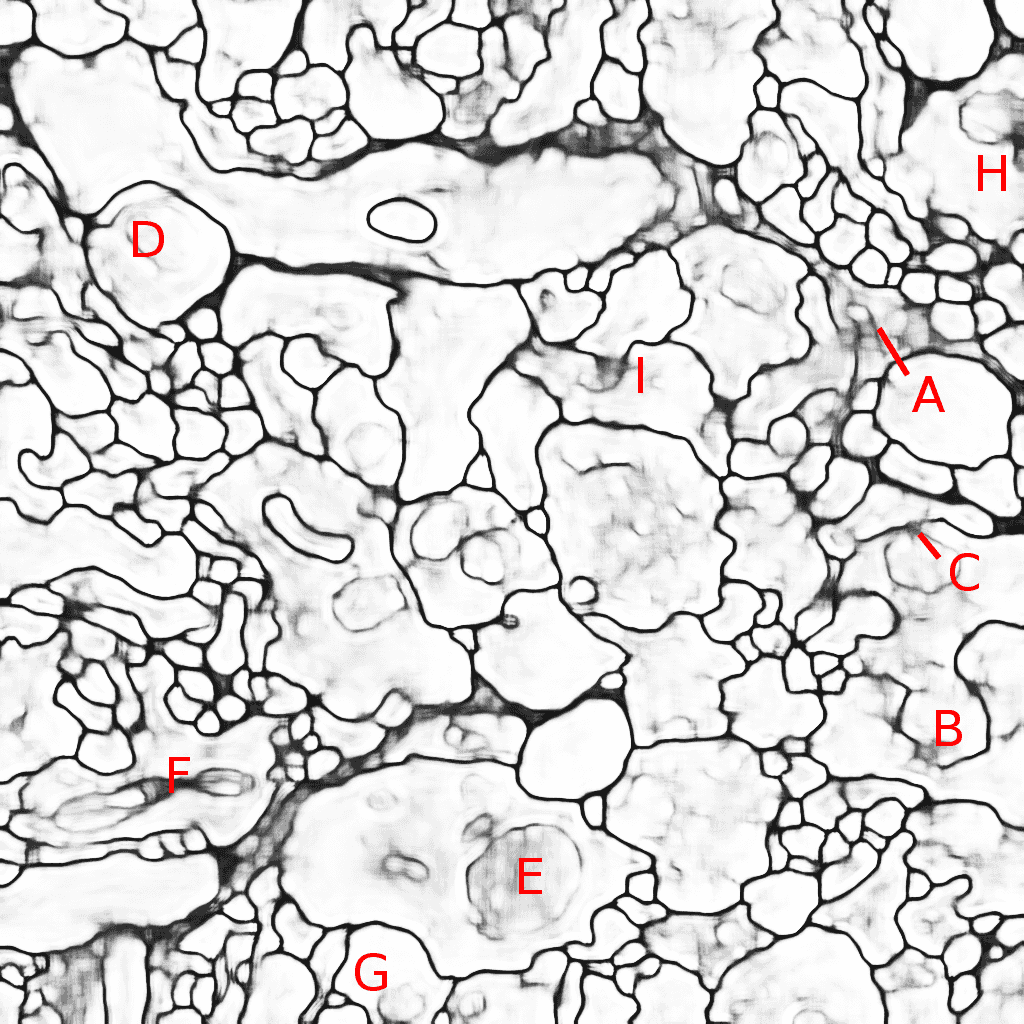}
		\caption{SK network, Softmax + Malis.}
	\end{subfigure}%
	\begin{subfigure}{.5\textwidth}
		\center
		\includegraphics[scale=0.18]{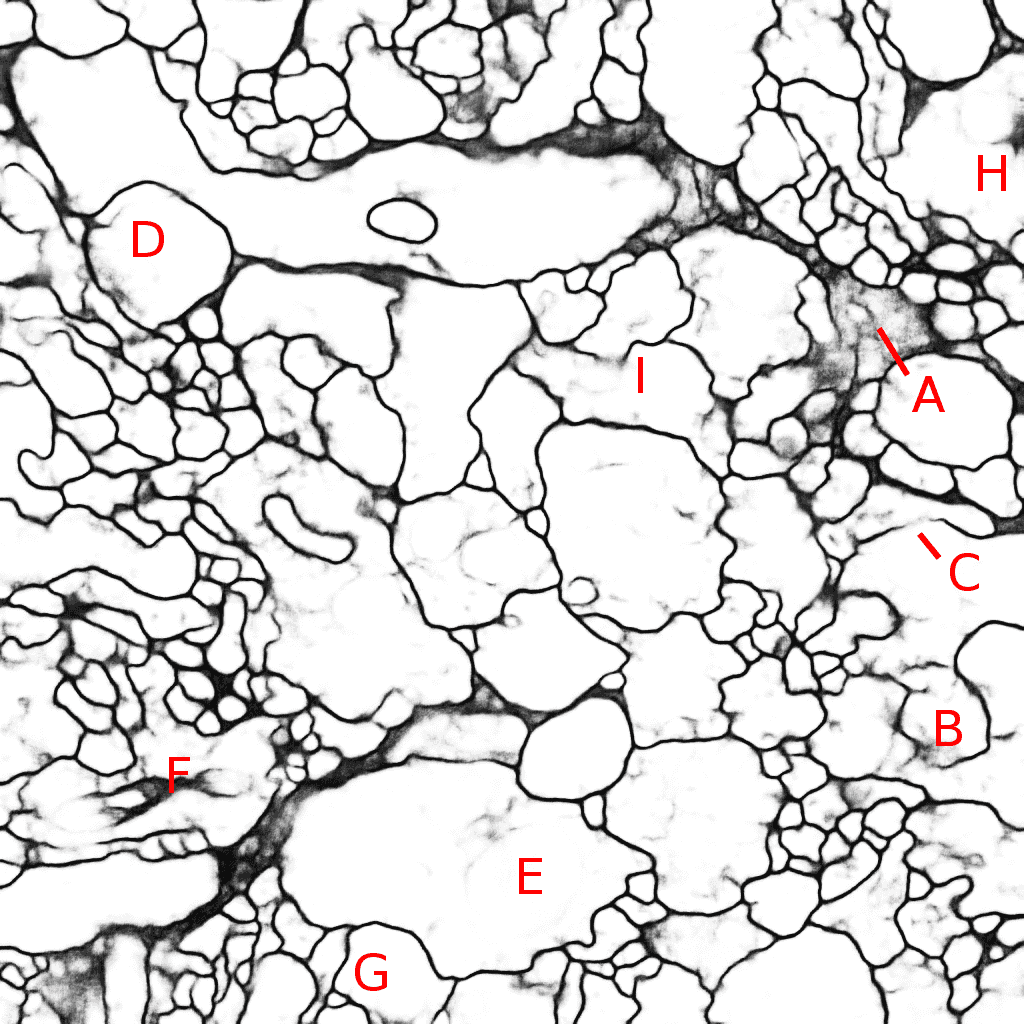}
		\caption{U network, Softmax + Malis.}
	\end{subfigure}%
	\pagebreak
	\newline
	\begin{subfigure}{0.5\textwidth}
		\center
		\includegraphics[scale=0.18]{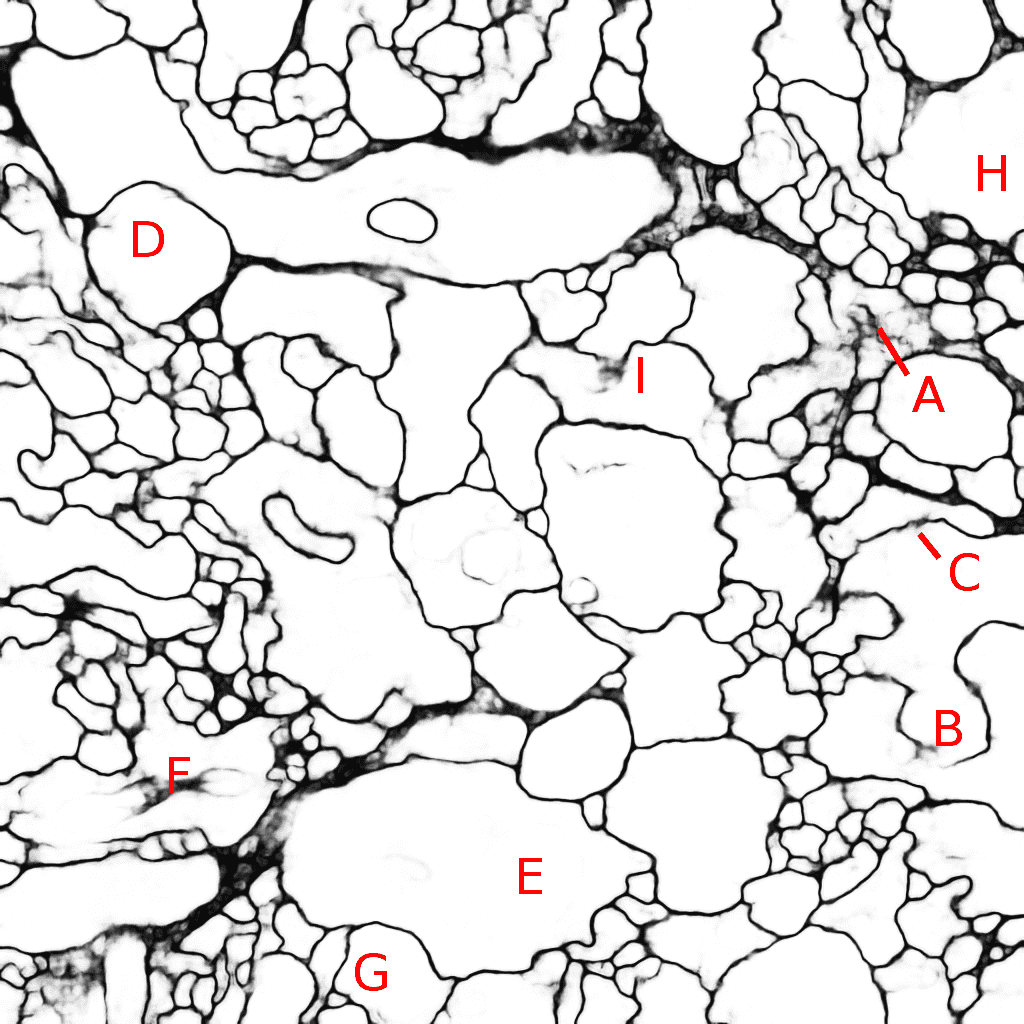}
		\caption{USK network, Softmax + Malis.}
	\end{subfigure}%
	\caption{Visual comparison of all results in Table \ref{tab:ds1ranking}.}
	\label{fig:ds1visualresults}
\end{figure}
Explanation to the labels in Figure \ref{fig:ds1visualresults}:
\begin{enumerate}[label=(\Alph*)]
	\item Glia cell that should be considered background. Malis loss does focus on separating foreground objects, so this part gets only labeled correctly in the trainings that use Softmax only (a, b and c).
	\item Diffuse membrane with light texture, which is hard to separate from cell interior. The separation is correct only in SK + Malis (d), while all other configurations lead to connected cells, although all trainings with Malis (d to i) show uncertainty, which could be sufficient to segment correctly. Diffuse parts are always on the edge of being pushed to foreground or background, which makes the segmentation fragile.
	\item Membrane with close proximity to a mitochondrion. Except for U networks that have been trained with Softmax (b and h), this gets labeled correctly. The downsampling of U-Net is sub-optimal here when pixel error (without scaling) instead of foreground separation is the training objective.
	\item Same case as label C, but here the mitochondrion is labeled with high certainty by all networks, leading to mislabeling of the nearby membrane.
	\item Removing an isolated mitochondrion with high certainty. The USK network (c and i) performed best on this task. Malis trainings are usually worse, again because Malis does not focus on pixel accuracy.
	\item Mitochondrion with long, thin structure. It consequently leads to a wrong classification as membrane because of its shape. This is not an issue, as isolated misclassifications within a cell can be rejected on a higher level when reconstructing the connectome.
	\item Mitochondrion close to the image border. As the training and classification uses border mirroring to make up for the missing context, it can lead to more errors near the border, independent of the network architecture.
	\item Same case as label G. Softmax + Malis on USK-Net performed best on removing the mitochondria on E, F, G and H.
	\item Diffuse membrane with a dark texture. It is expected that this gets completely classified as membrane. However, this was not the case on all networks and trainings. The worst case applies on U-Net (b and h), where the cells are almost connected.
\end{enumerate}

As expected, membranes get labeled thinner with Malis as this error criterion is stopping to provide loss at membranes as soon as two cells are sufficiently separated (see Section \ref{sec:malisloss}). Even though the separation border is thin, training with Malis separates cells very well and scores best on both DS1 (Table \ref{tab:ds1ranking}) and DS2 (Table \ref{tab:ds2ranking}).

The visual results match the numerical evaluation. USK performs best on pixel error with Softmax and best on rand error with Softmax + Malis. However, when using Softmax, there is not a huge difference in pixel error and visual results between the networks. Malis does not work very well on SK networks (d), leaving a lot of uncertainty in the prediction. It can still be a good segmentation, as thresholding of gray scale values can be applied. The numeric evaluation \cite{fijiscript} does test different thresholds, which is why the score is rather good (see Table \ref{tab:ds1ranking}) despite the uncertainty.

\section{Analysis on DS2}
\label{sec:ds2analysis}

\subsection{Training}
Training on DS2 was similar to training on DS1. The main difference is that the patch prior was not used on both Softmax and Malis. Instead, the error masking was enabled when using Softmax, which gives thicker borders. This is motivated by having input images which are slightly more blurred and thus the cell membranes are less sharp and harder to distinguish from cell interior.

Here, the SK network already converged using 2000 training iterations of Softmax before switching over to Malis. Starting with Malis directly was also not possible.

\subsection{Numerical}

\begin{table}[H]
\begin{center}
	\begin{tabular}{|l|l|l|r|r|r|}
		\hline
		\textbf{Rank} & \textbf{Network} & \textbf{Loss Function} & \textbf{Rand} & \textbf{Warping} & \textbf{Pixel}\\\hline
		1. & SK & Softmax + Malis & \textbf{0.060110507} & \textbf{0.000495529} & 0.106053779\\\hline
		2. & USK & Malis & 0.085927407 & 0.000848007 & 0.110390552\\\hline
		3. & SK & Malis & 0.086975487 & 0.000572968 & 0.107365432\\\hline
		4. & SK & Softmax & 0.087380844 & 0.000585556 & 0.075981492\\\hline
		5. & U & Softmax + Malis & 0.097356122 & 0.000940704 & 0.101856259\\\hline
		6. & USK & Softmax & 0.102450251 & 0.000851440 & \textbf{0.073163943}\\\hline
		7. & U & Malis & 0.121984927 & 0.001038742 & 0.111951817\\\hline
		8. & USK & Softmax + Malis & 0.128440534 & 0.000858688 & 0.101919859 \\\hline
		9. & U & Softmax & 0.148045042 & 0.001293564 & 0.083922396\\\hline
	\end{tabular}
	\captionof{table}{DS2 error evaluation (lower is better).}
	\label{tab:ds2ranking}
\end{center}
\end{table}

Interestingly, the USK-Net with Softmax + Malis loss training performs unexpectedly worse on the dataset DS2 than on DS1, where it performed best.
What is common for both DS1 and DS2 is that the USK network combined with Softmax training performs best on the pixel error.
Overall, this data set has a ranking much harder to explain than on DS1. Visual inspection reveals that the USK-Net was often overconfident when labeling the cell interior, which connected cells that should be separated. On the DS1 data set, this did not happen.

\begin{table}[H]
\begin{center}
	\begin{tabular}{|l|r|r|r|}
		\hline
		\textbf{Network} & \textbf{Rand} & \textbf{Warping} & \textbf{Pixel}\\\hline
		U & 0.0382 & 0.000353 & 0.0611\\\hline
	\end{tabular}
	\captionof{table}{(Source: Ronneberger \etal \cite{2015arXiv150504597R}).}
	\label{tab:unetres}
\end{center}
\end{table}
Finally, the results of U-Net obtained by Ronneberger \etal \cite{2015arXiv150504597R} (Table \ref{tab:unetres}) could not be reached, probably due to having used less transformations to extend the training data. It is not inherently clear if the U network would perform better than SK and USK given the bigger training data set.

Ideas to improve training include:
\begin{itemize}
	\item Weight maps to scale the loss instead of only masking it.
	\item Add elastic deformations, shifting and scaling instead of only rotation and mirroring to increase the amount of training data.
	\item Experiment with different loss functions than Malis and Softmax, or alternate between them during training.
\end{itemize}

\subsection{Visual}

The visual analysis is based on the image number 16 of the DS2 test stack (see Section \ref{sec:dataset2}).

\begin{figure}[H]
	\center
	\includegraphics[scale=0.37]{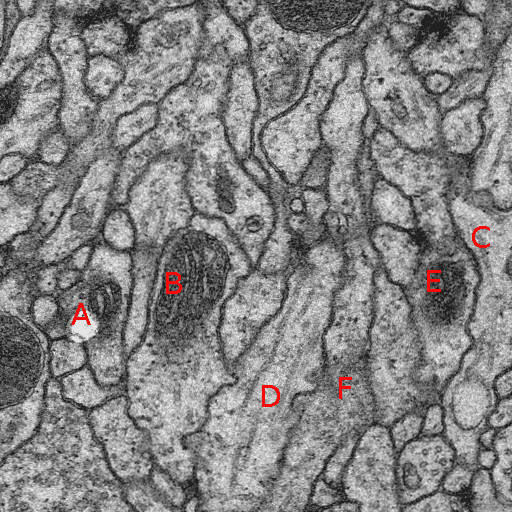}
	\caption{DS1 raw image, 512 by 512 pixels (Source: ISBI challenge \cite{ISBI2012}, \cite{CaffeNeuralModels}).}
	\label{fig:ds2araw}
\end{figure}

Explanation to the labels in Figure \ref{fig:ds1visualresults}:

The visual results of Softmax training (a to c) have thicker membrane labels than on DS1. This is because the error masking was enabled here and, as a result, the network has seen the same amount of error pixels for both foreground and background.

\begin{enumerate}[label=(\Alph*)]
	\item A bright spot, which is an error in the electron microscopy image. The membrane affected by it is misclassified by all networks and trainings. The local contrast enhancement with CLAHE did not help here. 
	\item Mitochondrion with close proximity parallel to a cell membrane. The USK network removes the membrane with all trainings (c, f and i). U-Net (b, e and h) performs a little better, but still merges the two cells. Only the SK network does it correctly (a, d and g), but has some uncertainty on the mitochondrion instead (d).
	\item Diffuse section of a cell membrane. This is not an issue on most training/architecture combinations, except for U-Net with Softmax (b and h).
	\item Oriented structures, even when faint, are partially labeled as membrane when sharp enough and of similar thickness as the membrane. This is no issue when isolated within the cell and not cutting through a cell that should be connected. With convolutional networks, this is hard to impossible to label correctly. It would require more high level knowledge of the object, such as if the predicted membrane is enclosing a cell or not.
\end{enumerate}

\begin{figure}[H]
	\center
	\begin{subfigure}{0.5\textwidth}
		\center
		\includegraphics[scale=0.37]{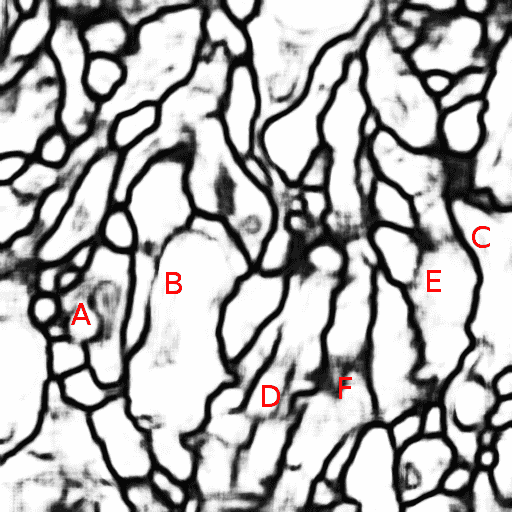}
		\caption{SK network, Softmax.}
	\end{subfigure}%
	\begin{subfigure}{.5\textwidth}
		\center
		\includegraphics[scale=0.37]{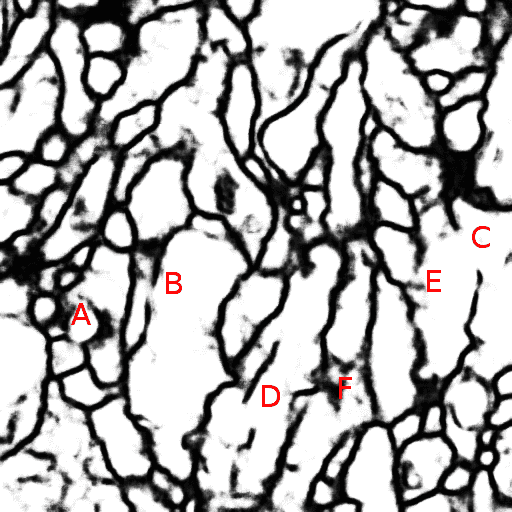}
		\caption{U network, Softmax.}
	\end{subfigure}%
	\newline
	\begin{subfigure}{0.5\textwidth}
		\center
		\includegraphics[scale=0.37]{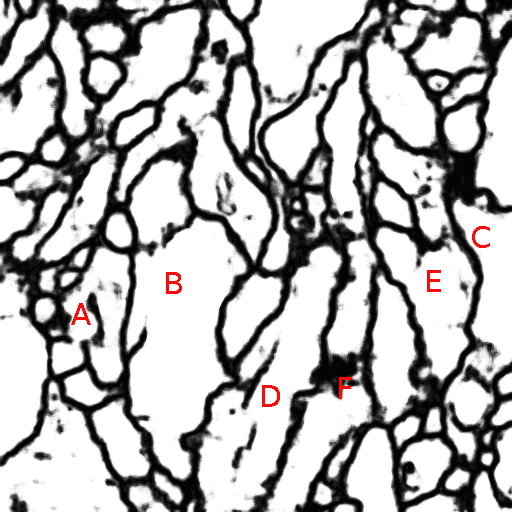}
		\caption{USK network, Softmax.}
	\end{subfigure}%
	\begin{subfigure}{.5\textwidth}
		\center
		\includegraphics[scale=0.37]{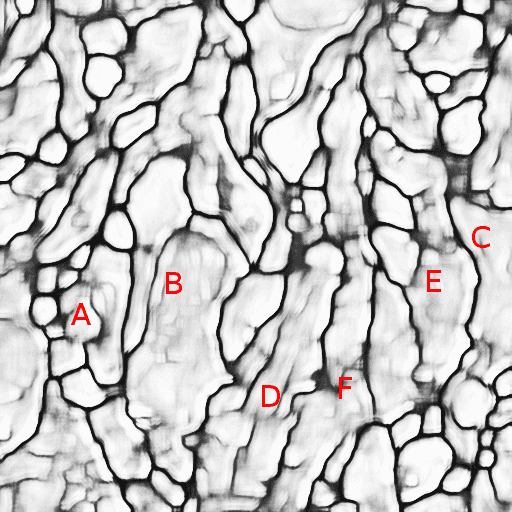}
		\caption{SK network, Malis.}
	\end{subfigure}%
	\newline
	\begin{subfigure}{0.5\textwidth}
		\center
		\includegraphics[scale=0.37]{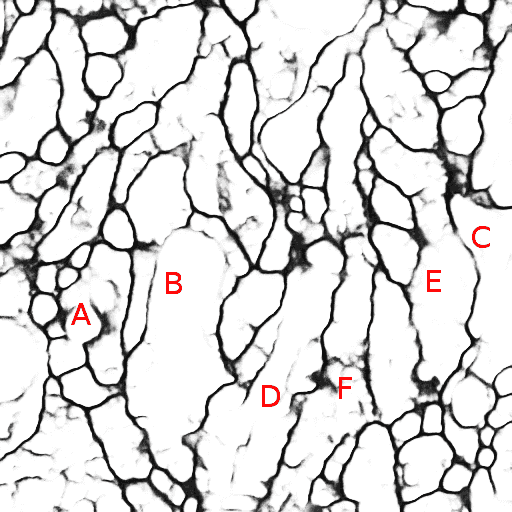}
		\caption{U network, Malis.}
	\end{subfigure}%
	\begin{subfigure}{.5\textwidth}
		\center
		\includegraphics[scale=0.37]{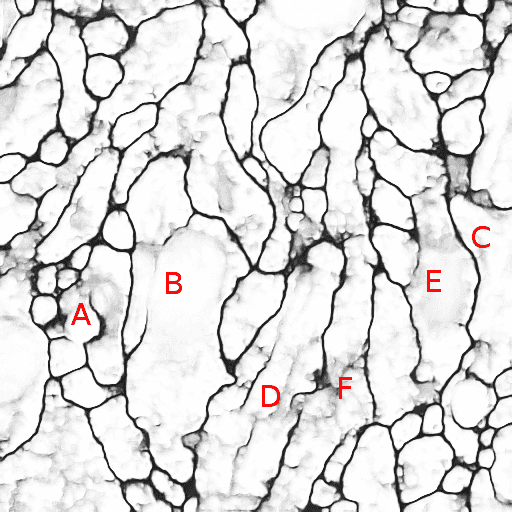}
		\caption{USK network, Malis.}
	\end{subfigure}%
	\newline
	\caption{Visual comparison of all results in Table \ref{tab:ds2ranking}.}
\end{figure}

\begin{figure}[H]
	\ContinuedFloat
	\begin{subfigure}{0.5\textwidth}
		\center
		\includegraphics[scale=0.37]{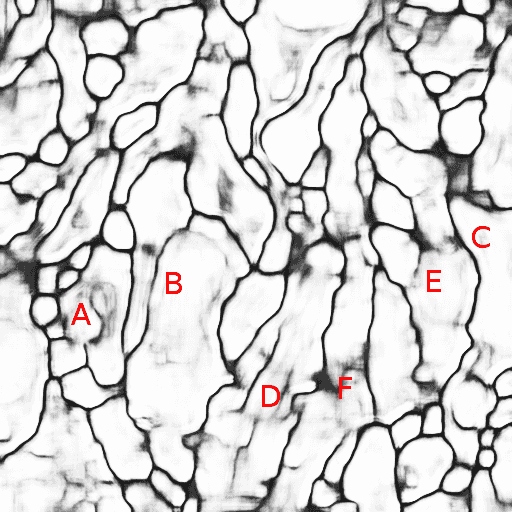}
		\caption{SK network, Softmax + Malis.}
	\end{subfigure}%
	\begin{subfigure}{.5\textwidth}
		\center
		\includegraphics[scale=0.37]{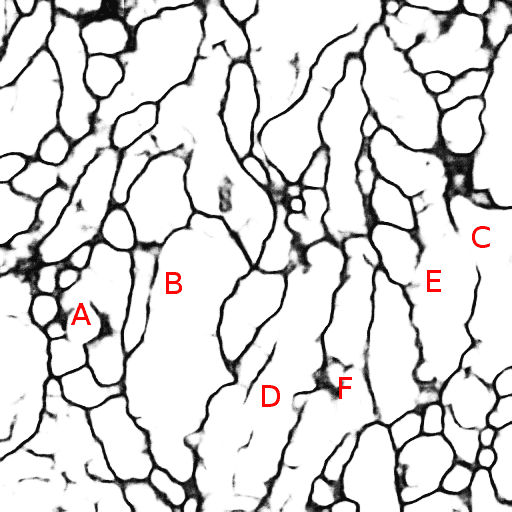}
		\caption{U network, Softmax + Malis.}
	\end{subfigure}%
	\pagebreak
	\newline
	\begin{subfigure}{0.5\textwidth}
		\center
		\includegraphics[scale=0.37]{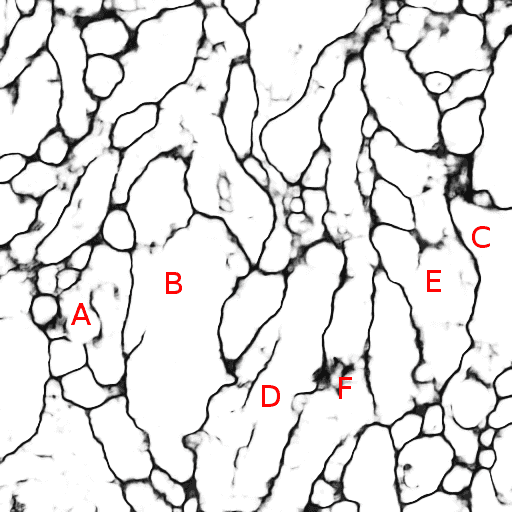}
		\caption{USK network, Softmax + Malis.}
	\end{subfigure}%
	\caption{Visual comparison of all results in Table \ref{tab:ds2ranking}.}
\end{figure}

\begin{enumerate}[label=(\Alph*), resume]
	\item Diffuse mitochondrion. The same situation as with B applies. Especially U- and USK-Net with Softmax training (b and c) get it wrong.
	\item Diffuse cell interior that is similar to the membrane texture. All networks see a membrane connection through this area. In the training data there are some examples of diffuse membranes, so the networks have slightly overfitted on the training data for this case.
\end{enumerate}

\addtocontents{toc}{\protect\newpage}
\chapter{Conclusion}
\section{Research Time Line}
An overview of the research time line, in order to give a context on what shaped the objectives and decisions made during the project:\\\newline
\begin{tabular}{|l|l|l|}
	\hline
	\textbf{From} & \textbf{To} & \textbf{Activity / Event}\\\hline
	05.11.2014 & 05.11.2014 & Collaboration request at UZH INI.\\\hline
	09.11.2014 & 22.01.2014 & Discussing ideas with Dr. Jan Funke.\\\hline
	14.12.2014 & 14.12.2014 & Hongsheng Li \etal paper released (SK kernels) \cite{2014arXiv1412.4526L}.\\\hline
	07.02.2015 & 07.02.2015 & Research proposal finished and accepted.\\\hline
	23.02.2015 & 23.02.2015 & Research beginning.\\\hline
	26.02.2015 & 06.03.2015 & Getting the sliding window network to work \cite{JulienMartel}.\\\hline
	08.03.2015 & 18.04.2015 & OpenCL backend development \cite{Caffe}.\\\hline
	10.04.2015 & 21.04.2015 & Discussing the project with AMD \cite{AMD}.\\\hline
	22.04.2015 & 22.04.2015 & Arrival of AMD's W9100 GPUs (hardware sponsoring).\\\hline
	19.04.2015 & 19.04.2015 & Pull request of the modified Caffe to BVLC \cite{jia2014caffe}.\\\hline
	09.05.2015 & 09.05.2015 & Public release of the Caffe Neural Models \cite{CaffeNeuralModels}.\\\hline
	09.05.2015 & 09.05.2015 & Public release of the Caffe Neural Tool \cite{CaffeNeuralTool}.\\\hline
	15.05.2015 & 15.05.2015 & Ronneberger \etal paper released (U-Net) \cite{2015arXiv150504597R}.\\\hline
	20.05.2015 & 25.06.2015 & Testing of U-Net and design of USK-Net.\\\hline
	27.06.2015 & 12.07.2015 & Collaboration at HHMI Janelia, Virginia, USA \cite{HHMIJ}.\\\hline
	29.06.2015 & 14.07.2015 & Implementing Malis loss and N-D SK kernels for Caffe.\\\hline
	12.07.2015 & 15.07.2015 & Critical source code development finished.\\\hline
	13.07.2015 & 20.08.2015 & Writing the report and final evaluation experiments.\\\hline
	24.08.2015 & - & Post-research support of \textit{Project Greentea} and ongoing\\
			   & & development in collaboration with AMD,\\
			   & & HHMI Janelia and the Caffe community.\\\hline
\end{tabular}

\newpage
\section{Implications}

The first idea for the research project was to implement strided kernels. However, with the release of the Hongsheng Li \etal \cite{2014arXiv1412.4526L} paper, the problem already got solved.
We got their source code and I was able to translate existing sliding window networks to strided kernel networks.\\
These events lead to a shift of focus to implement the OpenCL backend and support a variety of hardware. This was important to see how existing CPU clusters and AMD GPUs could be used instead of only nVidia GPUs.

A nice side effect of completely re-writing the whole Caffe library to OpenCL was gaining a complete understanding of the library, the bottlenecks, how all layers work and what the most important objectives for optimizations and network design are.

It turned out that running networks across devices does not give an advantage in the case of SK, U and USK architectures, as perfect scaling is possible when running independent instances of the network on each device. This only requires from the devices to have enough memory to hold the networks. This assumption was met when AMD's W9100 GPUs became available to me.

SK networks did not scale as desired and up to 100'000 pixel classifications per second were only about 1/10th of the desired speed. The original ideas to speed up the layers of the SK network by using methods such as multi device execution, Fourier transform convolutions or direct convolutions did not work.\\
With the release of the Ronneberger \etal paper \cite{2015arXiv150504597R}, the research focus was shifted to analyzing the U-Net approach, which is able to classify up to one megapixel per second.\\
Training of U-Net was more difficult than SK-Net and thus I tried to implement my own network architecture based on the findings of SK-Net and U-Net, which resulted in the experimental USK-Net. The USK-Net performs similarly to U-Net and produced better results with small training data sets (see Chapter \ref{ch:results}).

Looking at ISBI 2012 results \cite{ISBI2012} and their test metrics, as well as the fact that one of the authors of the Malis criterion \cite{2009arXiv0911.5372T}, Dr. Srinivas Turaga, was at HHMI Janelia for collaboration, lead to the development of an additional loss layer for Caffe (see Section \ref{sec:malisloss}).

Finally, the last feature implemented before freezing the source code was N dimensional strided kernel support for max pooling and convolution layers, as this was a feature requested by Dr. Stephan Saalfeld and Dr. Srinivas Turaga at HHMI Janelia. This can be used to run modified SK, U and USK network architectures on 3D blocks of volumetric-isotropic data sets, or even 3D over time (4D).

\section{Difficulties Encountered}
The obvious difficulty was to keep up with the general research in pixelwise classification of images, as important papers \cite{2014arXiv1412.4526L}, \cite{2015arXiv150504597R} were released during the project research. A shift of focus from the original plans were required a few times. This includes taking into account new results and dropping planned approaches.

It was also a lot of work to keep up with the changes of the Caffe library \cite{jia2014caffe}, as they changed many core aspects such as network file format and shape specifications for memory blobs. This broke compability with existing code from Honghsheng Li \etal's approach \cite{2014arXiv1412.4526L} as well as the existing sliding window network \cite{JulienMartel}, \cite{CaffeNeuralModels}. Constantly pulling new changes from the BVLC master branch \cite{BVLCCaffe} and adapting my own branch to those changes was necessary. The benefit gained by doing this is that backwards compatibility to the official version is always guaranteed and that my own branch was ahead during the whole scope of the project.

With programming the Caffe Neural Tool, the diversity of formats for labels and input data was complex to handle. Especially loading and storing \textit{TIF} pictures that can have a variety of pixel formats and support stacking multiple images in a single file can be tricky.

At last, it was not always obvious why a network does learn the expected features or not. Training parameter tests require up to ten hours of training on a GPU, which is very acceptable during production, but rather cumbersome during debugging. Evaluation and training of the networks for numeric results was only possible after freezing critical parts of the source code (computation kernels and layer implementations), because the results can differ after fixing bugs and other changes of the library. This resulted in having only two weeks left for this stage.

\section{Reproducibility of Results}
The results obtained in this report are guaranteed to be reproducible by the use of the following software pipeline, using CUDA or OpenCL hardware equivalent to the hardware used in this report.

Repositories belonging to \textit{Project Greentea}:
\begin{itemize}
	\item Caffe \cite{Caffe}
	\subitem URL: \url{https://github.com/naibaf7/caffe}
	\subitem Commit checksum: \texttt{f84c2a4fb8d633bc7d8fc9771eb06a3cf2215212}
	\item Caffe Neural Tool \cite{CaffeNeuralTool}
	\subitem URL: \url{https://github.com/naibaf7/caffe_neural_tool}
	\subitem Commit checksum: \texttt{780e7dd72e4f88c80729e2b33d6c0137d479016a}
	\item Caffe Neural Models \cite{CaffeNeuralModels}
	\subitem URL: \url{https://github.com/naibaf7/caffe_neural_models}
	\subitem Commit checksum: \texttt{dbb06d8352aa9c2ba99458cb9e9068500ebacc11}
\end{itemize}

As long as the ISBI 2012 challenge is ongoing, the data set DS2 results can be reproduced on their website \cite{ISBI2012}.
The training and test data for the data set DS1 remains included in the Caffe Neural Models repository.

\section{Outlook}

In this outlook I give a brief introduction on future plans for the improved Caffe version \cite{Caffe}, extended use cases for the models \cite{CaffeNeuralModels}, missing features for the Caffe Neural Tool \cite{CaffeNeuralTool} and ideas that did not fit into the time scope of the project.

\subsection{Device Abstracted Backend}
Currently, the OpenCL, CUDA and (legacy) CPU backend are implemented side-by-side and there is quite some code duplication in the Caffe library. To support future multi-device training methods and remove redundancy, the backend should be further unified so that the only remaining code duplication resides with the actual compute kernels used inside the layers. This will minimize bugs that occur on only one backend and make software verification much easier to handle. It will also shorten the time required for newly developed layers to become available on all devices.\\
At the time of the project, the improved Caffe library \cite{Caffe} drops full support on the legacy CPU backend in favor of an OpenCL hybrid solution (see Section \ref{sec:openclhybrid}) on CPUs. Some tuning to work correctly on NUMA processors (see Section \ref{sec:numaissues}) is still required. The CPU backend remains as a fallback for layers that do not work on OpenCL and CUDA, such as the Malis loss criterion (see Section \ref{sec:malisloss}).

\subsection{Improving Training Data}
As a first step to improve results, all network architectures should be evaluated using more training data, acquired artificially or from more ground truth.\\
It is advisable to try out all models on a given data set, as there is no clear winner among the networks. Results may vary strongly as the numerical analysis revealed (see Section \ref{sec:ds1analysis} versus \ref{sec:ds2analysis}).

\subsection{Parameter Grid Search}
The network architectures presented here are only examples of a whole class of possible networks. Many parameters such as kernel sizes and how SK networks are combined with U type networks can be evaluated. Especially deep multi-path networks can be useful, merging the feature maps of different architectures before making the ouput label predictions. The new USK architecture is such an example.

\subsection{Testing of Volumetric Architectures}
Depending on the data set, SK, U and USK networks can be configured in many more ways for 3D than for 2D. For example, the depth direction is likely to have less physical resolution than the width and height dimension, due to how the data is acquired with slicing and electron microscopy. This should be considered when choosing the kernel size , kernel stride and span of the depth dimension.
An example is the ISBI 2012 dataset which spans 2 x 2 x 1.5 microns with a resolution of 4 by 4 by 50 nm/pixel \cite{TrackEM}, \cite{ISBI2012}, \cite{Cardona2010}.

\subsection{Improving Test Metrics}
Visual inspection and the error metrics used in this report can only give information about how accurate the label predictions are compared to the ground truth. For connectomics, this may not be the most important objective. The tools which will further process the segmentations may be able to correct certain errors in the predictions by testing how likely the final result is, while other merge and split errors lead to uncorrectable errors. This is an objective that could be more useful when selecting the network architecture, loss function and training method.

\section{Final Words}
This research project combines many disciplines, such as high performance computing, machine learning, visual computing and a bit of connectomics. I was able to implement most of the originally planned features and found replacements for ideas that turned out to work badly. Finally, the results of the research include a useful, versatile stack of Open Source software (\textit{Project Greentea}) that can be extended in the future. During working on the project, it was already possible to establish a growing user base \cite{CaffePR}. The models and tools introduced with this project can be used efficiently on a large variety of data sets and hardware, making it very flexible.
The collaboration at HHMI Janelia was also a great experience. Hardware sponsoring by AMD shows that programming of the \textit{Project Greentea} and efficient machine learning libraries in general is of high interest also for hardware manufacturers.
\appendix

\chapter{Network Architectures}
\label{app:net}
\section{SK-Net}
\label{app:sknet}
\begin{figure}[H]
	\centering
	\includegraphics[page=2,width=0.95\textwidth]{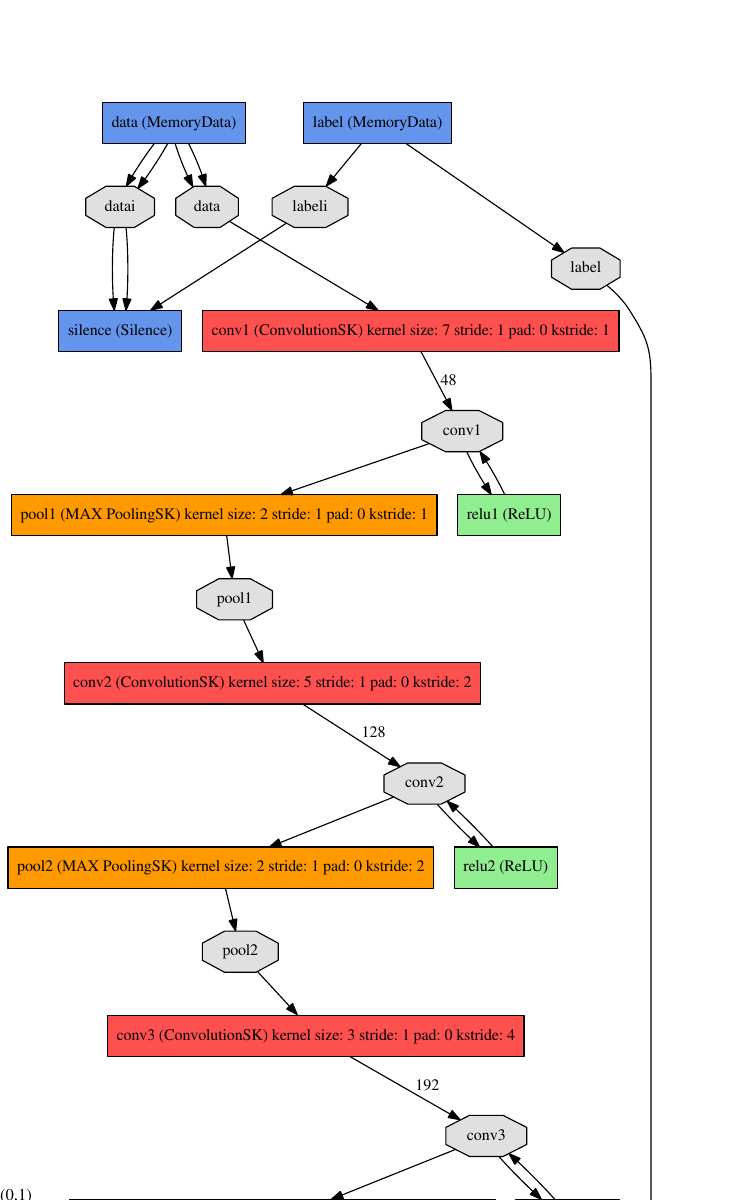}
\end{figure}
\begin{figure}[H]
	\centering
	\includegraphics[page=1,width=0.95\textwidth]{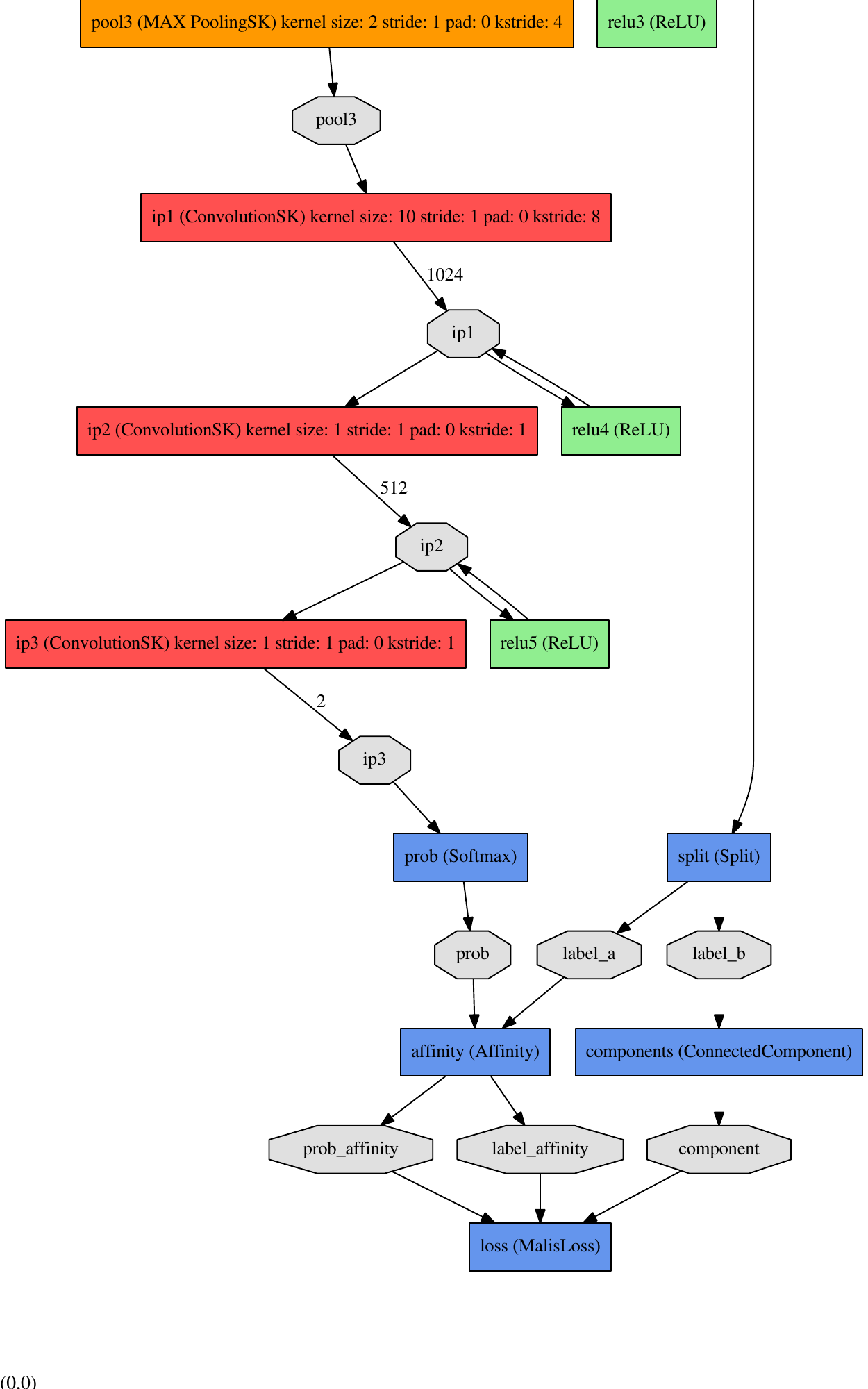}
\end{figure}
\section{U-Net}
\label{app:unet}
\begin{figure}[H]
	\centering
	\includegraphics[page=6,width=0.95\textwidth]{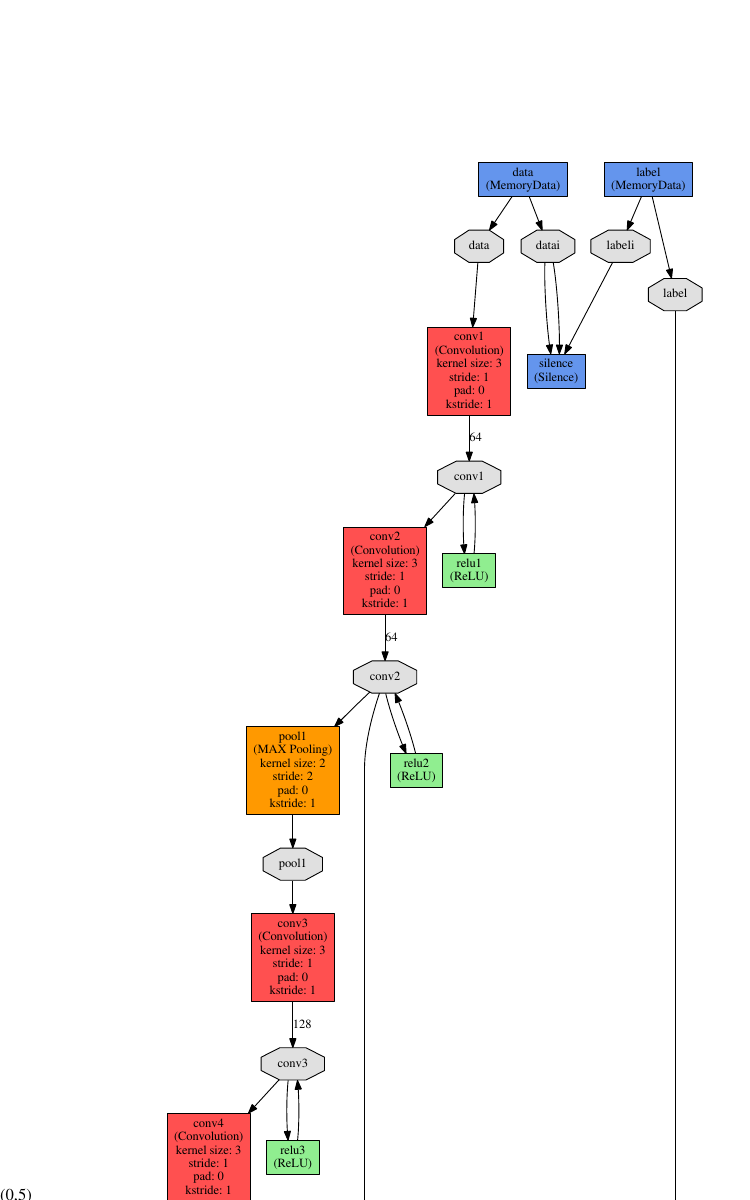}
\end{figure}
\begin{figure}[H]
	\centering
	\includegraphics[page=5,width=0.95\textwidth]{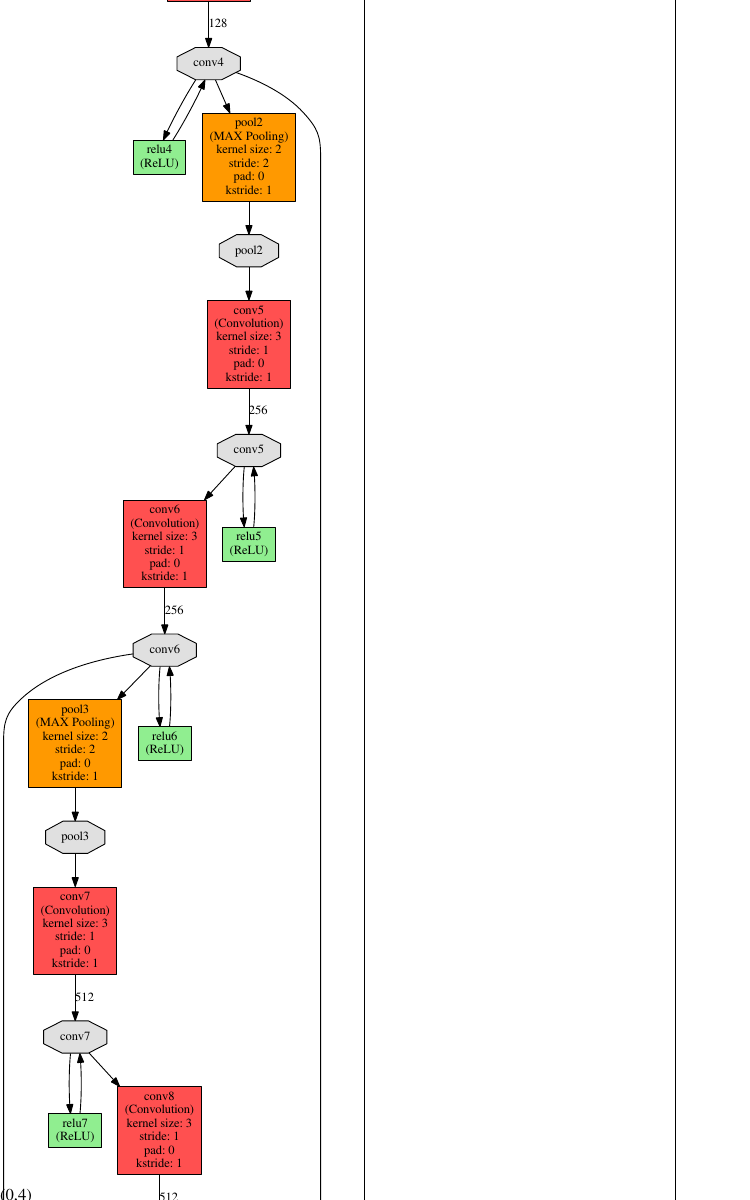}
\end{figure}
\begin{figure}[H]
	\centering
	\includegraphics[page=4,width=0.95\textwidth]{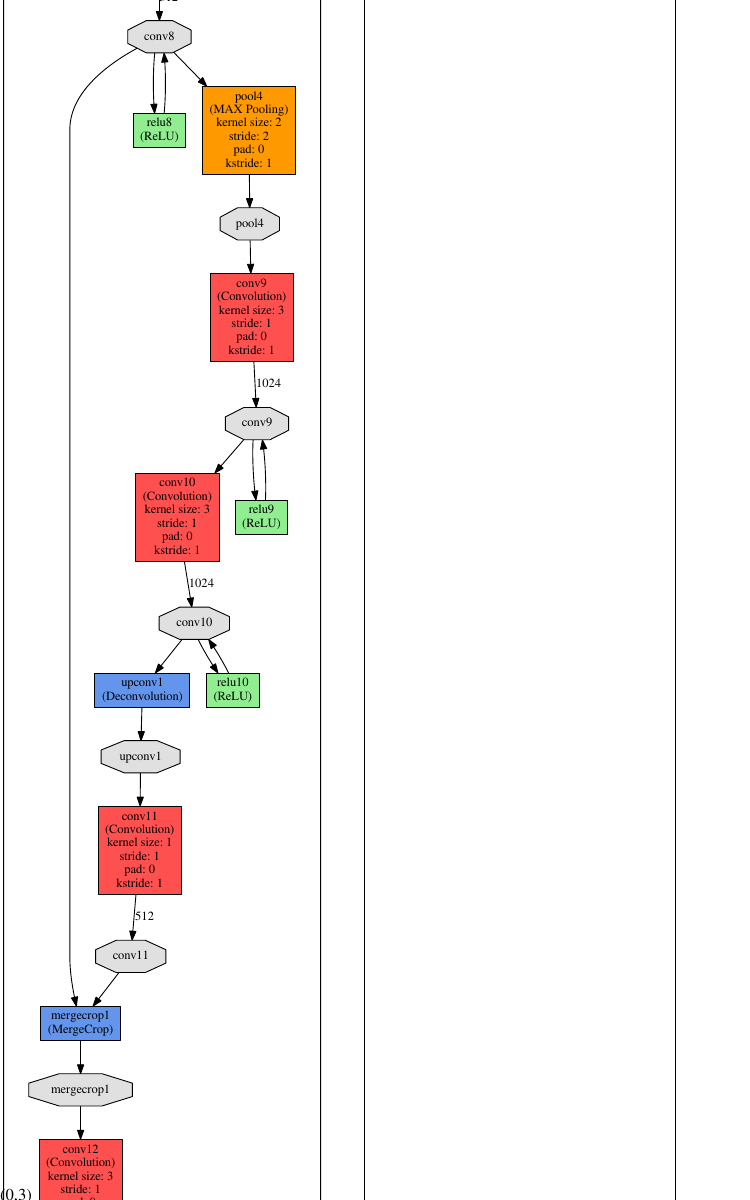}
\end{figure}
\begin{figure}[H]
	\centering
	\includegraphics[page=3,width=0.95\textwidth]{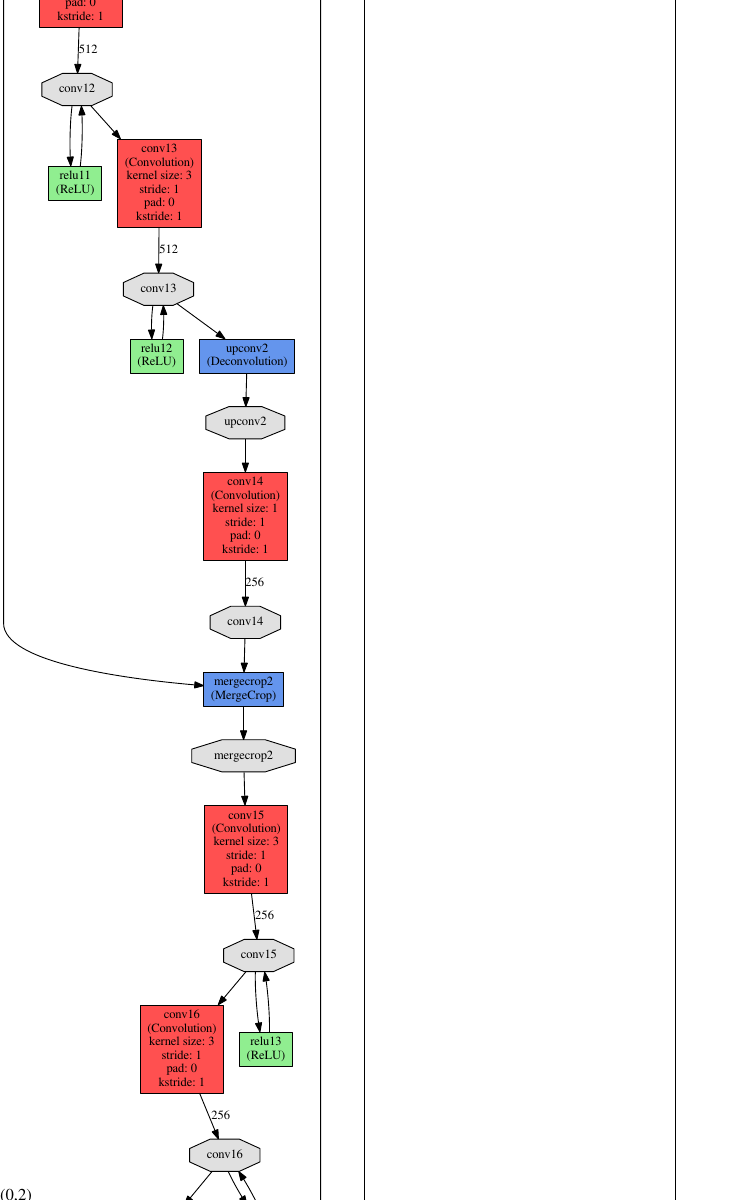}
\end{figure}
\begin{figure}[H]
	\centering
	\includegraphics[page=2,width=0.95\textwidth]{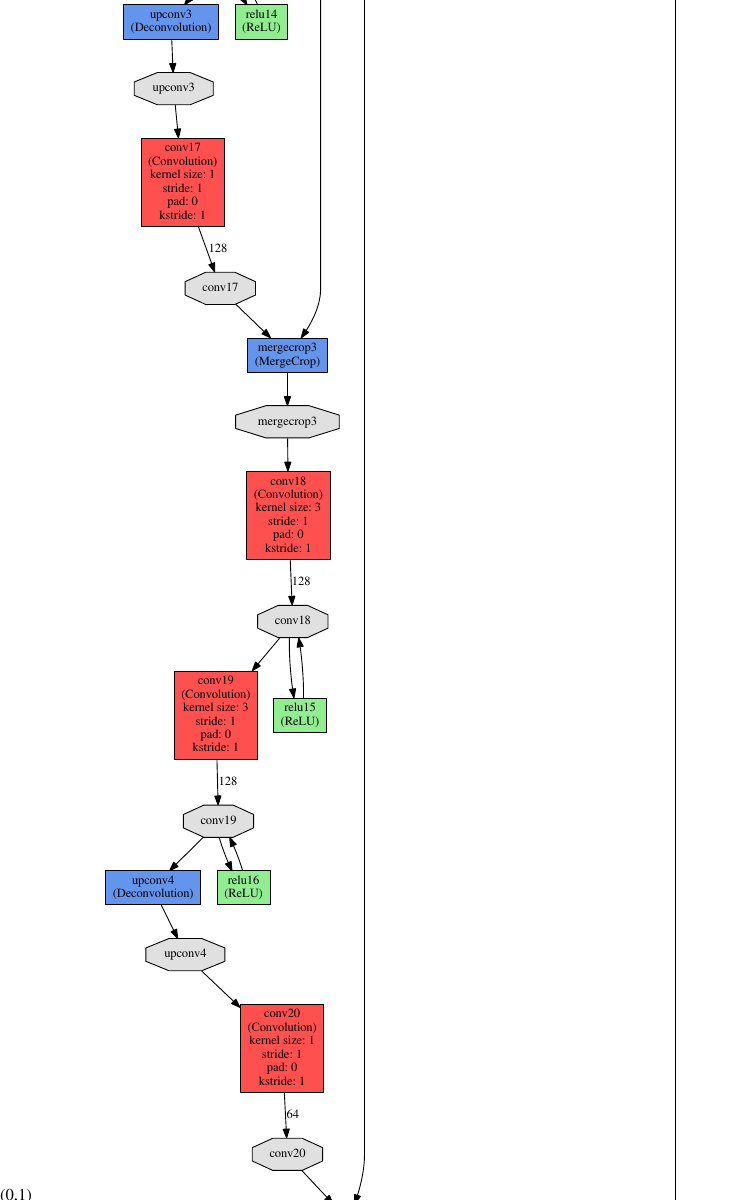}
\end{figure}
\begin{figure}[H]
	\centering
	\includegraphics[page=1,width=0.95\textwidth]{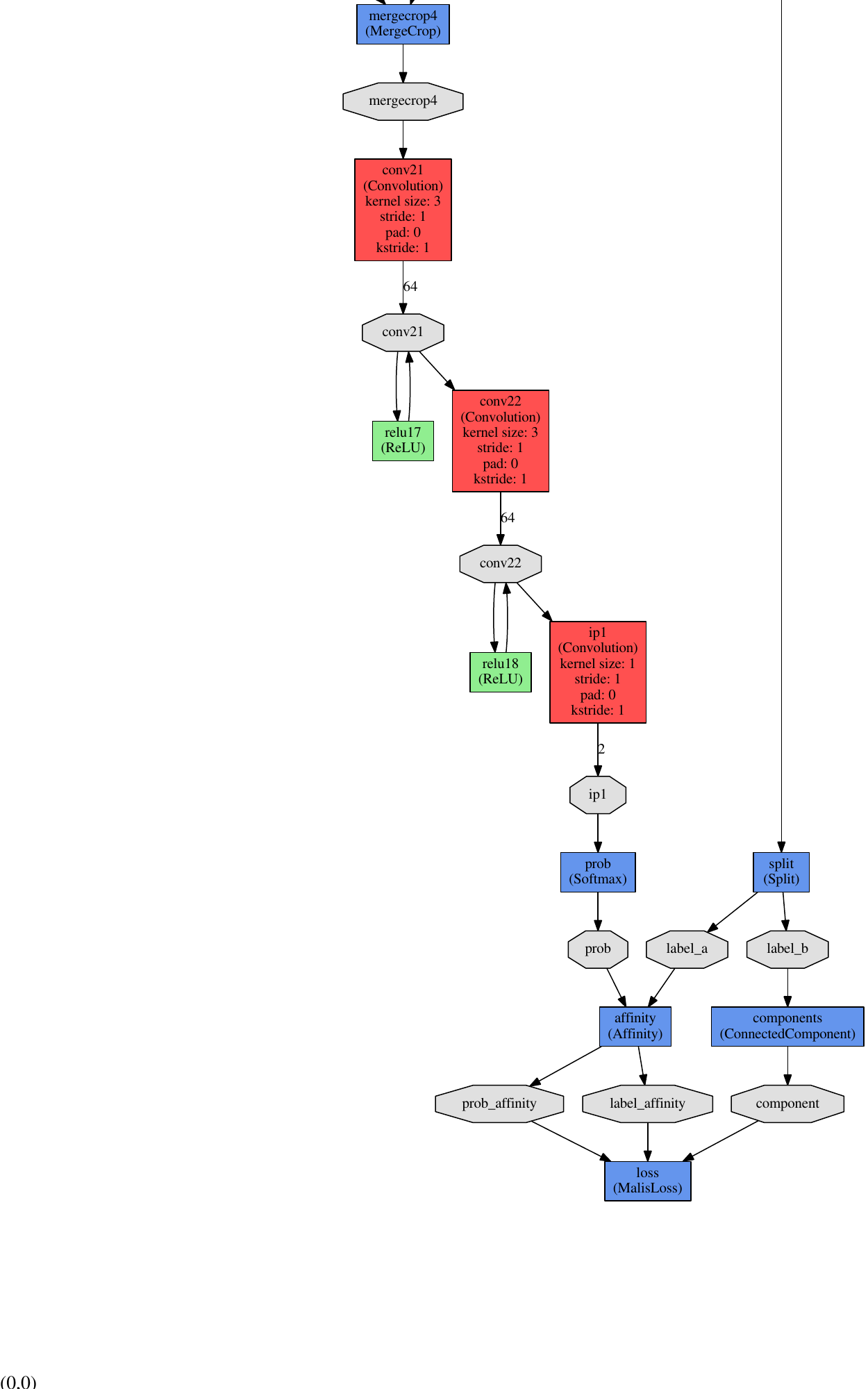}
\end{figure}
\section{USK-Net}
\label{app:usknet}
\begin{figure}[H]
	\centering
	\includegraphics[page=5,width=0.95\textwidth]{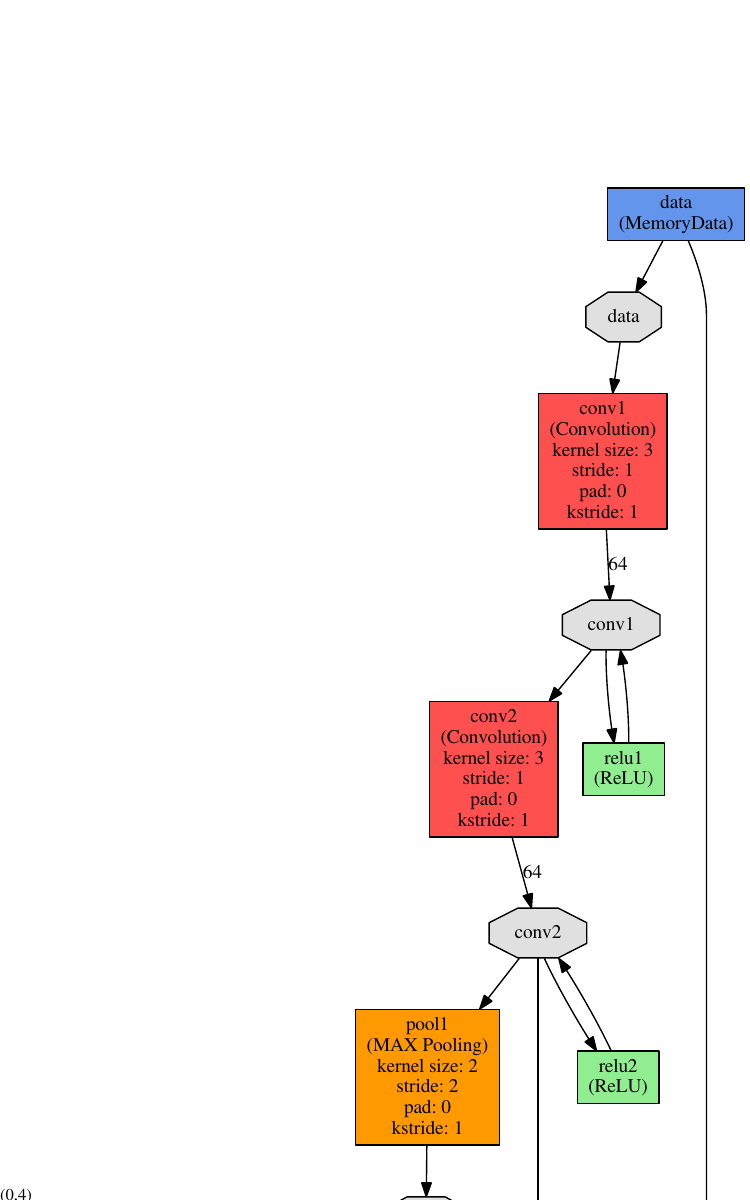}
\end{figure}
\begin{figure}[H]
	\centering
	\includegraphics[page=4,width=0.95\textwidth]{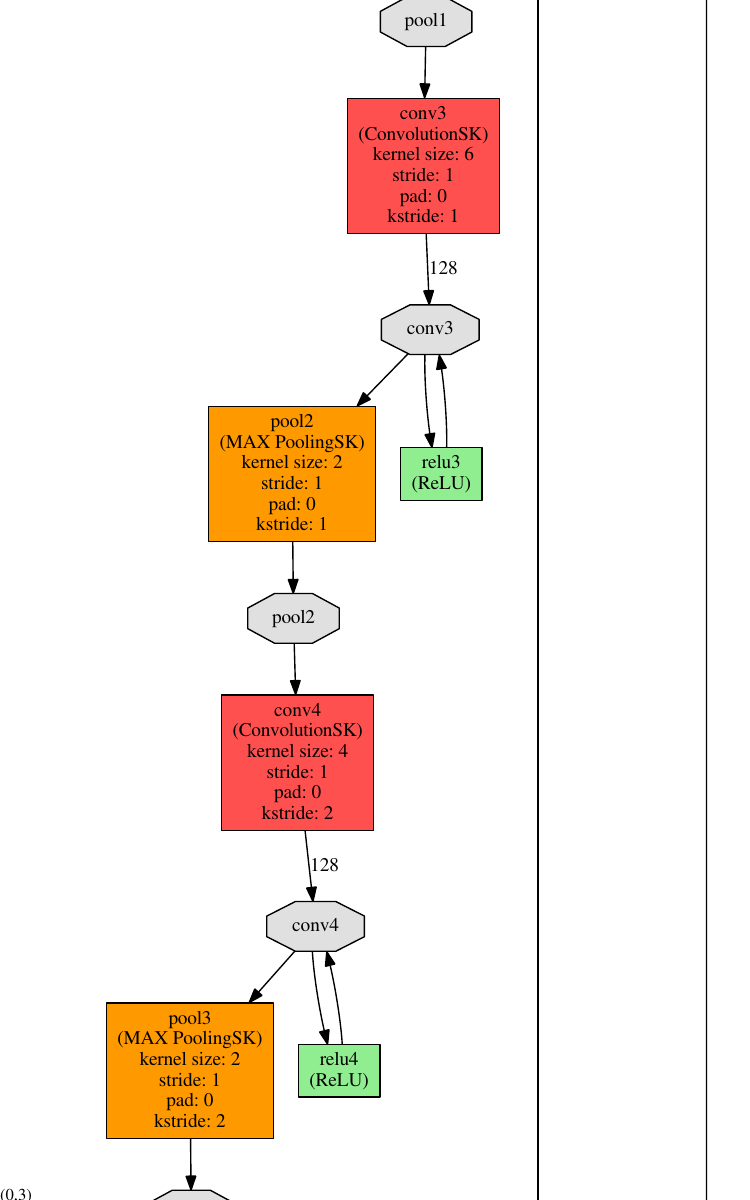}
\end{figure}
\begin{figure}[H]
	\centering
	\includegraphics[page=3,width=0.95\textwidth]{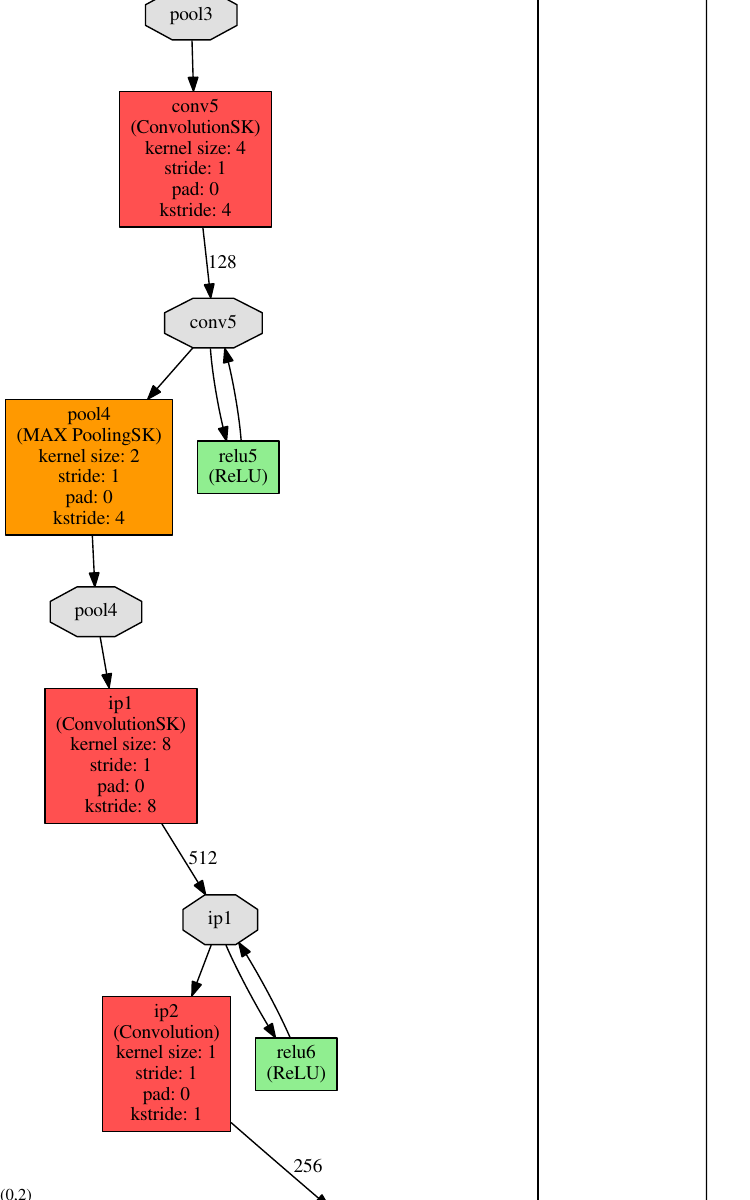}
\end{figure}
\begin{figure}[H]
	\centering
	\includegraphics[page=2,width=0.95\textwidth]{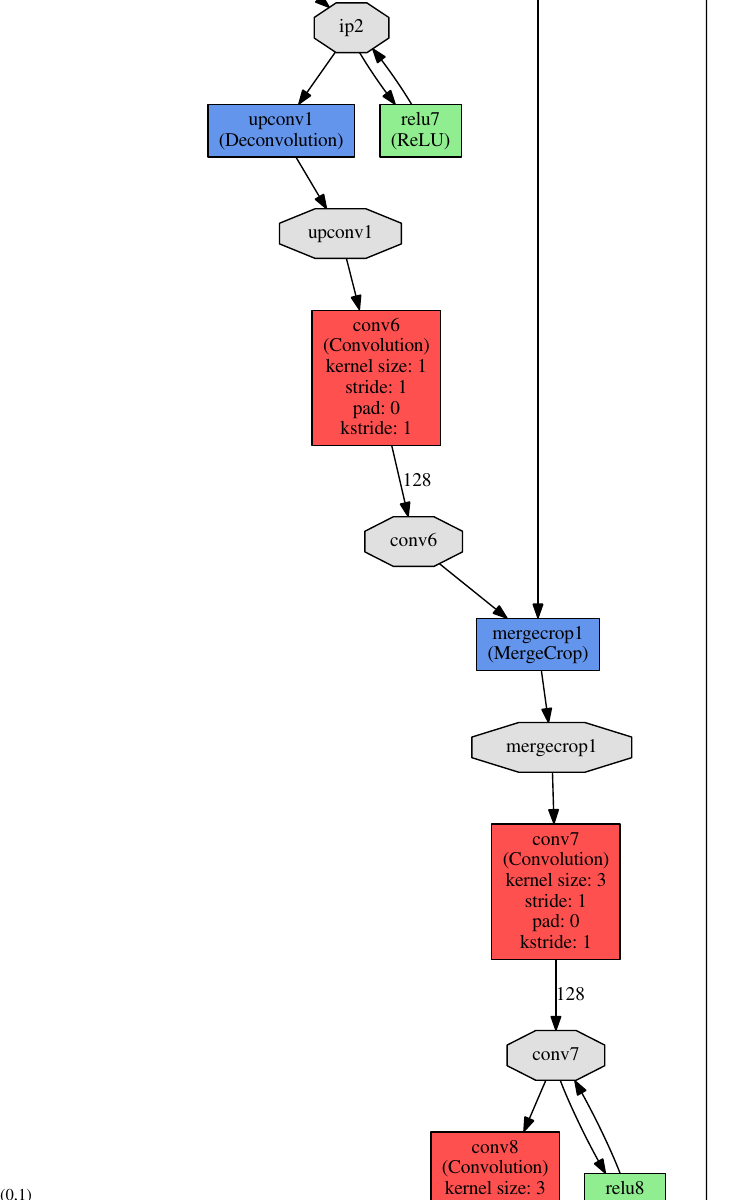}
\end{figure}
\begin{figure}[H]
	\centering
	\includegraphics[page=1,width=0.95\textwidth]{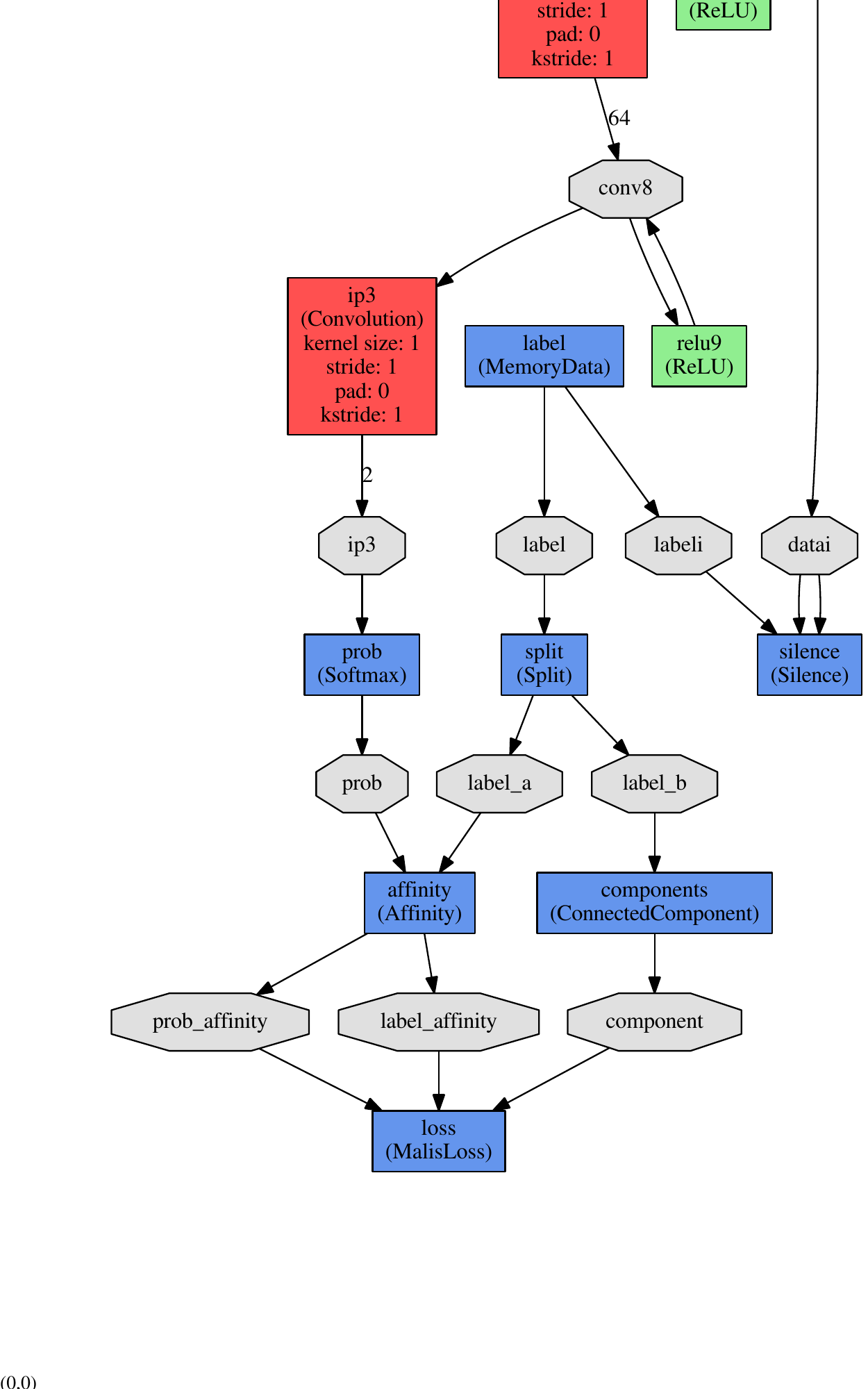}
\end{figure}

\backmatter

\chapter*{References}
\printbibliography[heading=none]

\end{document}